\documentclass[runningheads]{llncs}

\usepackage{eccv}

\usepackage{eccvabbrv}
\newcommand{\name}{CoMind\xspace}

\usepackage{arydshln}
\newcommand{\myparagraph}[1]{\noindent\textbf{#1.}\;}

\usepackage[normalem]{ulem}
\newcommand{\best}[1]{\cellcolor{ForestGreen!20}$\mathbf{{#1}}$}
\newcommand{\second}[1]{\cellcolor{LimeGreen!20}\uline{{$#1$}}}
\newcommand{\third}[1]{\cellcolor{yellow!20}$#1$}

\DeclareRobustCommand{\bestcap}[1]{\colorbox{ForestGreen!20}{\textbf{#1}}}
\DeclareRobustCommand{\secondcap}[1]{\colorbox{LimeGreen!20}{\uline{#1}}}
\DeclareRobustCommand{\thirdcap}[1]{\colorbox{yellow!20}{#1}}

\usepackage[para]{footmisc}
\usepackage{boldline}

\usepackage{silence}
\usepackage{float}
\usepackage{hyperref}
\usepackage{pdfpages}

\usepackage[T1]{fontenc}
\usepackage[utf8]{inputenc}
\usepackage{listings}
\usepackage{zi4} % nicer monospaced font
\usepackage{xcolor}

\definecolor{quoteBg}{HTML}{FCFBF9}
\definecolor{quoteBorder}{HTML}{C8BFB5}
\definecolor{quoteText}{HTML}{2A2A2A}
\definecolor{quoteTitle}{HTML}{7A6D60}

\usepackage{hyperref}  
\usepackage{graphicx}
\usepackage{booktabs}
\PassOptionsToPackage{table}{xcolor}
\usepackage{listings}
\usepackage{wrapfig}
\usepackage{ulem}
\usepackage{nicefrac}       % compact symbols for 1/2, etc.
\usepackage{microtype}      % microtypography
\usepackage{pifont}
\usepackage{wrapfig}
\usepackage{multirow}
\usepackage[skins,listings]{tcolorbox}
\usepackage[accsupp]{axessibility}  % Improves PDF readability for those with disabilities.

\title{\name: Understanding Collaborative Human Activity from Multiple Minds and Views} 
\titlerunning{\name: Understanding Collaborative Human Activity}

\author{Alexey Gavryushin*\textsuperscript{\textdagger}\inst{1} \quad Dingxi Zhang*\inst{1} \quad Zhao Huang*\inst{1} \\
Alexandros Delitzas \inst{1, 2} \quad Jiaqi Chen \inst{1} \quad Ben Ellis \inst{1} \quad Cedric Zöllner \inst{1} \\ Manthan Patel \inst{1} \quad Manuel Kaufmann \inst{1} \quad 
Marc Pollefeys\inst{1,3} \quad Xi Wang \inst{1, 4, 5}}

\authorrunning{A. Gavryushin et al.}

\institute{$^1$ETH Zurich, Switzerland \quad $^2$MPI for Informatics, Germany \quad $^3$Microsoft Switzerland \quad $^4$TU Munich, Germany \quad $^5$MCML, Germany \\
}
\begin{document}

% do NOT put "breakable" here
\newtcblisting{verbatimquote}[1][]{
  enhanced,
  listing only,
  colback=quoteBg,
  colframe=quoteBorder,
  boxrule=0.4pt,
  arc=1mm,
  left=4mm,
  right=4mm,
  top=3mm,
  bottom=3mm,
  width=\textwidth,
  title=#1,
  coltitle=quoteTitle,
  fonttitle=\bfseries\ttfamily\scriptsize,
  before skip=8pt,
  after skip=8pt,
  listing options={
    basicstyle=\ttfamily\scriptsize\color{quoteText},
    breaklines=true,
    breakatwhitespace=false,
    breakindent=0pt,
    postbreak=\mbox{},
    columns=fullflexible,
    keepspaces=true,
    showstringspaces=false,
    xleftmargin=0pt,
    xrightmargin=0pt
  }
}

\maketitle

\begin{NoHyper}
%\vspace{1cm}
\def\thefootnote{*}\footnotetext{Equal contribution, alphabetical order.}\def\thefootnote{\dag}\footnotetext{Corresponding author.}\def\thefootnote{\arabic{footnote}}
\end{NoHyper}

{
\vspace*{-0.8cm}
\begin{figure}
    \centering
    \includegraphics[width=0.94\textwidth]{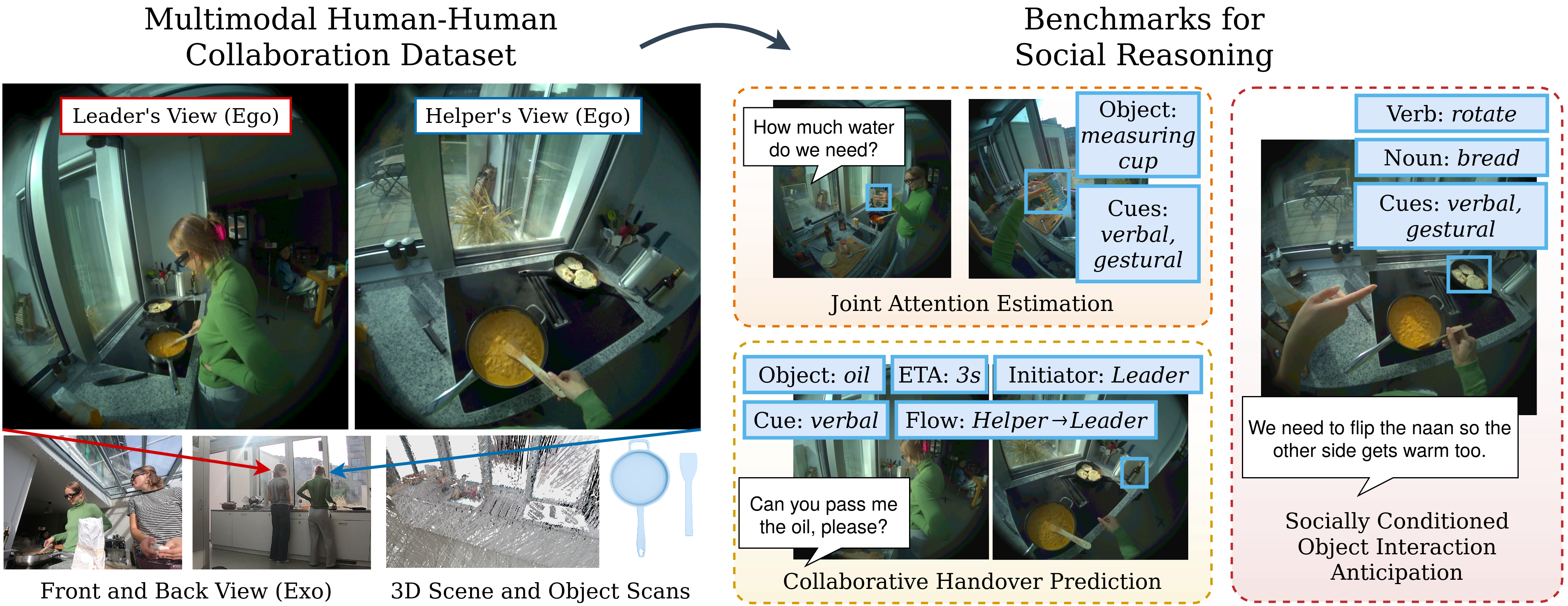}
    \caption{\textbf{The \name dataset} captures 41 hours of human-human collaboration across 80 sessions, combining egocentric/exocentric videos, audio, gaze, and high-quality 3D scene and object scans, as well as rich annotations for three novel social reasoning tasks.}
    % \caption{\textbf{The \name dataset}  features 44 hours of human-human collaboration sessions, captured across 80 recordings with a multitude of modalities and dense with scenarios involving social context. We introduce and provide rich annotations for three novel tasks testing social reasoning and theory of mind.}
    \label{fig:teaser}
\end{figure}
\vspace{-1.2cm}
}

 \begin{abstract}
Human-human collaboration is a fundamental aspect of everyday life, essential to success in a wide range of goal-directed activities from household tasks to professional teamwork. 
While much research has focused on modeling coordination and task execution, the cognitive processes that support such collaboration, particularly Theory of Mind (the ability to infer others’ mental states), remain difficult to study in natural settings. 
To address this gap, we introduce a novel egocentric and exocentric video dataset capturing real-world collaboration in cooking scenarios. 
The dataset integrates multi-perspective video, high-quality audio, gaze tracking, and 3D scene and object scans, with annotations for shared object attention, social cues and interactions between agents, as well as agent-object interactions. %actions, objects, and dialogue. 
We establish benchmarks for Joint Attention Estimation, Action Anticipation, and Collaborative Handover Prediction, enabling research on multimodal perception, proactive assistance, and collaborative planning. 
By providing temporally aligned, richly annotated multimodal data, \name facilitates the development and evaluation of AI systems capable of modeling complex social interactions and reasoning about human behaviors in collaborative environments. Our dataset and benchmarks are made available at \url{https://comind.ethz.ch/}.
\vspace{-0.3cm}
\end{abstract}

\section{Introduction}
\label{sec:intro}

% MAIN POINT OF PARAGRAPH: human collaboration needs to be explicitly studied and modeled
Human collaboration is central to success in many goal-directed activities, from everyday tasks to professional teamwork.
Understanding it is key to building AI systems that can seamlessly work with humans.
Effective collaboration requires the ability to infer and adapt to others’ mental states, intentions, and goals, a concept also known as \textit{Theory of Mind} (ToM) \cite{tomas2005, apperly2010mindreaders}.
ToM is achieved through the continuous exchange of \textit{social cues} \cite{adams2017social}, which are multimodal signals (\eg, eye gaze, body pose, and targeted dialogue) that agents use to establish shared mental states, transmit intent, and facilitate social interaction. 

However, computational modeling of ToM remains underexplored, as previous ToM benchmarks \cite{shi2025muma, rabinowitz2018machine, villa2025moments, li2025egotom, zhang2025metamind, chen2024tombench} mainly focus on text-based reasoning with multiple-choice prompts.
This is due to the lack of datasets capturing the multimodal social cues for human collaboration in natural, real-world settings~\cite{CSIBRA2009148, Schilbach_Timmermans_Reddy_Costall_Bente_Schlicht_Vogeley_2013}.
While egocentric datasets \cite{damen2018scaling, sener2022assembly101, HoloAssist, grauman2022ego4d, zhang2022egobody, jang2019epic, banerjee2024hot3d, huang2024egoexolearn, grauman2024ego, perrett2025hd} capture single-agent activity, few focus on \textit{multi-agent collaboration}. 
Existing datasets with \textit{multiple} humans \cite{jia2020lemma, liu2024core4d, wiederhold2023hoh, zhang2022egobody, khirodkar2023egohumans, guo2023ft} involve short-horizon tasks, limited interactions, or a minimal recording setup that omits critical social cues such as eye gaze or verbal communication. 
Other works in simulation\cite{chang2024partnrbenchmarkplanningreasoning, Watch-And-Help, zhang2024buildingcooperativeembodiedagents} lack the authenticity of unscripted collaboration.
These limitations hinder learning from and modeling ToM in real-world contexts.

% MAIN POINT OF PARAGRAPH: what are we doing more specifically? and why?
In this work, we present \name, a \textbf{new egocentric–exocentric multimodal dataset}, designed to study ToM in collaborative, long-horizon tasks.
Our dataset focuses on unscripted cooking tasks, a domain inherently rich in shared intentions and interactions.
%, and capturing a natural leader-helper dynamic. 
% We designate a ``leader'' to drive the primary task, while a ``helper'' continuously reads social cues to proactively anticipate needs \cite{pesquita2018predictive}, directly mirroring proactive multi-agent collaboration. 
Our dataset features (1) synchronized egocentric and exocentric video, (2) gaze and hand tracking, (3) audio recordings and transcripts, (4) camera trajectories with spatial mapping, (5) dense 3D scene scans, (6) a set of 3D scans of commonly used objects, and (7) fine-grained social cue annotations.

We additionally design \textbf{three tasks} of increasing difficulty to formalize ToM benchmarking.
Our first task, \textit{Joint Attention Estimation}, tests the ability to establish mutual awareness. 
The second task, \textit{Socially Conditioned Object Interaction Anticipation}, predicts an upcoming action and its social cues, testing the ability to infer intent. 
Our third task, \textit{Collaborative Handover Prediction}, predicts the timing and specifics of physical assistance before any reaching motion begins, testing proactive physical coordination.

We benchmark recent vision-language models across these tasks and reveal a significant performance deficiency, showing that current state-of-the-art methods struggle with social perception. 
We further validate the utility of \name by fine-tuning an open-source baseline model on our training data, which yields a substantial improvement in performance and establishes our dataset as a crucial foundation for socially aware AI systems.

\section{Related Work}
\label{sec:related_work}

\begin{table*}[!t]
% \vspace{-5pt}
\centering
\caption{\textbf{Overview of egocentric datasets.} 
% For Modality, \includegraphics[width=0.02\textwidth]{figures/video.pdf} denotes video, \includegraphics[width=0.02\textwidth]{figures/gaze.pdf} denotes gaze, \includegraphics[width=0.02\textwidth]{figures/imu.pdf} denotes IMU, \includegraphics[width=0.015\textwidth]{figures/3d.pdf} denotes 3D scans. 
For Modality, 
\includegraphics[width=0.02\textwidth]{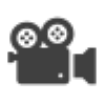} denotes video,
\includegraphics[width=0.02\textwidth]{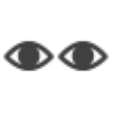} denotes gaze,
\includegraphics[width=0.02\textwidth]{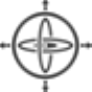} denotes IMU,
\includegraphics[width=0.02\textwidth]{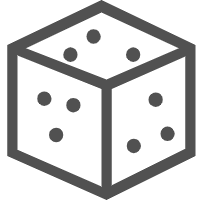} denotes semi-dense point cloud,
\includegraphics[width=0.02\textwidth]{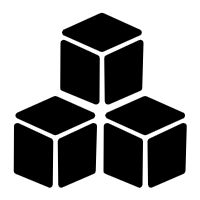} denotes dense 3D scene scans,
\includegraphics[width=0.02\textwidth]{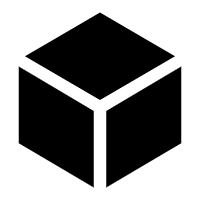} denotes 3D object scans,
\includegraphics[width=0.02\textwidth]{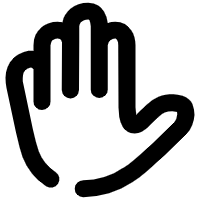} denotes hand tracking. Hours (Ego) refers to the cumulative duration of distinct egocentric video recordings.
Our dataset provides unprecedented depth by offering rich social cues and collaborative annotations alongside multi-view 3D grounding. 
}
\label{tab:compare_dataset}
\huge
% \vspace{-10pt}
\setlength\tabcolsep{15pt}
\resizebox{\textwidth}{!}{
\begin{tabular}{@{}lllcccccccc@{}}
\toprule[3pt]
\textbf{Benchmark} & \textbf{Domain} & \textbf{Modality} & \textbf{Hours (Ego)}   & \textbf{\begin{tabular}[c] {@{}c@{}}Average \\ Duration \end{tabular}} & \textbf{Exo} & \textbf{\begin{tabular}[c]{@{}c@{}}Collaboration\\Dynamics \end{tabular}} & \textbf{\begin{tabular}[c]{@{}c@{}}Social Cue \\ Annotation \end{tabular}} & \textbf{\begin{tabular}[c]{@{}c@{}}\#Humans\\per session \end{tabular}} & \textbf{\begin{tabular}[c]{@{}c@{}}Total \\ participants \end{tabular}} \\
\midrule
EPIC-KITCHENS-100~\cite{Damen2022RESCALING}  &    Kitchen     &          \includegraphics[width=0.05\textwidth]{figures/video.pdf}                     &100        &      8.5 min       &     \textcolor{red}{\ding{55}}      &          \textcolor{red}{\ding{55}} & \textcolor{red}{\ding{55}} & 1       &  32      \\
HD-EPIC~\cite{perrett2025hd} & Kitchen &
 \includegraphics[width=0.05\textwidth]{figures/video.pdf} \includegraphics[width=0.05\textwidth]{figures/gaze.pdf}
\includegraphics[width=0.05\textwidth]{figures/imu.pdf}
\includegraphics[width=0.04\textwidth]{figures/3dpc.png}
\includegraphics[width=0.04\textwidth]{figures/hand.png}
& 41 & 15.9 min & \textcolor{red}{\ding{55}} & \textcolor{red}{\ding{55}} & \textcolor{red}{\ding{55}} & 1 & 9 \\

EPFL-Smart-Kitchen-30~\cite{bonnetto2025epfl} & Kitchen &
 \includegraphics[width=0.05\textwidth]{figures/video.pdf} \includegraphics[width=0.05\textwidth]{figures/gaze.pdf}
\includegraphics[width=0.05\textwidth]{figures/imu.pdf}
\includegraphics[width=0.04\textwidth]{figures/3dpc.png}
\includegraphics[width=0.04\textwidth]{figures/hand.png}
& 29.7 & 35.9 min & \textcolor{teal}{\checkmark} & \textcolor{red}{\ding{55}} & \textcolor{red}{\ding{55}} & 1 & 16 \\

Assembly101~\cite{sener2022assembly101}  &    Task Execution     &          \includegraphics[width=0.05\textwidth]{figures/video.pdf}                     &167        &      7.1 min       &     \textcolor{teal}{\checkmark} &   \textcolor{red}{\ding{55}} & \textcolor{red}{\ding{55}} & 1       &  53     \\
Ego4D~\cite{grauman2022ego4d}      &    Daily Activities    &            \includegraphics[width=0.05\textwidth]{figures/video.pdf} \includegraphics[width=0.05\textwidth]{figures/gaze.pdf}
\includegraphics[width=0.05\textwidth]{figures/imu.pdf}
\includegraphics[width=0.04\textwidth]{figures/3dpc.png}
\includegraphics[width=0.04\textwidth]{figures/hand.png} &3,670        &    22.8 min        &     \textcolor{red}{\ding{55}}      &         \textcolor{red}{\ding{55}} & \textcolor{teal}{\checkmark}  & 1       & 923    \\
EgoExo4D~\cite{grauman2024ego}    &    Skilled Activities      &    \includegraphics[width=0.05\textwidth]{figures/video.pdf} \includegraphics[width=0.05\textwidth]{figures/gaze.pdf}
\includegraphics[width=0.05\textwidth]{figures/imu.pdf}
\includegraphics[width=0.04\textwidth]{figures/3dpc.png} 
\includegraphics[width=0.04\textwidth]{figures/hand.png} &221        &           2.6 min  &    \textcolor{teal}{\checkmark}       &      \textcolor{teal}{\checkmark}  & \textcolor{red}{\ding{55}} & 1-2    & 740         \\
EgoExoLearn~\cite{huang2024egoexolearn}     &   Task Execution      &    \includegraphics[width=0.05\textwidth]{figures/video.pdf} \includegraphics[width=0.05\textwidth]{figures/gaze.pdf}
\includegraphics[width=0.05\textwidth]{figures/imu.pdf} &120         &      13.4 min     &       \textcolor{teal}{\checkmark}      &          \textcolor{red}{\ding{55}}  & \textcolor{red}{\ding{55}} & 1   & -       \\
HoloAssist~\cite{HoloAssist}     &   Assistive Task      &    \includegraphics[width=0.05\textwidth]{figures/video.pdf} \includegraphics[width=0.05\textwidth]{figures/gaze.pdf}
\includegraphics[width=0.05\textwidth]{figures/imu.pdf}
\includegraphics[width=0.04\textwidth]{figures/3dpc.png} 
\includegraphics[width=0.04\textwidth]{figures/hand.png} &166            &      4.8 min     &      \textcolor{red}{\ding{55}}    &          \textcolor{teal}{\checkmark} & \textcolor{red}{\ding{55}} & 1-2     &   222  \\

LEMMA~\cite{jia2020lemma} & Task Execution  & \includegraphics[width=0.05\textwidth]{figures/video.pdf} 
\includegraphics[width=0.04\textwidth]{figures/3dpc.png}      & 10  & 2 min      & \textcolor{teal}{\checkmark}       & \textcolor{teal}{\checkmark} & \textcolor{red}{\ding{55}} & 1-2 & 8 \\

EgoLife~\cite{yang2025egolife} & Daily Life  & \includegraphics[width=0.05\textwidth]{figures/video.pdf} \includegraphics[width=0.05\textwidth]{figures/gaze.pdf}
\includegraphics[width=0.05\textwidth]{figures/imu.pdf}
\includegraphics[width=0.04\textwidth]{figures/3dpc.png}       & 266   & 44.3 h      & \textcolor{teal}{\checkmark}         & \textcolor{teal}{\checkmark} &\textcolor{red}{\ding{55}} & 6 & 6 \\
IndEgo~\cite{Chavan2025IndEgo} & Industrial Tasks  & \includegraphics[width=0.05\textwidth]{figures/video.pdf} \includegraphics[width=0.05\textwidth]{figures/gaze.pdf}
\includegraphics[width=0.05\textwidth]{figures/imu.pdf}
\includegraphics[width=0.04\textwidth]{figures/3dpc.png} 
\includegraphics[width=0.04\textwidth]{figures/hand.png} & 197   & 3.4 min      & \textcolor{teal}{\checkmark}        & \textcolor{teal}{\checkmark} & \textcolor{red}{\ding{55}} & 1-2 & 20 \\
\midrule
\textbf{Ours} & Kitchen  & \includegraphics[width=0.05\textwidth]{figures/video.pdf} \includegraphics[width=0.05\textwidth]{figures/gaze.pdf}
\includegraphics[width=0.05\textwidth]{figures/imu.pdf}
\includegraphics[width=0.04\textwidth]{figures/3dpc.png}  
\includegraphics[width=0.04\textwidth]{figures/3dobject.png}
\includegraphics[width=0.04\textwidth]{figures/3dscene.png}
\includegraphics[width=0.04\textwidth]{figures/hand.png} & 81 &  30.5 min      & \textcolor{teal}{\checkmark}        & \textcolor{teal}{\checkmark} & \textcolor{teal}{\checkmark} & 2 & 125 \\
\bottomrule[3pt]
\end{tabular}}
\vspace{-10pt}
\end{table*}

% \subsection{Ego-Exo Datasets and Benchmarks}
\myparagraph{Ego-Exo Datasets and Benchmarks}
Egocentric vision research has produced several large-scale datasets that capture human activity from first-person perspectives, providing valuable multimodal information such as gaze, hand pose, audio, and depth \cite{damen2018scaling, sener2022assembly101, HoloAssist, grauman2022ego4d, zhang2022egobody, jang2019epic, banerjee2024hot3d, huang2024egoexolearn, grauman2024ego, perrett2025hd}. 
% For example, EPIC-Kitchens \cite{damen2018scaling} focuses on unscripted daily cooking activities, while Ego4D \cite{grauman2022ego4d} expands to over 3,000 hours of egocentric video covering diverse daily life scenarios, both enabling fine-grained action recognition and object interaction benchmarks. 
% Assembly101 \cite{sener2022assembly101} captures long-horizon procedural tasks in furniture assembly for action understanding and hand-object interaction. 
Large scale datasets like EPIC-KITCHENS-100~\cite{damen2018scaling, Damen2022RESCALING} and Ego4D \cite{grauman2022ego4d} have significantly advanced single-agent action understanding through multi-modal annotation.
More recent efforts capture ego- and exocentric paired views \cite{huang2024egoexolearn, grauman2024ego, jia2020lemma, qiu2025egome, sener2022assembly101} to bridge first- and third-person perspectives. 
Ego-Exo4D \cite{grauman2024ego} synchronizes head cameras with external views for skilled physical activities, while HoloAssist \cite{HoloAssist} captures multi-view instructional collaboration. 
However, these datasets still predominantly treat the primary actor in isolation, focusing on procedural execution rather than joint coordination. 
Conversely, socially aware datasets \cite{zhang2022egobody, khirodkar2023egohumans, mclean2025embody3d} capture multi-person interactions but focus on conversational dynamics and human poses rather than shared physical goals. 
As detailed in Table \ref{tab:compare_dataset}, existing literature lacks works that capture continuous, goal-driven physical collaboration with detailed annotations for social cues, which are important for modeling intent during collaboration. 
Our dataset addresses this by integrating long-context multi-agent ego-exo videos with diverse sensor data, rich annotations, and 3D scans of the environment and objects, providing the multimodal data sources necessary to study real-world human-human collaboration from multiple viewpoints.
% Ego-Exo4D \cite{grauman2024ego} links synchronized egocentric headcams with external camera views, while HoloAssist \cite{HoloAssist} augments egocentric video with task instructions and multi-view capture to facilitate step-by-step assistance research.
% In socially aware settings, datasets such as EgoBody \cite{zhang2022egobody} and EgoHuman \cite{khirodkar2023egohumans} incorporate multi-person interactions, body pose, and gaze, but typically center on short-term or structured exchanges rather than extended, goal-driven, and natural collaboration.

% Although these works have substantially advanced understanding of perception and action from a first-person viewpoint, most remain focused on single-agent activities or brief dyadic interactions as shown in Table \ref{tab:compare_dataset}. They rarely capture the rich, temporally extended, multi-modal collaboration dynamics—such as joint attention shifts, proactive assistance, and role negotiation—required for studying ToM in real-world tasks.

% In contrast, within our dataset, these brief collaboration-reflecting motions are annotated with participant dialogues and a broader task context that focuses on the achievement of specific goals, providing a more comprehensive and context-rich understanding.

% The insights from these datasets, particularly in activity understanding and object interaction, provide essential foundations for our work, where we seek to understand how agents perceive and react to the collaborative environment from an individual agent's perspective.

% \subsection{Multi-Agent Collaboration}
\myparagraph{Multi-Agent Collaboration}
Beyond single-agent settings, recent works \cite{zhang2024buildingcooperativeembodiedagents, chang2024partnrbenchmarkplanningreasoning, Watch-And-Help, shridhar2020alfred, puig2018virtualhomesimulatinghouseholdactivities} have investigated multi-agent collaboration in simulation, examining how humans and embodied agents coordinate in shared tasks. 
% PARTNR \cite{chang2024partnrbenchmarkplanningreasoning} leverages large language models for task planning and evaluation in Habitat 3.0 \cite{puig2023habitat}, while CoELA \cite{zhang2024buildingcooperativeembodiedagents} introduces memory-augmented agents to optimize communication for collaborative tasks. 
% Watch-And-Help \cite{Watch-And-Help} enables goal inference from demonstration. 
% However, these works lack multimodal real-world cues, mutual assistance, and typically address short-horizon tasks with clearly defined steps.
While these simulators excel at modeling high-level task coordination, they abstract away the sensor noise and lack the rich, multimodal real-world cues essential for natural human teamwork.
On the other hand, real-world datasets \cite{jia2020lemma, liu2024core4d, khirodkar2023egohumans, zhang2022egobody, wiederhold2023hoh, yang2025egolife, Chavan2025IndEgo} capture the complexity of physical interaction in shared spaces. 
% Specifically, 
LEMMA \cite{jia2020lemma} collects multi-view videos in short-term collaborative activities without verbal communication; HOH \cite{wiederhold2023hoh} focuses on object handover sequences and geometric execution of the process; EgoLife \cite{yang2025egolife} targets daily household activities for training long-term egocentric assistants, and Indego \cite{Chavan2025IndEgo} focuses on industrial scenarios and procedural task understanding. 
While these datasets offer rich sensory information, most center on short-duration or narrowly scoped interactions and provide limited insight into the dynamic and socially cued interactions in multi-agent collaboration.
\myparagraph{Benchmarking Theory of Mind}
\label{subsec:benmark_tom}
While the importance of Theory of Mind (ToM) in collaboration is well-established \cite{tom1978, jointaction}, its computational evaluation remains a challenge.
Recent works benchmark modern AI systems, such as multimodal large language models, for ToM\cite{shi2025muma, rabinowitz2018machine, villa2025moments, li2025egotom, zhang2025metamind, chen2024tombench}. 
However, these works typically evaluate for ToM as a passive Video Question Answering task, probing abstract belief systems through text, often using unrealistic multiple-choice prompts. 
In natural physical collaboration, there are no predefined prompts; agents must continuously infer mental states and respond with real-time physical actions. 
Therefore, rather than performing post-hoc textual analysis, we target open-ended, complex ToM reasoning. %high-level embodied reasoning. %low-level embodied reasoning. 
To benchmark for this ToM ability, we formalize Theory of Mind as three interdependent vision tasks.
%To computationally model this actionable social perception, our dataset grounds ToM in three highly interdependent, observable vision tasks:

% \paragraph{Joint Attention Detection.} 
\noindent\textit{Joint Attention Estimation.} 
Gaze behavior is a fundamental cue for social interactions. 
While gaze estimation has been extensively studied in exocentric \cite{recasens2015they, chong2020detecting} and egocentric settings \cite{fathi2012learning, grauman2022ego4d, qiu2025egome}, most works predict \textit{single} agent gaze targets. 
Multi-agent research has addressed shared attention \cite{fan2019understanding, de2023temporal} and mutual gaze \cite{marin2014detecting, medina2021suarez}, but typically in third-person videos that lack the synchronized first-person perspective needed for understanding egocentric collaboration. 
Egocentric works on social gaze \cite{ye2015detecting, huang2020ego} explore when another person looks at the camera wearer, and few works explicitly target joint attention in goal-oriented tasks. 
By providing synchronized multi-agent gaze data, we frame ``joint attention'' as not just identifying the geometric gaze intersection, but as understanding shared mental attention \cite{reddy2005understanding}, which is paramount to understanding socially cued actions.

% \paragraph{Socially Conditioned Object Interaction Anticipation.} 
\noindent\textit{Object Interaction Anticipation.} 
Object interaction anticipation has been widely studied in exocentric \cite{vondrick2016anticipating,abu2018will,gao2017red,chen2022gatehub,rizve2023pivotal} and egocentric \cite{rodin2022untrimmed,qi2021self,girdhar2021anticipative,furnari2022towards,damen2018scaling,grauman2022ego4d, pasca2024transfusion} videos. 
Approaches range from recurrent models \cite{furnari2020rolling,osman2021slowfast,shi2018action} and transformer-based methods \cite{girdhar2021anticipative,grauman2022ego4d,mascaro2023intention} to multimodal variants that integrate audio or object features \cite{furnari2020rolling,zhang2023object,mittal2022learning}, and more recently to LLMs \cite{zhao2023antgpt,huang2023palm}. 
However, most egocentric action anticipation works focus on single-agents. 
While works like LEMMA \cite{jia2020lemma} incorporate multi-agent scenarios using 3D-CNNs \cite{osman2021slowfast} and transformers \cite{himemformer}, they generally predict individual actions independently, ignoring inter-agent conditioned behaviors.
Our benchmark addresses this critical gap %by introducing Socially Conditioned Object Interaction Anticipation. Instead of predicting isolated behaviors based on an individual's past, our task requires 
anticipating a \textit{partner's} future STA variables (verb, noun, and spatial bounding box) strictly conditioned on prior social cues.
% \todo{this section should be more 'object interaction' focused, similar to our task}

% \paragraph{Collaborative Handover Prediction.} 
\noindent\textit{Collaborative Handover Prediction.} 
The physical transfer of objects is a critical component of collaboration. 
Previous research studies handovers from a kinematic, grasp synthesis, and safety perspective \cite{strabala2013towards, taheri2020grab, zhang2023object, fu2025gigahands}, and predominantly within the context of Human-Robot Interaction~\cite{ortenzi2021object, cakmak2011human}.
% In computer vision, works analyzing human-human handovers \cite{liu2022joint, mascaro2023intention, wiederhold2023hoh} have successfully categorized isolated handover phases with the focus on the reactive, geometric execution of the transfer. 
In computer vision, works analyzing handovers \cite{liu2022joint, mascaro2023intention, wiederhold2023hoh} focus on segmenting and reconstructing the mechanics of the handover itself (i.e. hand trajectories, grasp phases, contact points).
Rather than just detecting the geometric reconstruction of interacting hands, our task serves to understand the cognitive precursors of a ``handover'' before any reaching motion occurs: who will hand what to whom, when, and prompted by what social cue if at all. 
We frame ``handover prediction'' as proactive intent inference, and measure a model's ability to anticipate agents' needs and provide seamless assistance.

\section{The \name Tasks and Dataset}

% \todo{write a brief overview of the structure of this section; also motivate why we decide to record in the kitchen }
This section introduces the \name dataset and the three tasks we design to formalize ToM benchmarking.
We begin by formally defining each task in \autoref{sec:task_definition}, describing the inputs, outputs, and evaluation objectives for Joint Attention Estimation, Socially Conditioned Object Interaction Anticipation, and Collaborative Handover Prediction. We then subsequently describe the data collection protocols, data processing, annotation pipeline, and dataset statistics.

\subsection{Proposed Tasks}
\label{sec:task_definition}

% \todo{Dingxi please take care of this part}\\

We propose three novel tasks designed to evaluate a model's ability to interpret social cues and understand multi-agent coordination.
% The tasks include: predicting jointly attended objects from synchronized egocentric views of two agents, anticipating socially conditioned actions of a first-person agent, as well as correctly estimating an object handover event between two agents. --> i think we mention this enough already in related works?
We designate a ``leader'' and a ``helper'' for each task, where the leader drives the primary task and the helper must proactively anticipate the leader's needs by understanding social cues.
This directly mirrors proactive multi-agent collaboration.
All tasks are grounded in a collaborative cooking scenario, as the kitchen provides a natural yet controlled environment well-suited for studying social coordination.
% This achieves a collaborative experience where multi-agents work together to achieve the same goals.

% \input{figures/joint_attention_examples}
\begin{figure*}[t]
\centering
\begin{minipage}{0.58\textwidth}
  \centering
  \includegraphics[width=\linewidth]{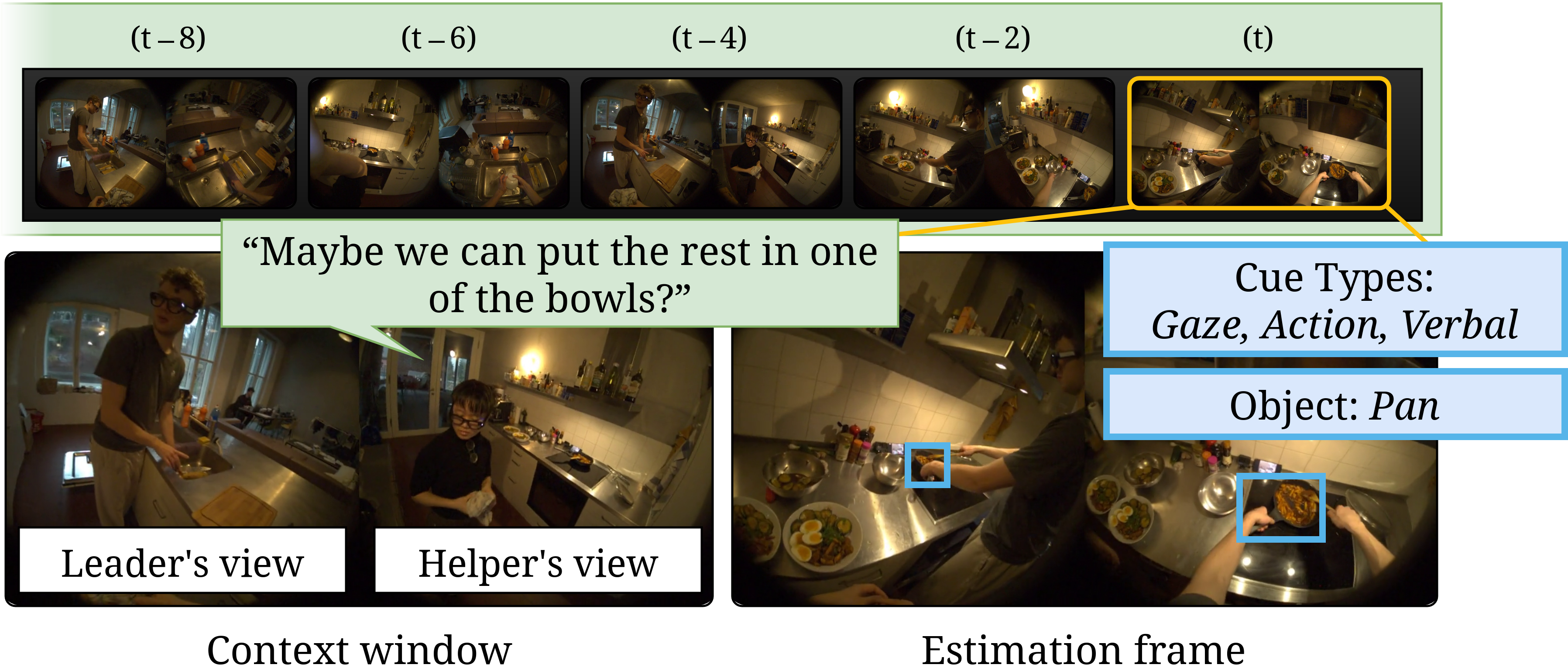}
\caption{\textbf{Joint Attention Estimation Task.} From the audio or transcribed speech in a context window, together with a target frame (green), the model predicts the jointly attended object, its bounding box in the left and right views, and the social cue type (blue).}
  \label{fig:joint-attention-task-figure}
\end{minipage}
\hfill
\begin{minipage}{0.38\textwidth}
  \centering
  \includegraphics[width=\linewidth]{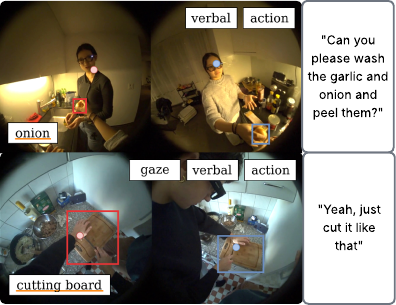}
\caption{\textbf{Example annotations.} Frames show joint attention cues, gaze points, and object bounding boxes with labels.}\label{fig:joint-attention-examples}
\end{minipage}
\vspace{-10pt}
\end{figure*}

\myparagraph{Task 1: Joint Attention Estimation} 
We define joint attention as a state of temporally and spatially synchronized focus on a shared semantic entity. 
The Joint Attention Estimation task is shown in \autoref{fig:joint-attention-task-figure}: For a frame at time $t$ (with a known ``joint attention'' event), given the associated context in the form of audio or transcribed speech  $T_{t'\rightarrow t}$ of the last 10 seconds ($t'=t-10s$) for both participants $p\in\{\ell,h\}$ and a final prediction frame $V^{(\ell,h)}_t$ showing both views, we predict
% \begin{equation}
$(b^{(l)}, b^{(h)}, c, q)$,
% \end{equation}
where $b^{(p)} = (x^{(p)}, y^{(p)}, w^{(p)}, h^{(p)})$  denotes a view-specific bounding box of the jointly attended object for each participant $p$. 
We further predict an object category $c$, as well as the type $q \subseteq \{\text{action}, \text{gaze}, \text{verbal}, \text{gestural}\}$ of the cue eliciting the joint attention on the object. %\todo{task figure and maybe 1 sentence at start to make this intuitive}
% \input{figures/joint_attention_task_figure}
% \todo{add a sentence similar to the last sentence of task 3}
This task measures the capacity to recognize a collaborative agent's shared mental state.

\begin{figure*}[t]
\centering
\begin{minipage}{0.65\textwidth}
  \centering
  \includegraphics[width=\linewidth]{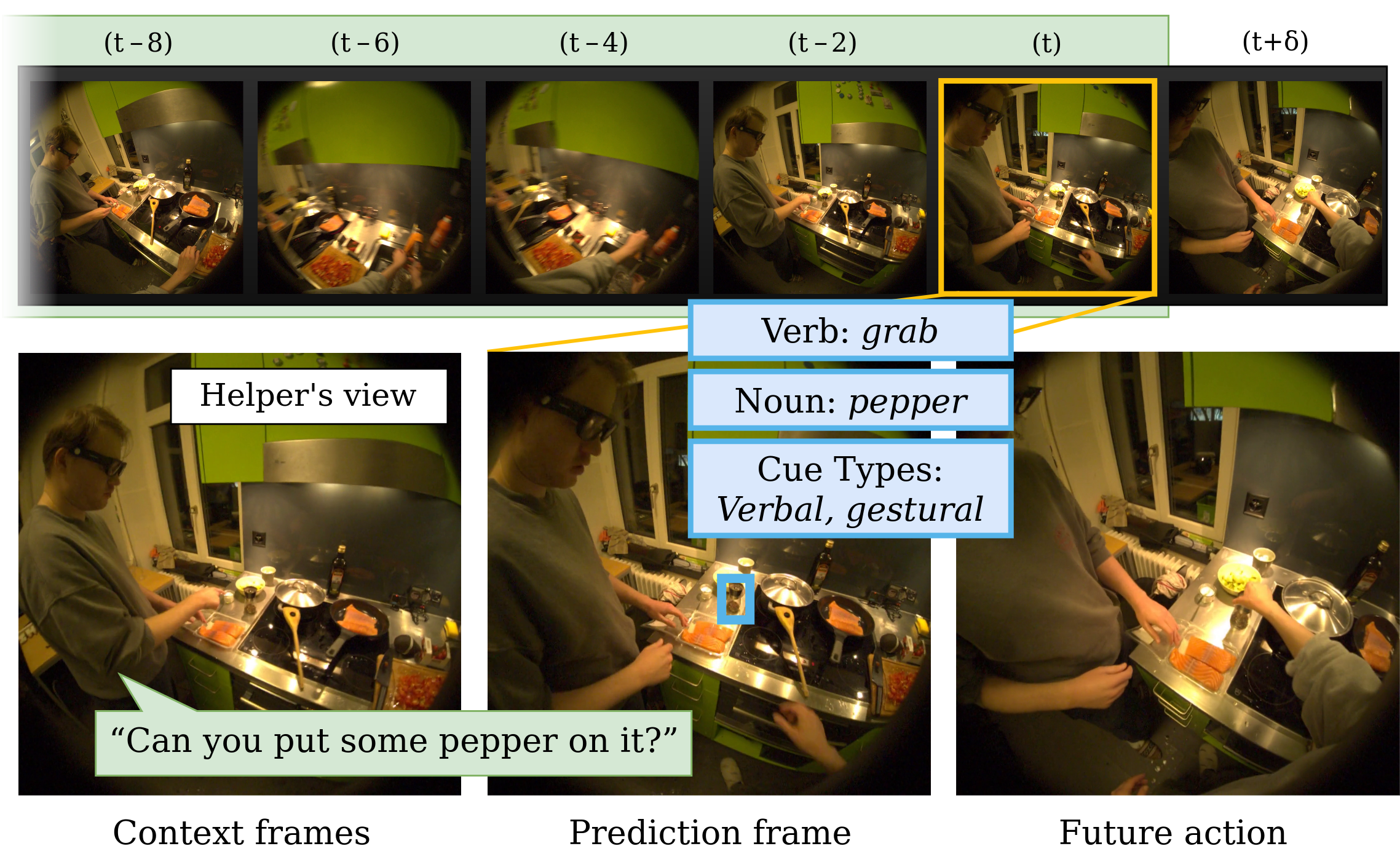}
  \caption{\textbf{Socially Conditioned Object Interaction Anticipation Task.} From context frames (or video) and audio or transcribed speech together with a prediction frame (green), the model predicts the noun and verb of the action to be performed, the interacted object's bounding box, and the social cue type (blue).}
  \label{fig:scoia-task-figure}
\end{minipage}
\hfill
\begin{minipage}{0.31\textwidth}
  \centering
  \includegraphics[width=\linewidth]{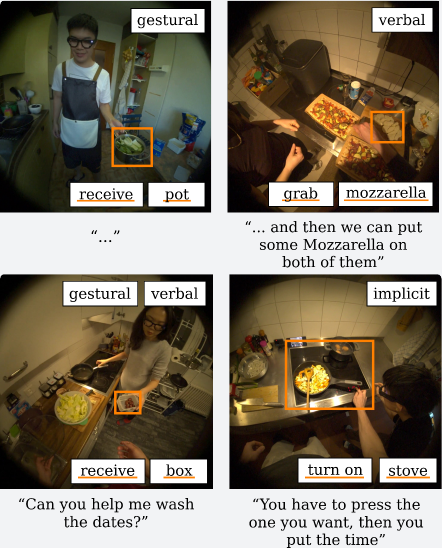}
  \caption{\textbf{Example annotations.} The prediction frame shows the ground-truth action verb, noun, object bounding box, and cue type(s).}
  \label{fig:scoia-examples}
\end{minipage}
\vspace{-16pt}
\end{figure*}

\myparagraph{Task 2: Socially Conditioned Object Interaction Anticipation} 
We define a socially conditioned action of participant $i \in \{\ell,h\}$ as any action performed that is directly triggered by a social cue from the other participant $j \in \{\ell,h\}, j\ne i$. In our setting, we fix $i=h$, i.e.~the action is always cued by the leader and performed by the helper. 
We fix $i=h$ to mimic a scenario where a helper agent, \ie, a robot, must predict what is expected of it.
% \todo{add motivation and explain why this fix of i=h makes sense}
%

%
Given the helper's recent observations as a frame history $V^{(h)}_{t'\rightarrow t}$ and audio or transcribed speech $T_{t'\rightarrow t}$ of the past 10 seconds ($t' = t - 10s$) of the collaborative session up to a frame $t$, the goal of this task is to predict the socially conditioned action which the helper is expected to perform in the immediate future after $t$. 
% \todo{task figure and maybe 1 sentence at start to make this intuitive} % comprising egocentric RGB frames, synchronized third-person views, audio, transcribed speech, gaze, and pose features (subset selectable depending on the benchmark variant).
We thus aim to predict the next structured action $a^{(h)}_{t+\delta}$ for $h$ (the helper) that has been socially conditioned on $\ell$ (the leader) at a future, unknown offset $0 < \delta \leq 10$ seconds.
The output of the task is
% \begin{equation}
$(v, n, b, q)$,
% \end{equation}
where we use a verb $v\in \mathcal{V}$ (e.g.~\textit{take}) and noun $n\in \mathcal{N}$ (e.g.~\textit{fork}) describing the action (e.g.~\textit{take fork}) that the helper will perform at time $t+\delta$. Here, $\mathcal{V}$ and $\mathcal{N}$ represent the verb and noun taxonomy of our dataset. Additionally, we predict a bounding box $b = (x,y,w,h)$ enclosing the object that the helper will interact with, in the 2D coordinates of the frame at $t+\delta$ of $h's$ video stream. Lastly, we also predict the cue type $q \in \{\mathrm{gestural}, \mathrm{verbal}, \mathrm{gestural \wedge verbal}, \mathrm{implicit}\}$ of the social cue which can be used to infer $v, n, b$.
%The task is illustrated in 
\autoref{fig:scoia-examples} shows a few examples. 
% \todo{add a sentence similar to the last sentence of task 3}
This task measures the capacity to anticipate a collaborative agent's intent, and grasp the social cues preceding the intent.

\myparagraph{Task 3: Collaborative Handover Prediction}
\begin{figure*}[t]
\centering
\begin{minipage}{0.6\textwidth}
  \centering
  \includegraphics[width=\linewidth]{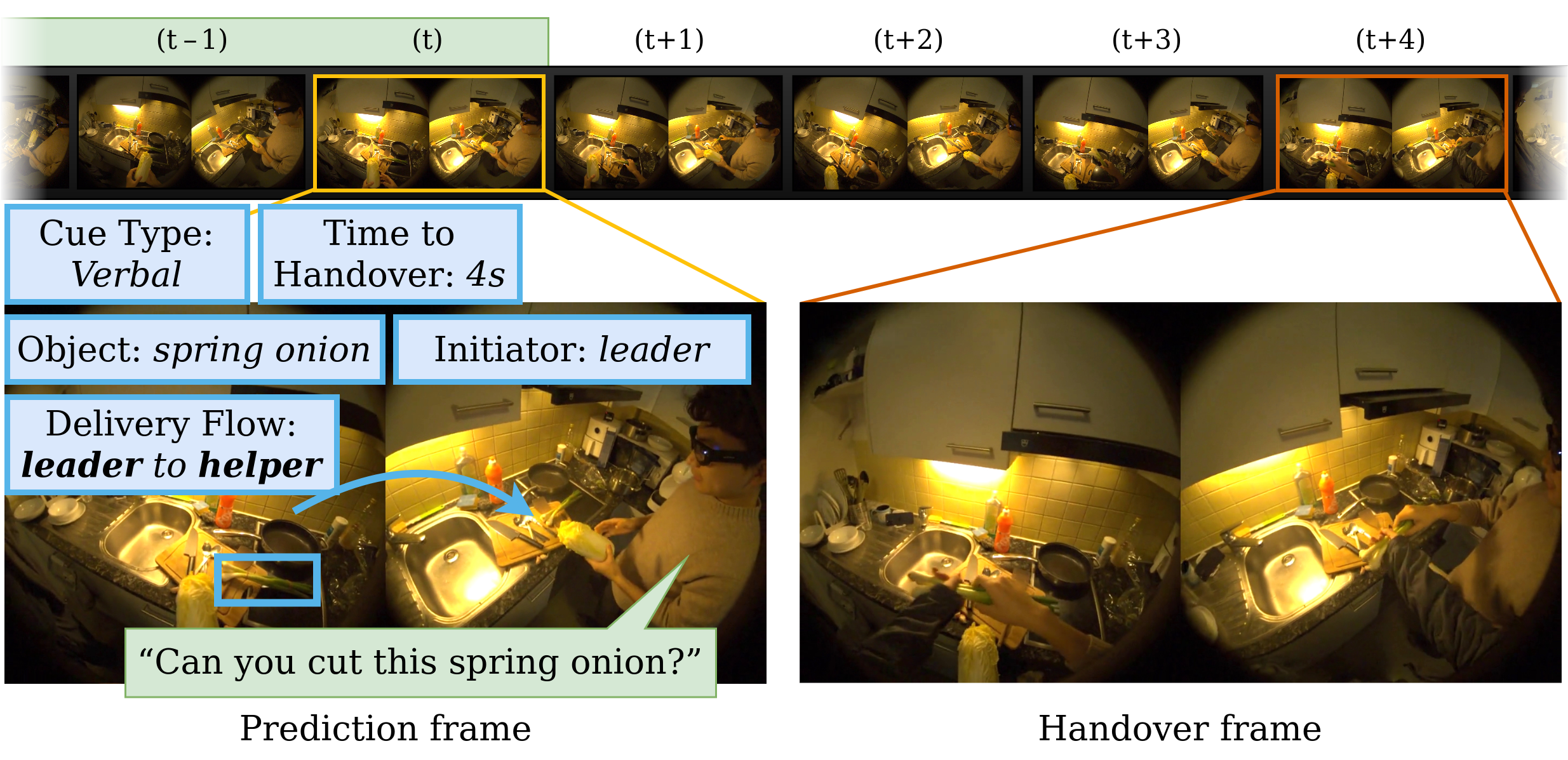}
\caption{\textbf{Collaborative Handover Prediction Task.} From context frames (or video) and audio or transcribed speech, together with a prediction frame (green), the model predicts the delivery flow (who hands to whom), the object category and bounding box in the view of the handing participant, the initiator, and the cue type (blue).}
\label{fig:handover_examples}
\end{minipage}
\hfill
\begin{minipage}{0.36\textwidth}
  \centering
  \includegraphics[width=\linewidth]{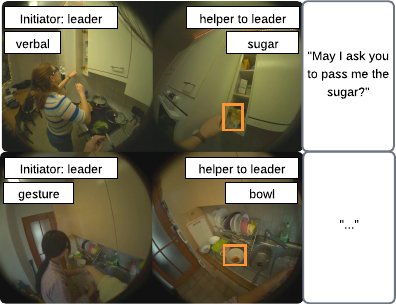}
\caption{\textbf{Example annotations.} Frames show handover initiator, cue type, delivery flow, and object bounding boxes with labels.}\label{fig:handover-examples}
\end{minipage}
\vspace{-16pt}
\end{figure*}
We illustrate the task in \autoref{fig:handover_examples}. Given a prediction frame $f$ \emph{before} any reaching or grasping motion towards a handed-over object, together with the two synchronized egocentric video clips
${V}^{(\ell)}_{t'\rightarrow t}$ and ${V}^{(h)}_{t'\rightarrow t}$,  
captured for 10 seconds before $t$ ($t' = t - 10s$), as well as the audio or transcribed speech $T_{t'\rightarrow t}$, the goal is to anticipate the specifics of the upcoming handover event using $f$ as a reference.
The task is formulated as predicting a single handover event
% \begin{equation}
$(\tau, s \rightarrow r, i, b, c, q)$,
% \end{equation}
where $\tau > 0$ denotes how long after $t$ the handover will occur (also referred to as Time-To-Handover, TTH),
$s \rightarrow r$ indicates the sender-to-receiver object delivery flow with $s, r \in \{\ell,h\}$ and $s \neq r$, $i \in \{\ell, h\}$ represents the person initiating the handover through a social cue,
$b=(x,y,w,h)$ is a bounding box localizing the object in the sender’s view when the leader and helper views are concatenated horizontally, $c$ is the object category, and
$q \subseteq \{\text{verbal}, \text{gestural}, \text{implicit}\}$ specifies the social cue type(s) used by $i$ to initiate the handover.
This task measures the ability to infer collaborative intent and partner need ahead of physical interaction, reflecting realistic assistive scenarios such as tool use in shared tasks.

\label{sec:dataset_method}

\subsection{Data Recording Protocol and Hardware Setup} 
\begin{figure}[t]
  \begin{minipage}[t]{0.65\textwidth}
    \includegraphics[width=\textwidth]{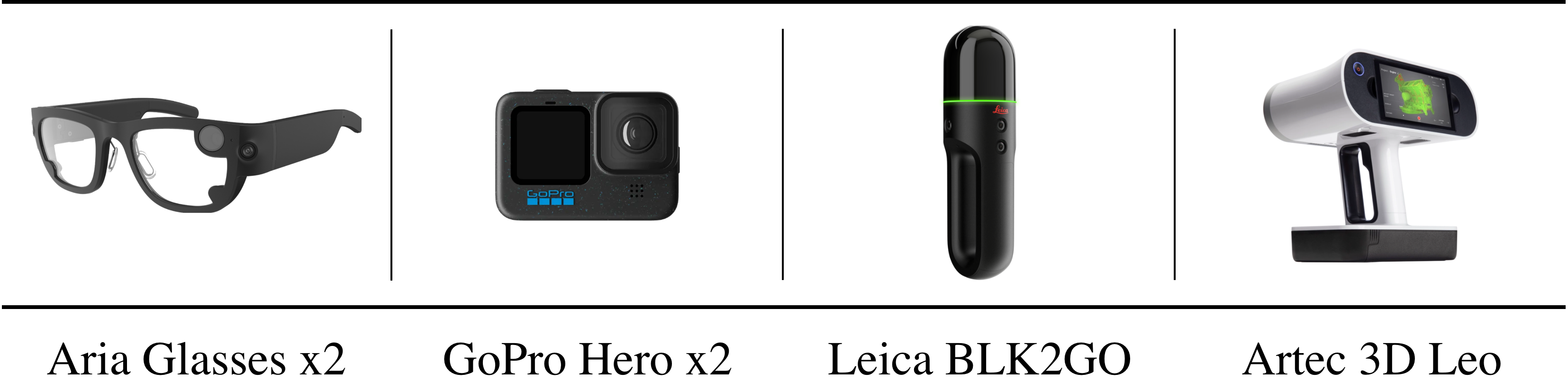}
    \vspace{-12pt}
    \caption{\textbf{Sensors used to create the CoMind dataset.} Two Aria glasses record the scene from each participant's view. Two GoPro Hero cameras capture exocentric views. The Leica BLK2Go and Artec 3D Leo are used for 3D scene and object scanning respectively. }
    % Participants record egocentric data using the head-worn Aria glasses. Point clouds of kitchens are obtained using the hand-held Leica BLK2GO scanner. Objects used for cooking are 3D-scanned using the Artec 3D Leo. Exocentric views of the participants are captured using two wide-angle GoPro Hero cameras.
    \label{fig:comind-hardware}
  \end{minipage}
  \hfill
  \begin{minipage}[t]{0.31\textwidth}
    \includegraphics[width=\textwidth]{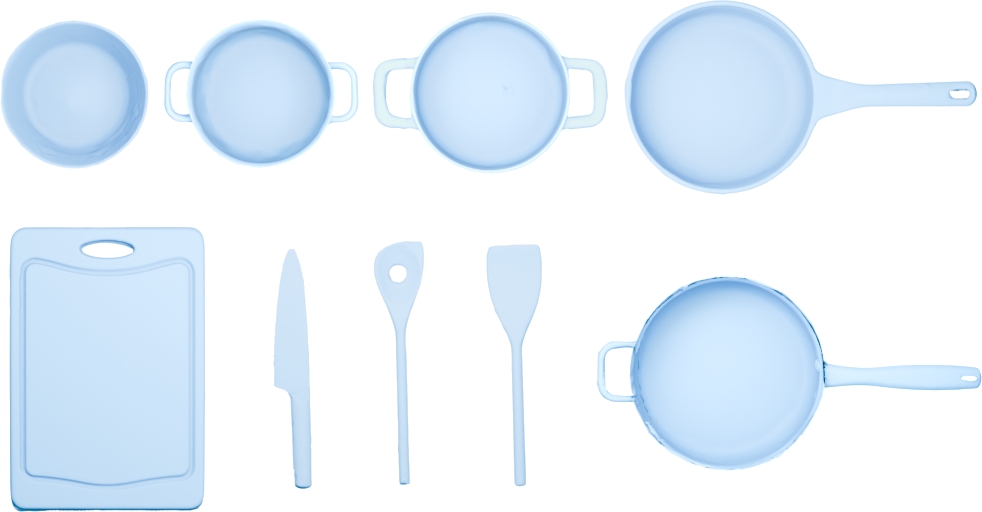}
    \vspace{-12pt}
    \caption{\textbf{3D scans of objects used in the recordings.} Participants work with objects that have been 3D-scanned. %to allow for applications such as object instance recognition and object pose reconstruction.
    }  % We let participants work with a common set of objects throughout all recordings, which were scanned to allow for applications such as object instance recognition and object pose reconstruction.
    \label{fig:object_meshes}
  \end{minipage}
  \vspace{-16pt}
\end{figure}

Data collection takes place in real-world kitchens to ensure a natural and diverse dataset, and all participants provide written informed consent.
Both participants wear the \textbf{Meta Aria Glasses}~\cite{engel2023projectarianewtool} to capture egocentric multi-modal recordings.
Prior to each session, we calibrate intrinsics of the two \textbf{GoPro cameras}, and the Aria glasses are calibrated for each participant for accurate eye-gaze tracking. 
We place one GoPro camera in front of the cooking area, capturing the participants from the front and providing a clear view of their upper-body and hand movements throughout the recording. 
A second GoPro is placed behind the cooking area, recording the participants from the back to ensure full-body visibility.
The Aria glasses are triggered to start recording synchronously.
A trained data collector oversees the recording session and monitors it remotely from another room via a video stream provided by a mobile phone placed in the kitchen space. 
Moreover, a dense 3D scan of the kitchen is captured by the data collector using a \textbf{Leica BLK2GO scanner}  prior to the session.
Lastly, we reuse a set of common cooking utensils (pans and pots) for each session, which are scanned using an \textbf{Artec 3D Leo} \cite{artec_leo_scanner} scanner to obtain high-fidelity object meshes. 
We provide a visualization of the sensors and hardware in \autoref{fig:comind-hardware} and the objects meshes in \autoref{fig:object_meshes}. 

%\begin{figure}[t]
 % \centering
  %\includegraphics[width=\textwidth]{figures/sensor_hardware_v11.png}
  %\includegraphics[width=0.65\textwidth]{figures/sensor_hardware_left.png}
  %\hfill
  %\includegraphics[width=0.27\textwidth]{figures/object_meshes_right_v2.png}
  %\caption{\textbf{Hardware used to create the CoMind dataset.} Participants record egocentric data using the head-worn Aria glasses. Point clouds of kitchens are obtained using the hand-held Leica BLK2GO scanner. Objects used for cooking are 3D-scanned using the Artec 3D Leo. Exocentric views of the participants are captured using two wide-angle GoPro Hero cameras.}
  %\label{fig:comind-hardware}
%\end{figure}

\subsection{Data processing}

%\todo{Zhao, Manthan, Alex}

\noindent\textbf{Synchronization.}
First, the Aria recordings are temporally aligned to sub-millisecond precision using TICSync~\cite{meta_mps_projectaria}.
Second, the GoPro recordings are time-synced to the Arias with SyncSink~\cite{six2015multimodal} using audio cross-correlation.
We also manually trim recordings to the same starting timestamp, to eliminate irrelevant frames before the start of the actual cooking.

\begin{figure}[t]
    \centering
    \includegraphics[width=0.98\textwidth]{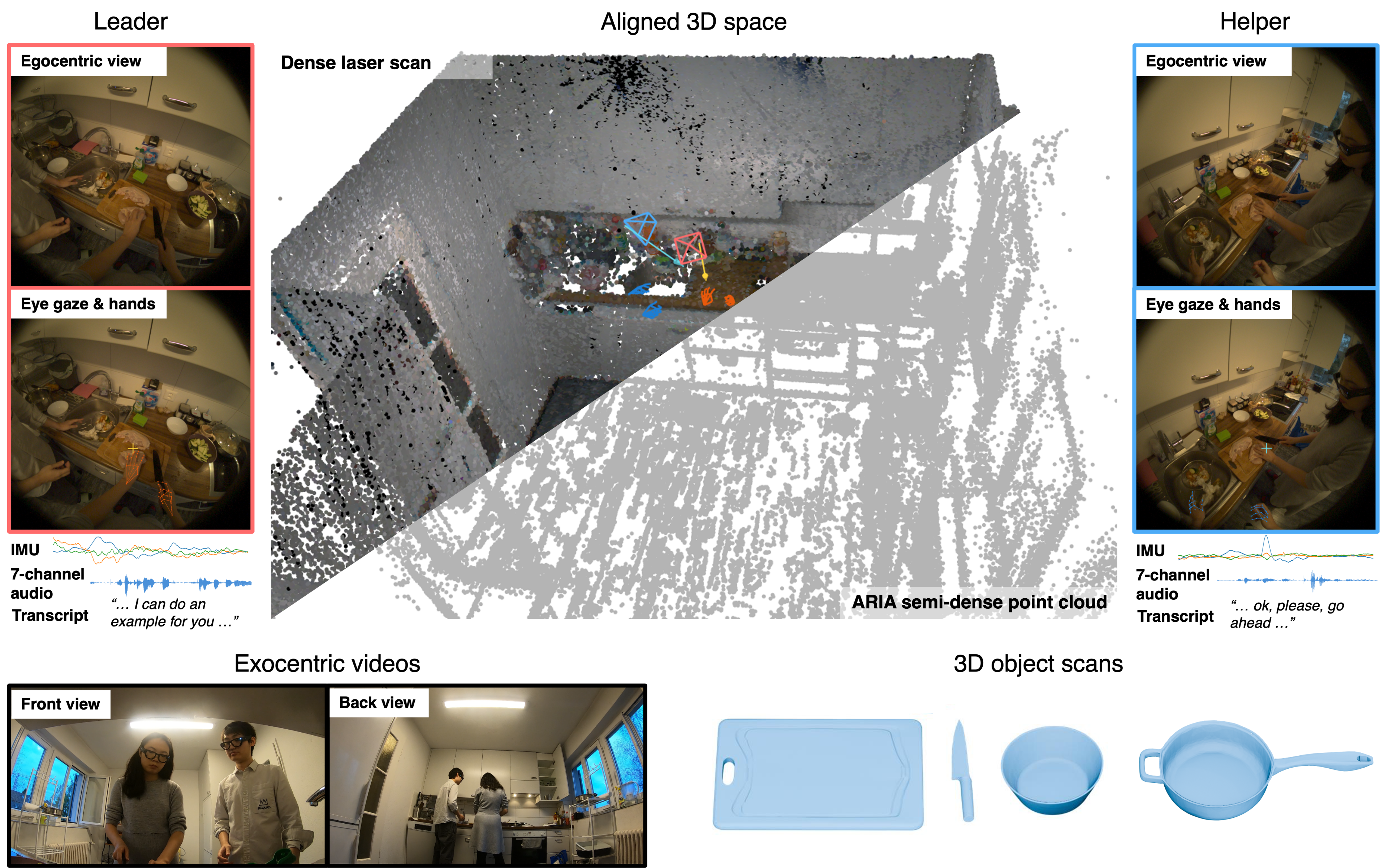}
    % \caption{\textbf{Modalities in the CoMind dataset.} The dataset provides synchronized multimodal recordings of collaborative sessions, including egocentric video with eye gaze and hand tracking, audio and IMU signals, exocentric camera views, an aligned 3D dense reconstruction of the scene, and high-quality 3D scans of the objects used in the environment.}
    \caption{\textbf{Modalities in the CoMind dataset.} The dataset provides synchronized multimodal recordings of collaborative sessions, including egocentric video with eye gaze and hand tracking, audio and IMU signals, exocentric camera views, camera trajectories and spatial mapping, dense 3D scans of the scene, and a set of high-quality 3D scans of commonly used objects.}
    \label{fig:modalities}
    \vspace{-8pt}
\end{figure}

\noindent\textbf{Post Processing.}
% We make available the raw and processed data from all sensors.
% The processed outputs from the Aria recordings include eye gaze and hand pose estimation, and camera trajectory data, obtained using Meta's Machine Perception Services~\cite{meta_mps_projectaria}. 
% For each recording, we provide a high-fidelity RGB point cloud of the environment from the BLK2GO Scanner, which is aligned to the coordinate frame of the egocentric hand and camera pose data.
We process all Aria recordings using the Machine Perception Services (MPS)~\cite{meta_mps_projectaria} to obtain high-fidelity spatial and behavioral data. This pipeline provides 6DoF camera trajectories, semi-dense point clouds, eye-gaze, and hand tracking. Specifically, the Multi-SLAM output from MPS is used to align the data from both participants into a shared coordinate frame, enabling consistent spatial reasoning across each participant pair.
Additionally, we extract audio transcripts of the narrations and dialogue using the WhisperX speech recognition model \cite{radford2023robust, bain2023whisperxtimeaccuratespeechtranscription}.
All modalities available in our dataset are visualized in \autoref{fig:modalities}. 

% \noindent\textbf{Trimming.}
% To eliminate irrelevant segments captured during device initialization, we perform temporal trimming on all streams. 
% As recording is initiated during the synchronization protocol, a latency period containing non-task interactions precedes the actual task commencement. 
% To reduce computational overhead and remove motion noise, we manually annotate a single start timestamp for a reference device and propagate this temporal boundary to all concurrent streams using the established synchronization mappings. 
% This ensures a globally consistent, task-centric temporal window across all multimodal data.

\noindent\textbf{Scan Alignment.}
To transform the 3D scan of the environment into Aria's world frame, we follow a two-stage procedure. In the first stage, we localize the BLK2GO images within Aria's semi-dense point cloud using Hierarchical Localization~\cite{sarlin2019coarse, sarlin2020superglue}.  Each successfully localized image yields an independent estimate of the rigid transformation between the two coordinate frames. We reject outlier estimates and compute a robust average over the inlier set to obtain an initial alignment. In the second stage, we refine this estimate using point-to-plane ICP~\cite{rusinkiewicz2001efficient} between the BLK2GO scan and the Aria semi-dense point cloud. All alignments are manually verified.

\vspace{-10pt}

\subsection{Data Annotation}
\label{sec:dataset_statistics}

% We make available the raw and processed data from all sensors.
% The processed outputs from the Aria recordings include eye gaze estimation, hand pose estimation, and camera trajectory data. 
% For each recording, we provide a high-fidelity RGB point cloud of the environment from the BLK2GO Scanner, which is aligned to the coordinate frame of the egocentric hand and camera pose data.
% To facilitate future applications involving object pose reconstruction, we also provide scans of the objects used.
% Additionally, we extract audio transcripts of the narrations and dialogue \cite{radford2023robust}. 
% All modalities available in our dataset are visualized in \autoref{fig:modalities}. 

In addition to the raw captured data, we manually annotate our recordings to provide detailed ground-truth for the three tasks defined above.
We use a mix of in-house and external professional human annotators. 
All annotators follow instructions and tutorial videos designed to ensure consistency and resolve ambiguity. 
External annotators pass their work through QA teams, and all annotations are further reviewed by at least one main contributor of our work. 
Annotations are performed using custom interfaces, built upon the VIA Video Annotator \cite{dutta2019vgg}, designed for each of the tasks. 
See Supp.~Mat.~for more details.

\subsection{Data Statistics}
\label{sec:statistics}
In total, our dataset contains 80 recordings from 55 kitchen spaces with 125 distinct participants. 
We accumulate 81h 26m of dual-view, or 40h 43m of single-view data. 
We provide a statistical analysis of our dataset, showing its diversity and size in terms of participants, recordings as well as the annotation taxonomy created for our considered tasks.

\begin{figure}[t]
  \centering

  \begin{minipage}{0.48\linewidth}
    \centering
    \includegraphics[width=\linewidth]{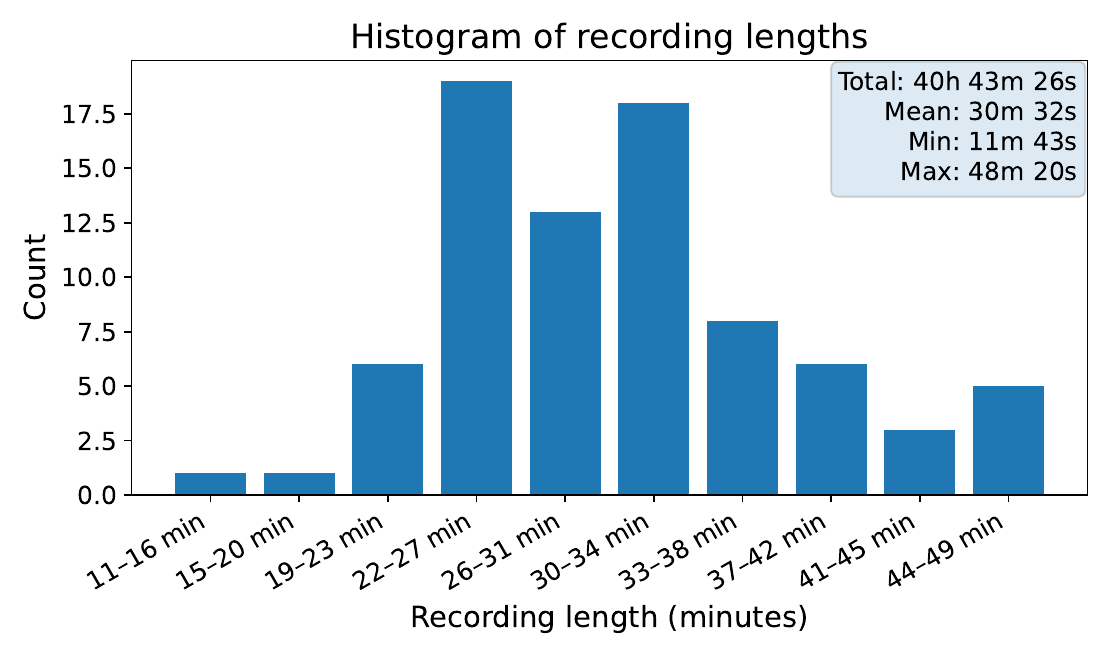}
    \caption{\textbf{Histogram of recording lengths.} Our lengthy recordings allow for the study of human-human collaborative work in long-context settings. In total, we record 41 hours of two-person resp.~81 hours of single-view data.}
    \label{fig:length_hist}
  \end{minipage}
  \hfill
  \begin{minipage}{0.48\linewidth}
    \centering
    \includegraphics[width=\linewidth]{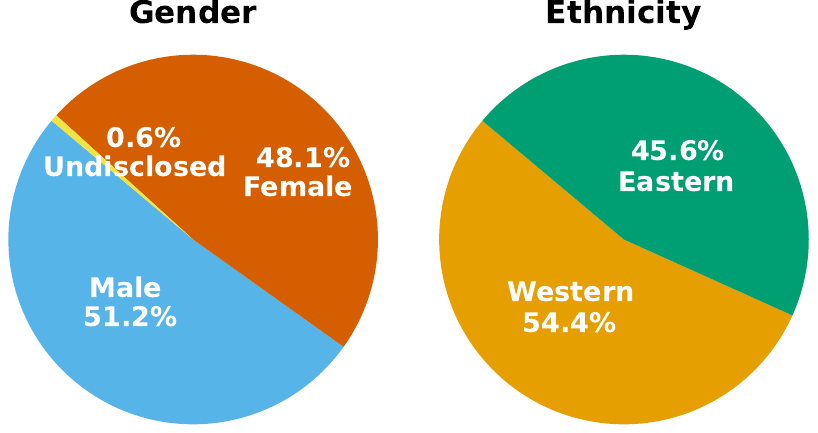}
    \caption{\textbf{Participant gender and ethnicity distribution.} We balance participants from Eastern and Western backgrounds, and maintain an equal gender distribution of our participants.}
    \label{fig:gender_hist}
    
  \end{minipage}
\vspace{-16pt}
\end{figure}

\noindent\textbf{Participants.} 
Our dataset features 125 unique participants organized into 80 collaborative pairs. 
% For each recording session, participants voluntarily assigned themselves to either the leader or helper role. 
% Participants self-assign the ``leader'' and ``helper'' roles.
% Typically, the individual with more culinary experience assumes the leader role to drive the task. \todo{--> put in supp??}
The participants have varying degrees of cooking expertise, and represent diverse nationalities and ethnic backgrounds. 
% All sessions, including natural dialogue and task coordination, were conducted exclusively in English. --> put this in supp if anything
A visualization of the demographics is provided in \autoref{fig:gender_hist}. 
Our dataset exhibits an approximately equal distribution of 48.1\% female-identifying participants compared to 51.2\% male-identifying and 0.6\% participants with undisclosed gender, whereas other egocentric datasets such as Ego4D \cite{grauman2022ego4d} and EgoExo4D \cite{grauman2024ego} report merely 45\% resp.~37\% of participants identifying as female. 
Furthermore, we achieve a good balance of 54.4\% participants with Western and 45.6\% with Eastern ethnicity. 
The participants' age at the time of recording ranges from 18 to 38, with most of them in their early to mid-20s.

\noindent\textbf{Recordings.} As shown in \autoref{fig:length_hist}, our recordings are on average 30 minutes long, allowing the study of long-horizon social interactions %for a given task and environment 
and making our dataset suited for future long-horizon benchmarks such as Moment Queries and Long-Term Action Anticipation from Ego4D \cite{grauman2022ego4d}.

\begin{figure*}[htbp]
\vspace{5pt}
  \centering

  \includegraphics[width=\linewidth]{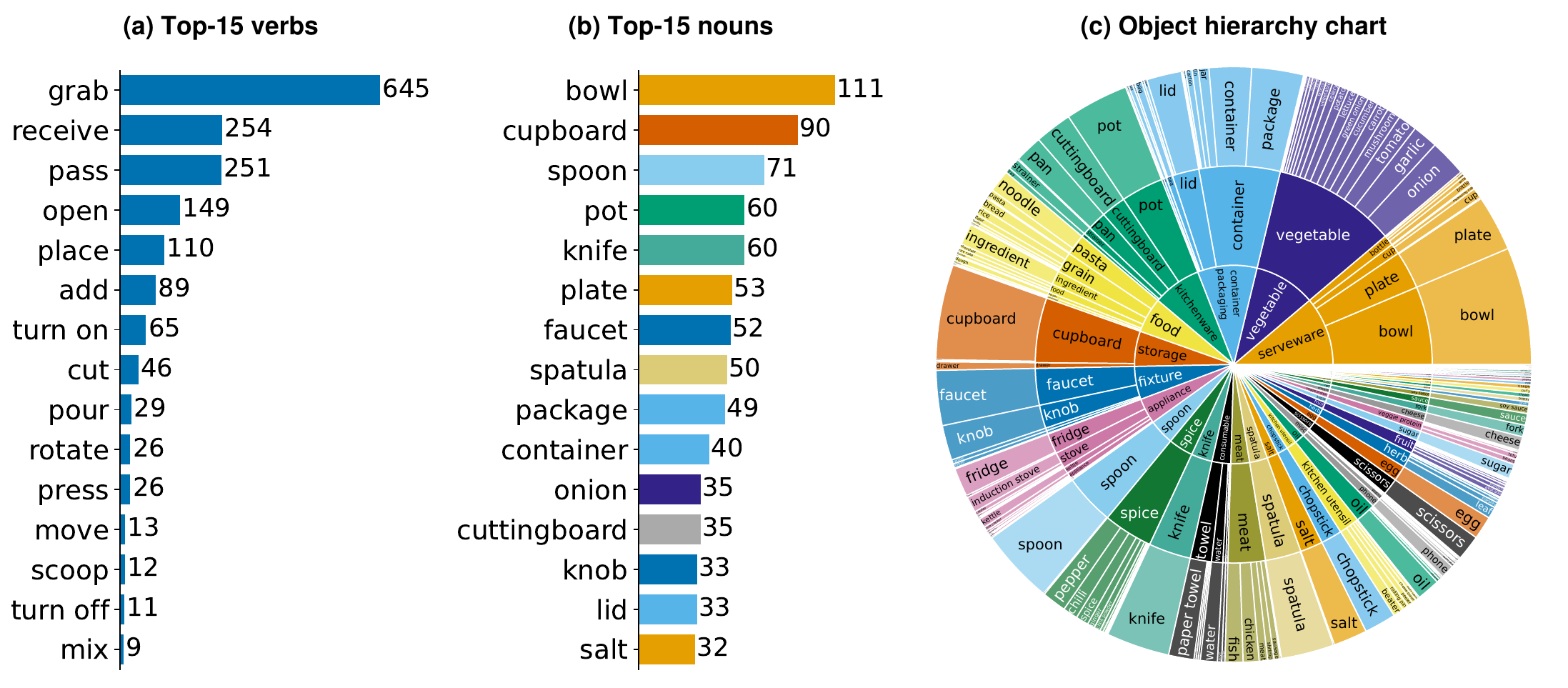}

  \caption{\textbf{Action verb and action noun distribution of the Socially Conditioned Object Interaction Anticipation task.} Our annotations feature a wide range of objects from 13 coarse (L3) and over 180 fine (L1) object categories, as well as 22 unique verbs.}
  \label{fig:scoia_object_hierarchy}
  \vspace{-16pt}
\end{figure*}

\noindent\textbf{Annotations.} For all tasks, we create a three-level hierarchy of object categories: L1 (finest) to L3 (coarsest), to account for different possible granularities when describing an object, e.g.~noodle (L1) vs.~pasta (L2) vs.~food (L3). We identify 281, 76 and 14 object categories on levels L1, L2 and L3, respectively. Details are provided in Supp.~Mat.

\begin{figure}[t]
  \vspace{-6pt}
  \centering

  \includegraphics[width=\linewidth]{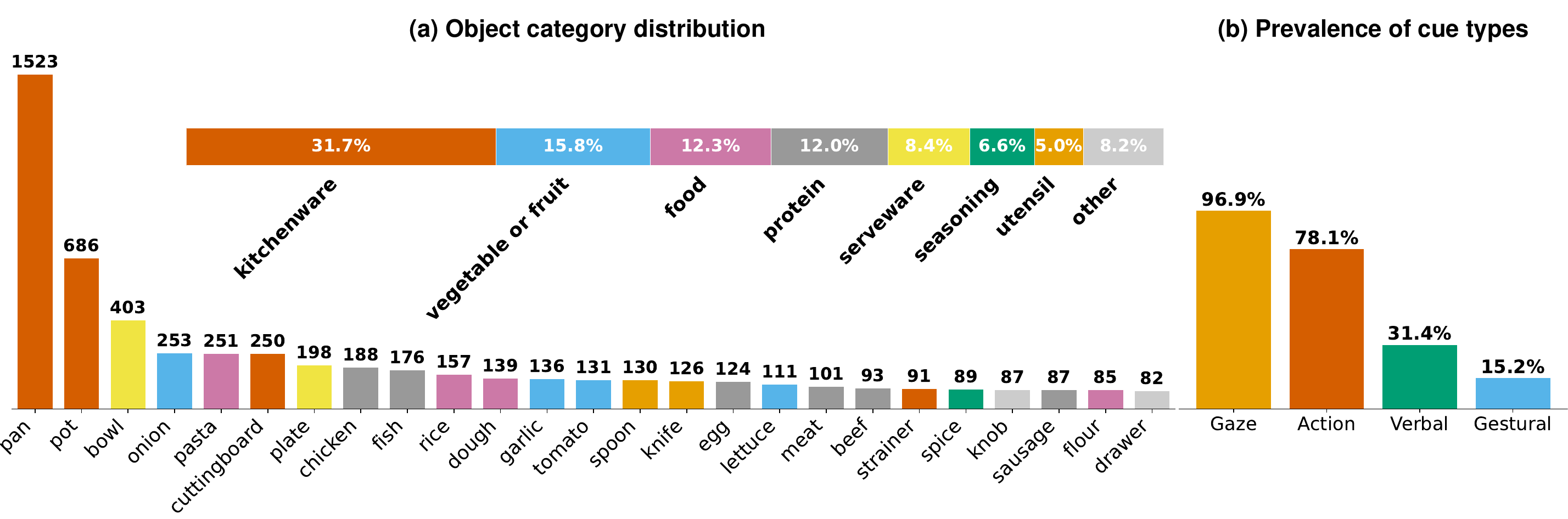}
  \vspace{-18pt}

  \caption{\textbf{Annotation statistics for the Joint Attention Estimation task.} The \name dataset allows investigating multiple indicators of joint attention between humans on a broad variety of objects.} % optional overall caption
  \label{fig:joint_attention_combined}
  \vspace{-12pt}
\end{figure}

\begin{figure}[t]
  \centering
  \includegraphics[width=\linewidth]{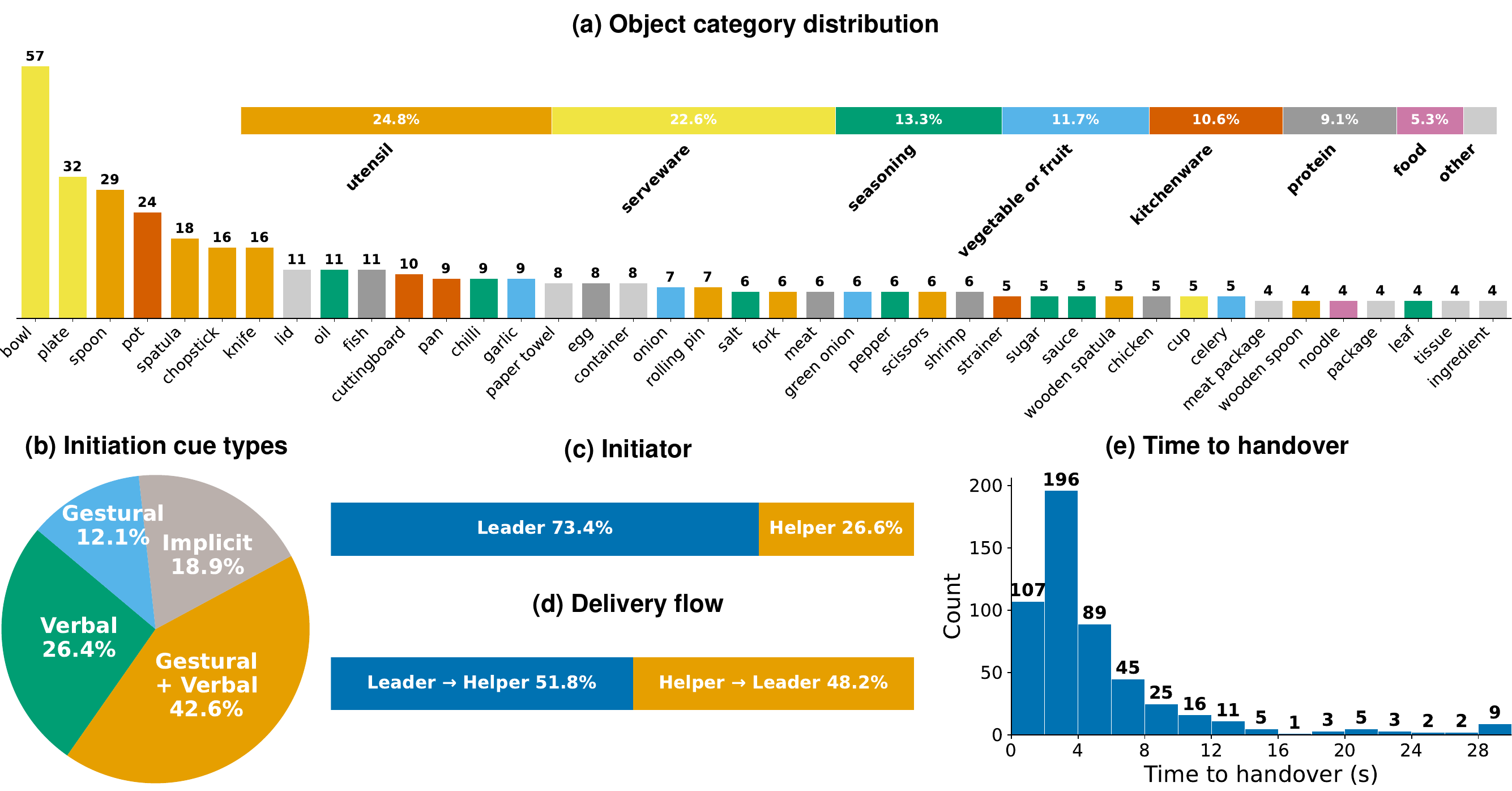}
\caption{\textbf{Annotation statistics for the Collaborative Handover Prediction task.} In addition to the category and bounding box of the handed-over object, we provide annotations about the Time-To-Handover (TTH), delivery flow direction, initiator and initiation cue for each handover event.}
  \label{fig:handover-stats}
  \vspace{-12pt}
\end{figure}

For the Socially Conditioned Object Interaction Anticipation task, we visualize this hierarchy, together with the top 15 most frequent verbs and nouns, in \autoref{fig:scoia_object_hierarchy}. We further provide statistics regarding the object categories for the Joint Attention Detection task in \autoref{fig:joint_attention_combined}. Statistics related to the Collaborative Handover Prediction task are illustrated in \autoref{fig:handover-stats}.

% This hierarchy is visualized in \autoref{fig:scoia_object_hierarchy}, with the inner circle corresponding to the highest (L3) and the outer circle corresponding to the lowest (L1) level. 

%\input{figures/noun_histogram}

\section{Baselines and Experiments}
\label{sec:experiments}

We benchmark numerous state-of-the-art closed- and open-source vision-language models (VLMs) on our three tasks, establishing a baseline for future research on human-human collaboration. Specifically, we benchmark Anthropic's Claude Opus (v4.5, v4.6) \cite{anthropic_claude_opus_4_5_system_card_2025,anthropic_claude_opus_4_6_announcement_2026}, Google's Gemini 3 Flash \cite{doshi2025gemini3flash}, OpenAI's GPT-4o \cite{openai_gpt_4o_system_card_2024} and GPT-5 families (v5.1, v5.2) \cite{openai_gpt_5_1_system_card_addendum_2025,openai_introducing_gpt_5_2_2025}, and the open-weight Gemma 4 31B IT \cite{gemma4} and Qwen3-VL series (8B, 32B) \cite{bai2025qwen3}. All models receive task-specific prompts instructing them to output predictions as structured JSON objects for standardized parsing. We partition the dataset into training and testing splits with disjunct participant sets, selecting 25 videos to provide test set annotations for all tasks. For visual processing, 5 video frames are uniformly sampled from the 10 seconds preceding the prediction frame, preserving fine-grained social cues while respecting VLM context limits. Transcribed speech of the last 10 seconds is also provided.
We further finetune our open-weight Qwen3-VL baselines (denoted with ``FT'' in the Tables~\ref{tab:JAEres},~\ref{tab:SCOIAres},~\ref{tab:CHPres}) and show substantial improvements, validating the effectiveness of our dataset for training socially aware models. Training details, more comparisons as well as visual results are provided in Supp.~Mat.
Lastly, we add a Random Sampling (RS) baseline which samples values according to the training split's distribution, and a Most Frequent (MF) baseline which picks the most frequent values, reflecting natural label priors. RS predicts bounding boxes according to the distribution's mean center point, width and height. MF uses the mean width and height and the image's center as the center point of the boxes.
 
% \todo{add gemini if we do use it; make clear which are open/close source models}
% \xw{what's the point of including different versions of the same model, i.e. claude 4.5 and 4.6, gpt 5.1 and 5.2?}
%\begin{itemize}
%    \item \textbf{.} We benchmark .
%    \item \textbf{GPT 4o.}
%    \item \textbf{GPT 5.} We benchmark versions 5.1 and 5.2.
%    \item \textbf{Qwen3-VL.} A state-of-the-art open-source vision-language model. We benchmark the 8B and 32B variants.  % 
%\end{itemize}
%\todo{Here we describe our experimental setup, metrics, baselines, and results for EACH task.} 

\subsection{Joint Attention Estimation}

\begin{table}[t]
\centering
% \vspace{-8pt}
\caption{\textbf{Quantitative results on Joint Attention Estimation.} \bestcap{Best}, \secondcap{second-best}, and \thirdcap{third-best} results are highlighted. $\mathrm{L\text{-}IoU@0.5}$ and $\mathrm{R\text{-}IoU@0.5}$ denote the fraction of predicted bounding boxes with $\mathrm{IoU} > 0.5$ with the ground-truth jointly attended object in the left and right egocentric streams, respectively. 
Cue Type reports cue classification accuracy, and Cat. L1 measures the fraction of correct L1 object category predictions. 
Models marked FT are fine-tuned on \name.}
\resizebox{\linewidth}{!}{

\label{tab:JAEres}
\begin{tabular}{l c c c c}
\toprule
Model 
& L-IoU@0.5~$\uparrow$ 
& R-IoU@0.5~$\uparrow$ 
& Cue Type~$\uparrow$ 
& Cat.~(L1)~$\uparrow$ \\
\midrule
Random Sampling & 0.1772 & 0.1337 & 0.3163 & 0.0533  \\
Most Frequent & 0.0004 & 0.0004 & 0.4647 & 0.2067 \\
\cdashline{1-5}
Claude Opus 4.5 & 0.0015 & 0.0025 & 0.2419 & 0.3433 \\
Claude Opus 4.6 & 0.0051 & 0.0015 & 0.3439 & 0.3487 \\
Gemini 3 Flash & \best{0.2164} & \best{0.2071} & {0.3685} & \third{0.4206} \\
GPT 4o & 0.0026 & 0.0003 & 0.0870 & 0.3451 \\
GPT 5.1 & 0.0042 & 0.0017 & 0.1133 & \second{0.4281} \\
GPT 5.2 & 0.0044 & 0.0017 & 0.1474 & \best{0.4285} \\
\cdashline{1-5}
Gemma 4 31B IT & {0.1295} & {0.1174} & \second{0.4628} & 0.3590 \\
Qwen3-VL 8B Instruct & 0.0000 & 0.0012 & 0.0074 & 0.2701 \\
Qwen3-VL 8B Instruct FT & \third{0.1483} & \third{0.1345} & \best{0.4775} & 0.3007 \\
Qwen3-VL 32B Instruct & 0.0001 & 0.0012 & 0.0740 & 0.3494 \\
Qwen3-VL 32B Instruct FT & \second{0.1562} & \second{0.1496} & \third{0.4572} & 0.3323 \\
%Qwen3-VL 32B Instruct FT & 0.0001 & 0.0016 & 0.1295 & 0.0000 \\
\bottomrule
\end{tabular}
}
\vspace{-5pt}
\end{table}

%\myparagraph{Baselines} Qwen3-VL-8B-Thinking, Gemini 3 Pro, InternVideo-Next, GLC transformer

\myparagraph{Metrics} We separately report the fraction of IoUs over 0.5 for the objects in the view of the left (L-IoU@0.5) as well as the right (R-IoU@0.5) person. We also provide the fraction of prediction-GT matches for the cue type. Lastly, we measure the matches of the object category with the ground-truth L1 category. Object categories are organized into three levels of granularity: L1 represents the most detailed set of categories, while L2 and L3 progressively merge them into more general groups, with L3 providing the coarsest categorization. %, as in the Handover task.

\myparagraph{Results} As shown in~\autoref{tab:JAEres}, zero-shot spatial localization for joint attention remains exceptionally challenging. Most proprietary VLMs (e.g., Claude, GPT families) fail almost entirely at bounding box generation, scoring near zero. Gemini 3 Flash is the notable exception, achieving the highest performance in spatial grounding (0.2164 resp.~0.2071 for L-IoU and R-IoU). Importantly, fine-tuning the Qwen3-VL models on \name yields substantial improvements. The fine-tuned 32B variant jumps from near-zero spatial scores to the second-best spatial performance on both L-IoU and R-IoU, while the finetuned 8B variant secures the highest accuracy for identifying the social Cue Type (0.4775). Finally, the clustered Category scores (0.27 to 0.43) across all models suggest that while current VLMs can semantically infer the target object from context, they severely lack the embodied visual grounding required to localize it across synchronous egocentric views.

\subsection{Socially Conditioned Object Interaction Anticipation}

\begin{table}[t]
\centering
\caption{\textbf{Quantitative results on Socially Conditioned Object Interaction Anticipation.} \bestcap{Best}, \secondcap{second-best}, and \thirdcap{third-best} results are highlighted. $\mathrm{IoU@0.5}$ denotes the fraction of predicted bounding boxes with $\mathrm{IoU} > 0.5$ with the ground-truth object to be interacted with. 
Cue Type reports cue classification accuracy, and Act. Verb resp.~Act. Noun (L1) measures the fraction of correct verb resp.~L1 object category predictions. 
Models marked FT are fine-tuned on \name.}
\resizebox{\linewidth}{!}{

\label{tab:SCOIAres}
\begin{tabular}{l c c c c}
\toprule
Model 
& IoU@0.5~$\uparrow$ 
& Cue Type~$\uparrow$ 
& Act.~Verb~$\uparrow$ 
& Act.~Noun (L1)~$\uparrow$ \\
\midrule
Random Sampling & 0.0152 & 0.3801 & 0.1854 & 0.0411 \\
Most Frequent & 0.0005 & 0.4040 & 0.3430 & 0.0662 \\
\cdashline{1-5}
Claude Opus 4.5 & 0.0034 & \third{0.5302} & 0.0819 & 0.3490 \\
Claude Opus 4.6 & 0.0091 & \second{0.5347} & 0.0987 & 0.3507 \\
Gemini 3 Flash & \third{0.1859} & {0.5298} & 0.0781 & 0.3775 \\
GPT 4o & 0.0009 & 0.4358 & 0.1046 & 0.3351 \\
GPT 5.1 & {0.0202} & 0.4556 & {0.1325} & \third{0.4119} \\
GPT 5.2 & 0.0038 & 0.5113 & 0.1060 & \best{0.4371} \\
\cdashline{1-5}
Gemma 4 31B IT & 0.1468 & 0.4173 & 0.0679 & 0.3225 \\
Qwen3-VL 8B Instruct & 0.0909 & 0.1151 & 0.0873 & 0.2288 \\
% Qwen3-VL 8B Instruct FT & 0.0455 & 0.4609 & 0.4556 & 0.3642 \\  % v6
Qwen3-VL 8B Instruct FT & \second{0.2425} & 0.4299 & \second{0.6429} & 0.4021 \\ % v5
Qwen3-VL 32B Instruct & 0.0035 & 0.4901 & \third{0.1380} & 0.3509 \\
Qwen3-VL 32B Instruct FT & \best{0.2513} & \best{0.5503} & \best{0.6508} & \second{0.4312} \\  % v7
% Qwen3-VL 32B Instruct FT & 0.0006 & 0.5060 & 0.1391 & 0.3550 \\
\bottomrule
\end{tabular}
}
\vspace{-8pt}
\end{table}

%\myparagraph{Baselines} Ego4Dv2 baseline, StillFast, Transfusion, LLaVa-Grounding.

\myparagraph{Metrics} IoU@0.5 and Cue Type are calculated as in the Joint Attention Detection task. We further report the fraction of matches with the ground-truth verb and L1 noun. As in the previous task, object categories follow the three-level hierarchy (L1–L3), with evaluation performed at the most fine-grained level (L1).

\myparagraph{Results} The results in~\autoref{tab:SCOIAres} indicate that predicting future interactions is significantly more complex than identifying current attention. Among the proprietary VLMs, Gemini 3 Flash maintains its lead in spatial grounding (0.1859 IoU), while Claude Opus 4.6 and GPT 5.2 excel in Cue Type and Noun categorization, respectively. Notably, fine-tuning Qwen3-VL on \name allows the 32B variant to take the lead in spatial grounding with 0.2513 IoU, followed by the 8B variant scoring second at 0.2425. The finetuned 32B model further scores highest among all models on the Cue Type and Act.\ Verb accuracy with 0.5503 and 0.6508, respectively. Notably, fine-tuning the Qwen3-VL models on \name results in a seven-fold and five-fold increase in their Action Verb accuracy, from 0.0873 and 0.1380 to 0.6429 and 0.6508. This substantial jump suggests that while zero-shot VLMs can recognize static objects and high-level social signals, domain-specific training is essential to accurately model the temporal action dependencies triggered by collaborative social cues.

\subsection{Collaborative Handover Prediction}

\begin{table}[t]
\centering
\vspace{-8pt}
\caption{\textbf{Quantitative results on Collaborative Handover Prediction.} \bestcap{Best}, \secondcap{second-best}, and \thirdcap{third-best} results are highlighted. $\mathrm{IoU@0.5}$ denotes the fraction of predicted bounding boxes with $\mathrm{IoU} > 0.5$ with the ground-truth object. 
Del. Flow, Initiator, and Init. Type report the fraction of correct predictions for the delivery flow, the initiating participant, and the initiation cue type, respectively. 
Cat. (L1) measures the fraction of correct L1 object category predictions. 
TTH reports the fraction of time-to-handover predictions within $0.25\,\mathrm{s}$ of the ground-truth. 
Models marked FT are fine-tuned on \name.}
\resizebox{\linewidth}{!}{

\label{tab:CHPres}
\begin{tabular}{l c c c c c c}
\toprule
Model & IoU@0.5~$\uparrow$ 
& Del.~Flow~$\uparrow$ 
& Initiator~$\uparrow$ 
& Init.~Type~$\uparrow$ 
& Cat.~(L1)~$\uparrow$ 
& TTH~$\uparrow$ \\
\midrule
Random Sampling & 0.0000 & 0.4476 & 0.6810 & 0.2905 & 0.0429 & 0.0714 \\
Most Frequent & 0.0000 & 0.4857 & 0.7857 & 0.4667 & 0.1048 & 0.0524  \\
\cdashline{1-7}
Claude Opus 4.5 & 0.0000 & 0.5659 & 0.6195 & \third{0.3073} & {0.3610} & \third{0.0976} \\ 
Claude Opus 4.6  & \third{0.0025} & \second{0.6146} & \second{0.6488} & 0.2976 & {0.3610} & \best{0.1317} \\
Gemini 3 Flash & \best{0.0313} & \best{0.6524} & 0.5810 & \second{0.3476} & 0.3524 & 0.0667 \\
GPT 4o & {0.0008} & 0.5433 & 0.5240 & 0.2644 & 0.3317 & {0.0962} \\
GPT 5.1 & 0.0000 & 0.5857 & 0.5952 & 0.2333 & \best{0.4000} & 0.0810 \\
GPT 5.2 & 0.0000 & {0.5952} & 0.5667 & 0.2905 & \second{0.3714} & 0.0810 \\
\cdashline{1-7}
Gemma 4 31B IT & \second{0.0130} & 0.5050 & 0.4100 & 0.2800 & 0.3400 & 0.0750 \\
Qwen3-VL 8B Instruct & 0.0000 & 0.5381 & 0.4667 & 0.1286 & 0.1905 & 0.0524 \\
% Qwen3-VL 8B Instruct FT & 0.0000 & 0.4762 & 0.6333 & 0.2810 & 0.3524 & 0.0333 \\ % v6
Qwen3-VL 8B Instruct FT & \third{0.0025} & 0.5714 & \third{0.6667} & {0.1429} & 0.1952 & \second{0.1143} \\ % v5
Qwen3-VL 32B Instruct & 0.0000 & 0.5381 & 0.6143 & 0.2952 & 0.3095 & 0.0381 \\
Qwen3-VL 32B Instruct FT & \second{0.0130} & \third{0.6143} & \best{0.7286} & \best{0.3714} & \third{0.3619} & 0.0905 \\ % v7
\bottomrule
\end{tabular}
}
\vspace{-16pt}
\end{table}

\myparagraph{Metrics} We report $\mathrm{IOU@0.5}$, the fraction of predicted bounding boxes with over $50\%$ IoU with the ground-truth bounding box. Moreover, we calculate the dataset-wide fraction of matches (accuracy) for the delivering flow, initiatior, and initiation type predictions. For the object category (Cat.), we measure the fraction of matches of the predicted category with the ground-truth L1 category, counting synonyms for the GT of the same granularity as correct predictions. Object categories follow the three-level hierarchy (L1–L3), with evaluation performed at the most fine-grained level (L1). Finally, we report the fraction of time-to-handover predictions falling within $0.25s$ of the ground-truth TTH.

\myparagraph{Results} As seen in \autoref{tab:CHPres}, handover prediction emerges as the most difficult task for spatial grounding, with Gemini 3 Flash leading at a mere 0.0313 IoU. While GPT-5.1 shows strength in semantic categorization (0.4000) and Claude 4.6 leads in temporal accuracy (0.1317 TTH), these proprietary models are noticeably less sensitive to social dynamics. In contrast, fine-tuning Qwen3-VL models on \name yields the best results for identifying the Initiator (0.7286) and Initiation Type (0.3714). This performance split suggests that while general VLMs can leverage broad heuristics for ``what'' and, to a lesser degree, for ``when,'' domain-specific training on \name is critical for teaching models the ``who'' and ``how'' of proactive collaboration. The across-the-board low scores highlight that seamless, human-like handover remains a challenging frontier for modern AI.

\section{Conclusion}
\label{sec:conclusion}

We introduced \name, a dataset of ego-/exocentric recordings collected for the study of human-human interaction in collaborative settings. 
The dataset features a large set of complementary modalities, involving both egocentric and environment data as well as static and dynamic data. 
We established benchmarks on three new tasks related to human collaboration, and manually collected rich and diverse ground-truth annotations for them in our dataset. 
We performed an evaluation of numerous vision-language models on our tasks, showing that the current state-of-the-art leaves much room for improvement. 
We further demonstrated the practical use of our dataset by finetuning an open-source model on its training set, leading to a boost in test-set performance.

% Now we have reached the maximum length of an ECCV \ECCVyear{} submission (excluding references and acknowledgements).
% References should start immediately after the main text, but can continue past p.\ 14 if needed. 
% \clearpage  % TODO FINAL: This \clearpage needs to be removed from both review and camera-ready versions.

\section*{Acknowledgements}
This work was supported by the Swiss National Science Foundation Advanced Grant 216260: Beyond Frozen Worlds: Capturing Functional 3D Digital Twins from the Real World and by European Union's Horizon Europe research and innovation programme under grant agreement number 101214398 (ELLIOT), fully funded by Swiss State Secretariat for Education, Research and Innovation (SERI). Alexandros Delitzas is also supported by the Max Planck ETH Center for Learning Systems (CLS).

\bibliographystyle{splncs04}
\bibliography{main}
\clearpage
\appendix
\section*{Supplementary Material}

\renewcommand{\thesection}{S\arabic{section}}  
\renewcommand{\thetable}{S\arabic{table}}  
\renewcommand{\thefigure}{S\arabic{figure}} 

%\renewcommand{\cftsecnumwidth}{20pt}
%\cftsetrmarg{50pt}

\setcounter{section}{0}
\setcounter{figure}{0}
\setcounter{equation}{0}

\makeatletter
\renewcommand{\tableofcontents}{%
  \section*{\contentsname}%
  \par\vspace{1em}%
  \@starttoc{toc}%
}
\makeatother

\section{Qualitative Examples}

\begin{figure}[ht]
    \centering
    
    \begin{subfigure}[t]{0.48\textwidth}
        \centering
        \includegraphics[width=\linewidth]{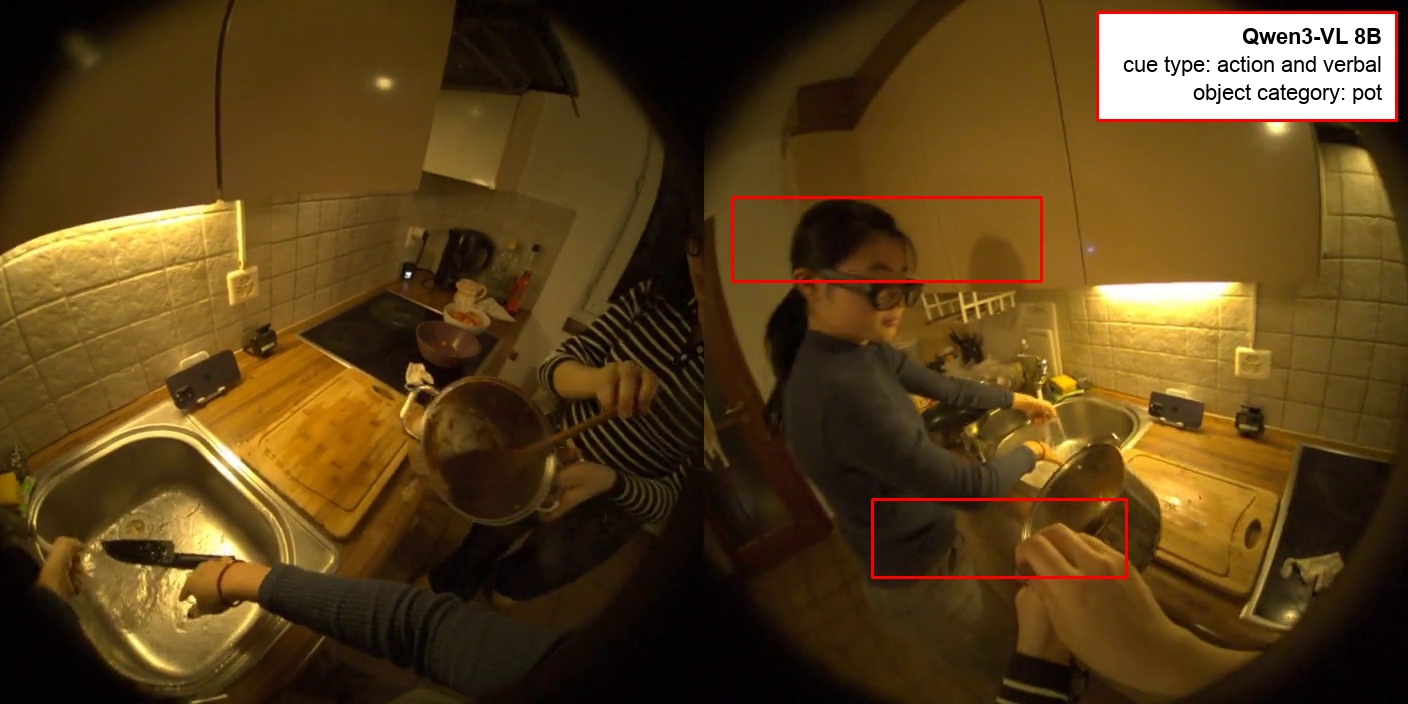}
        %\caption{Second}
    \end{subfigure}
    \hfill
    \begin{subfigure}[t]{0.48\textwidth}
        \centering
        \includegraphics[width=\linewidth]{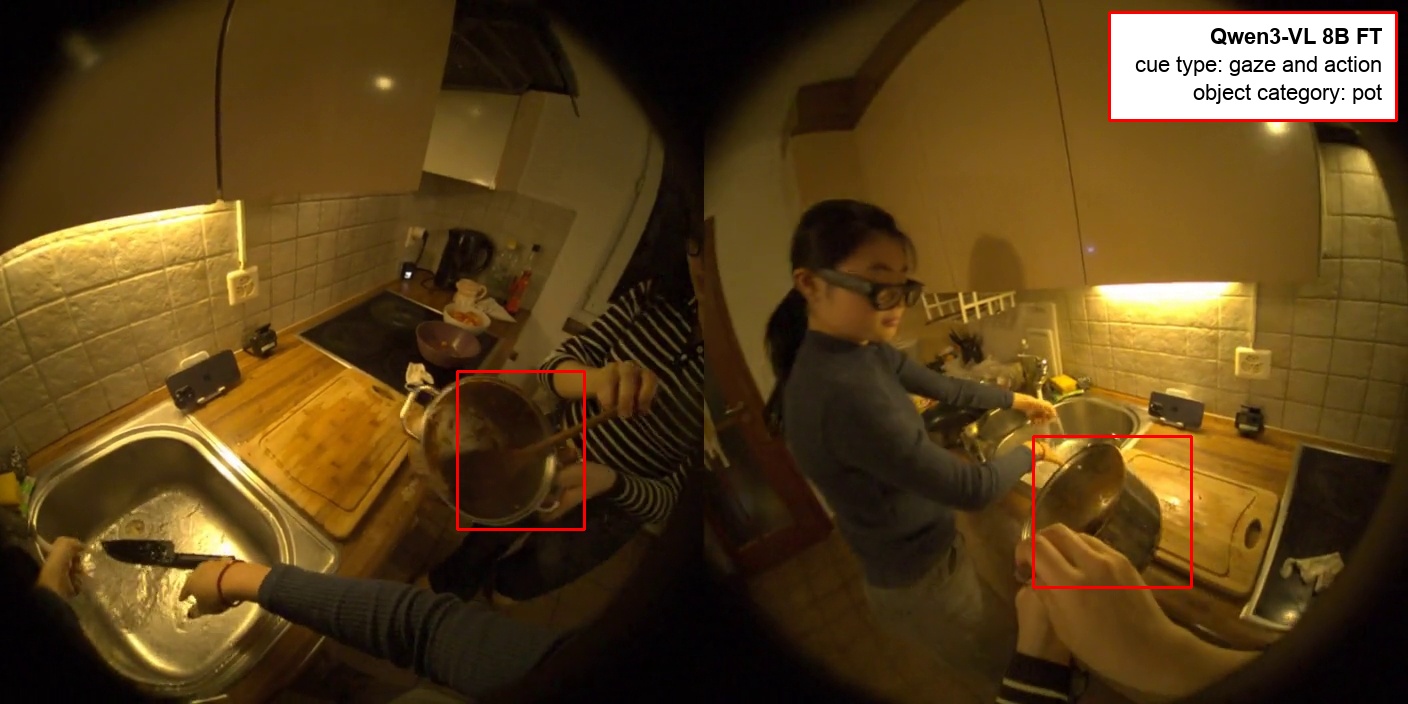}
        %\caption{Second}
    \end{subfigure}
    
    \vspace{0.5em}
    
    \begin{subfigure}[t]{0.48\textwidth}
        \centering
        \includegraphics[width=\linewidth]{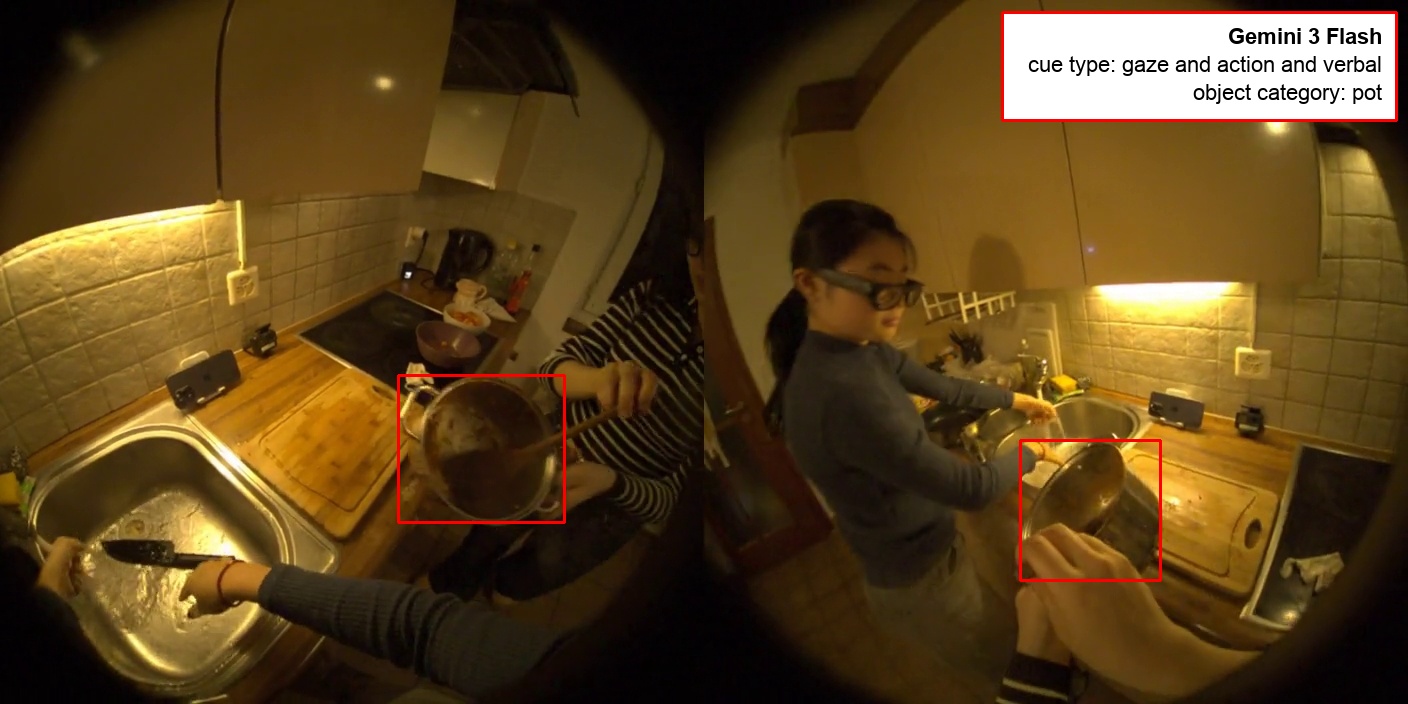}
        %\caption{Second}
    \end{subfigure}
    \hfill
    \begin{subfigure}[t]{0.48\textwidth}
        \centering
        \includegraphics[width=\linewidth]{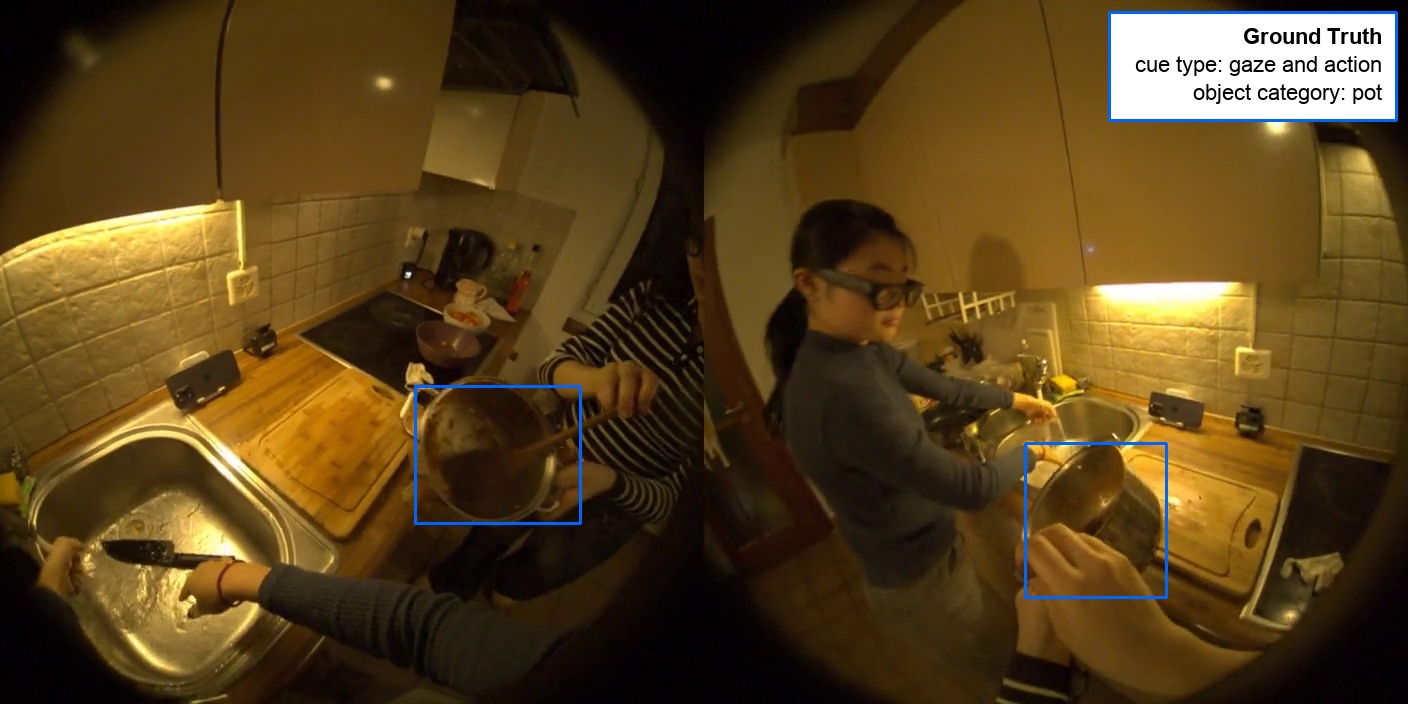}
        %\caption{Second}
    \end{subfigure}
    
    \caption{\textbf{Baseline comparison for the Joint Attention Estimation task.} We visualize and compare the predictions of Qwen3-VL 8B with those of Qwen3-VL 8B finetuned on the CoMind training set and Gemini 3 Flash, and provide the ground-truth solution given by our dataset. Finetuning on our training set improves the test set predictions.}
    \label{fig:supp_joint_attention_baseline_comp}
\end{figure}

\begin{figure}[ht]
    \centering
    \begin{subfigure}[t]{0.3\textwidth} % 0.48
        \centering
        \includegraphics[width=\linewidth]{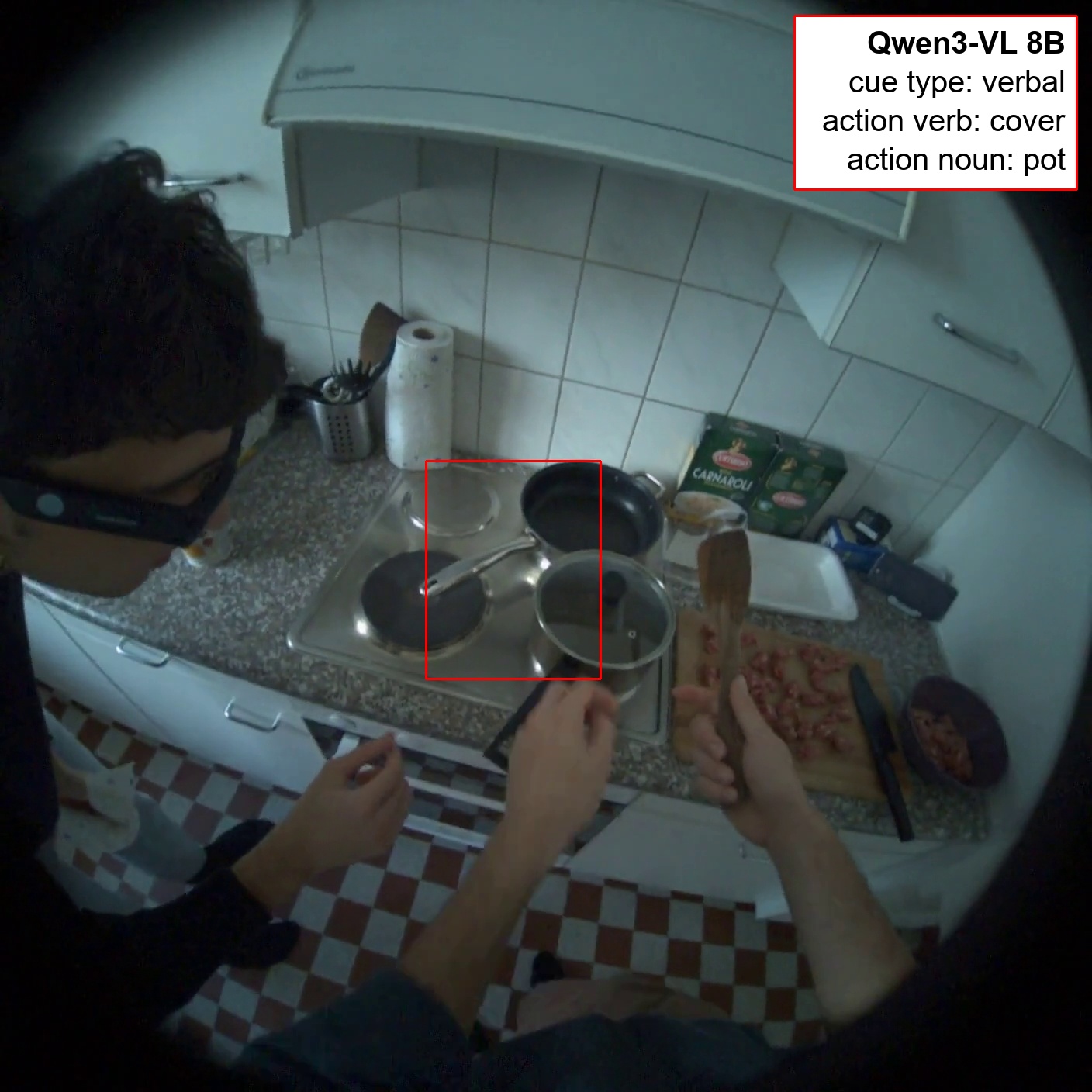}
        %\caption{Second}
    \end{subfigure}
    \begin{subfigure}[t]{0.3\textwidth}
        \centering
        \includegraphics[width=\linewidth]{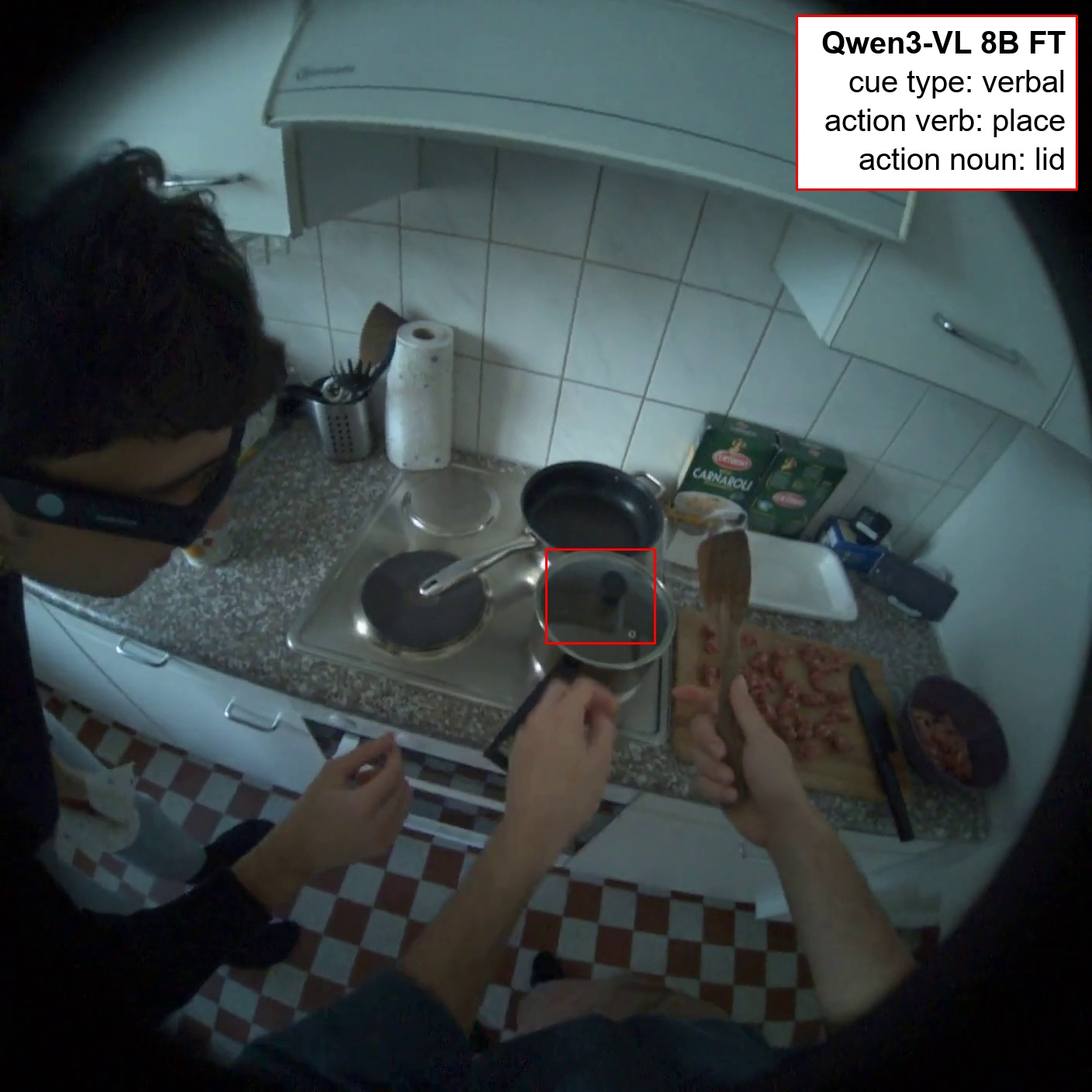}
        %\caption{Second}
    \end{subfigure}

    \vspace{0.5em}

    \begin{subfigure}[t]{0.3\textwidth}
        \centering
        \includegraphics[width=\linewidth]{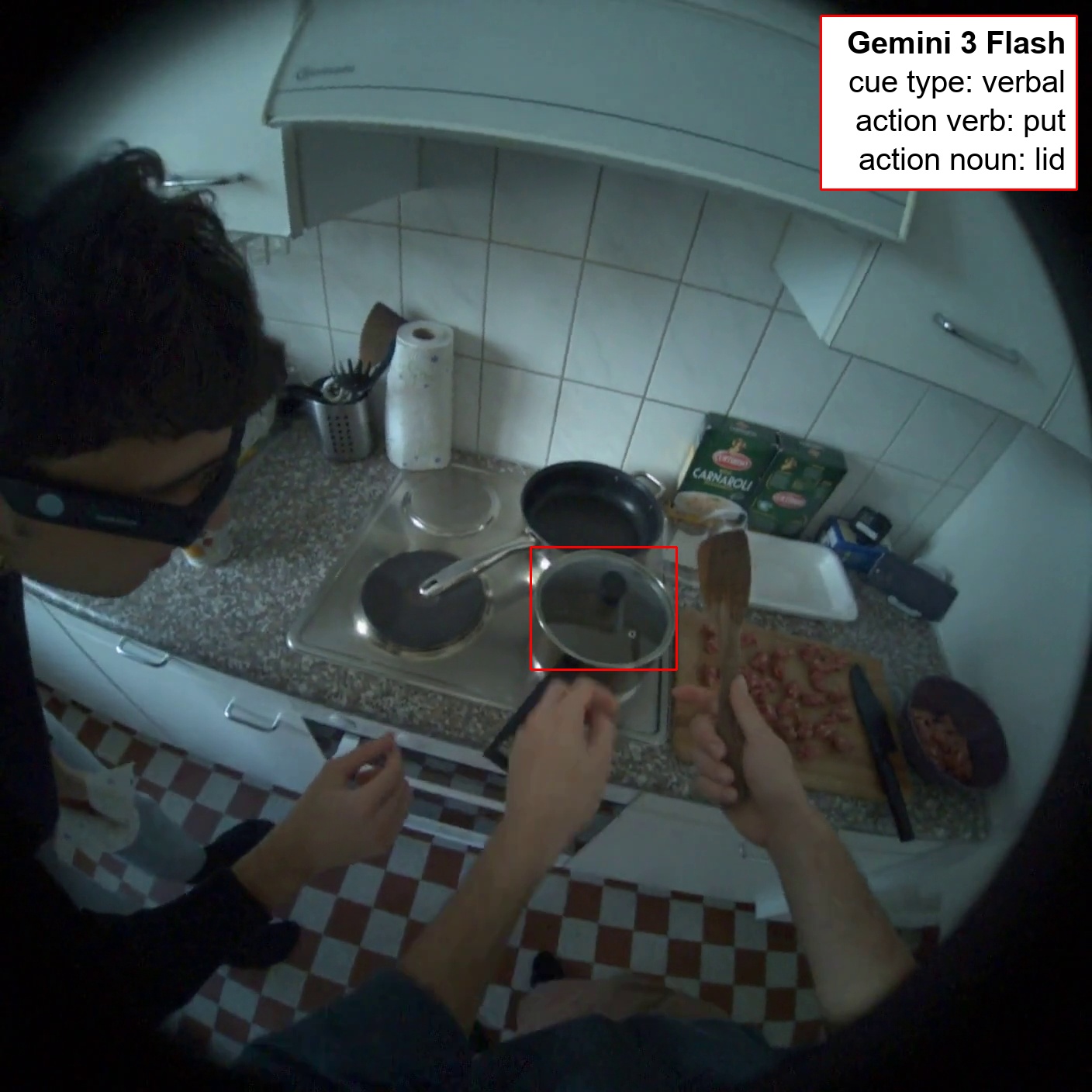}
        %\caption{Second}
    \end{subfigure}
    \begin{subfigure}[t]{0.3\textwidth}
        \centering
        \includegraphics[width=\linewidth]{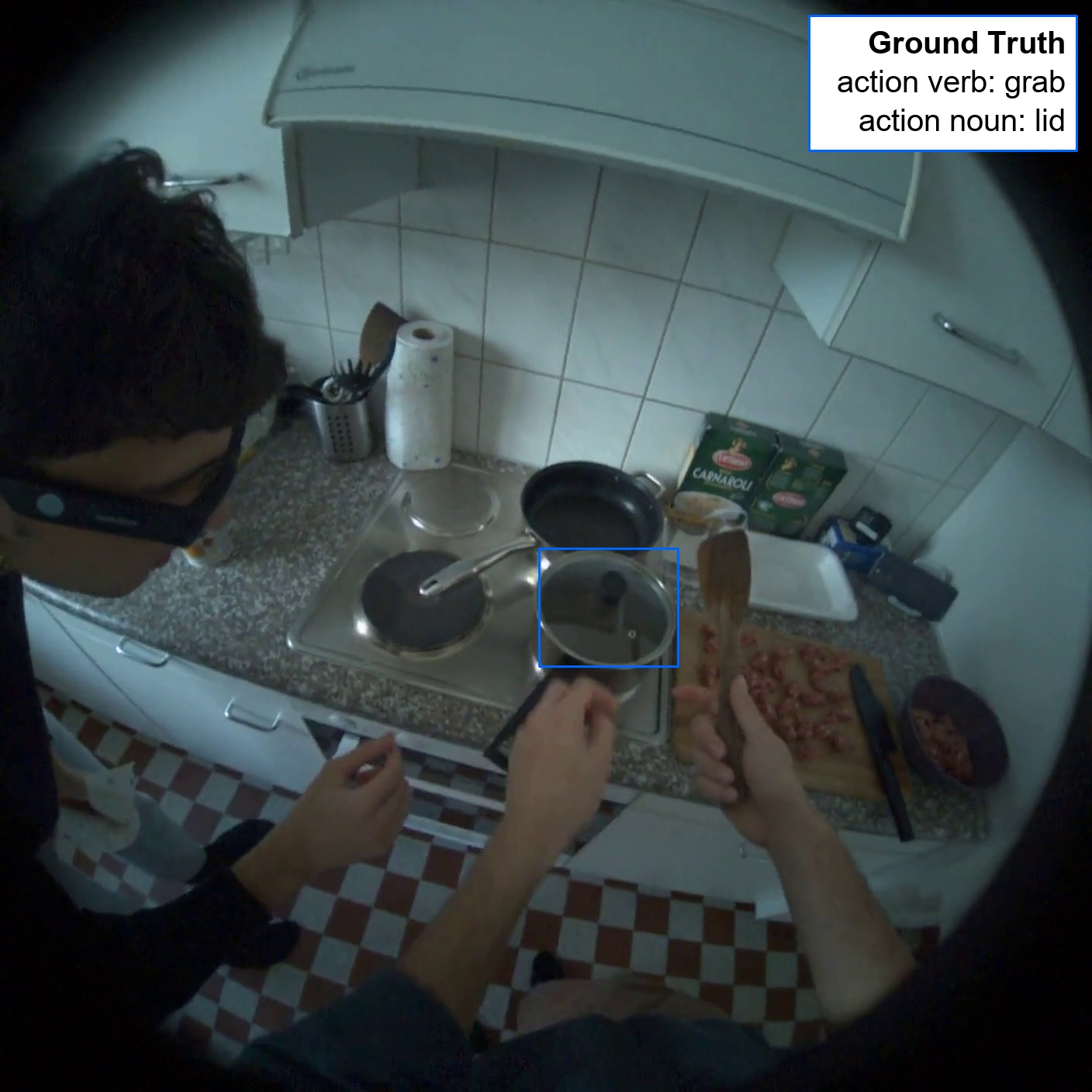}
        %\caption{Second}
    \end{subfigure}

    \vspace{1.5em}

    \begin{subfigure}[t]{0.3\textwidth} % 0.48
        \centering
        \includegraphics[width=\linewidth]{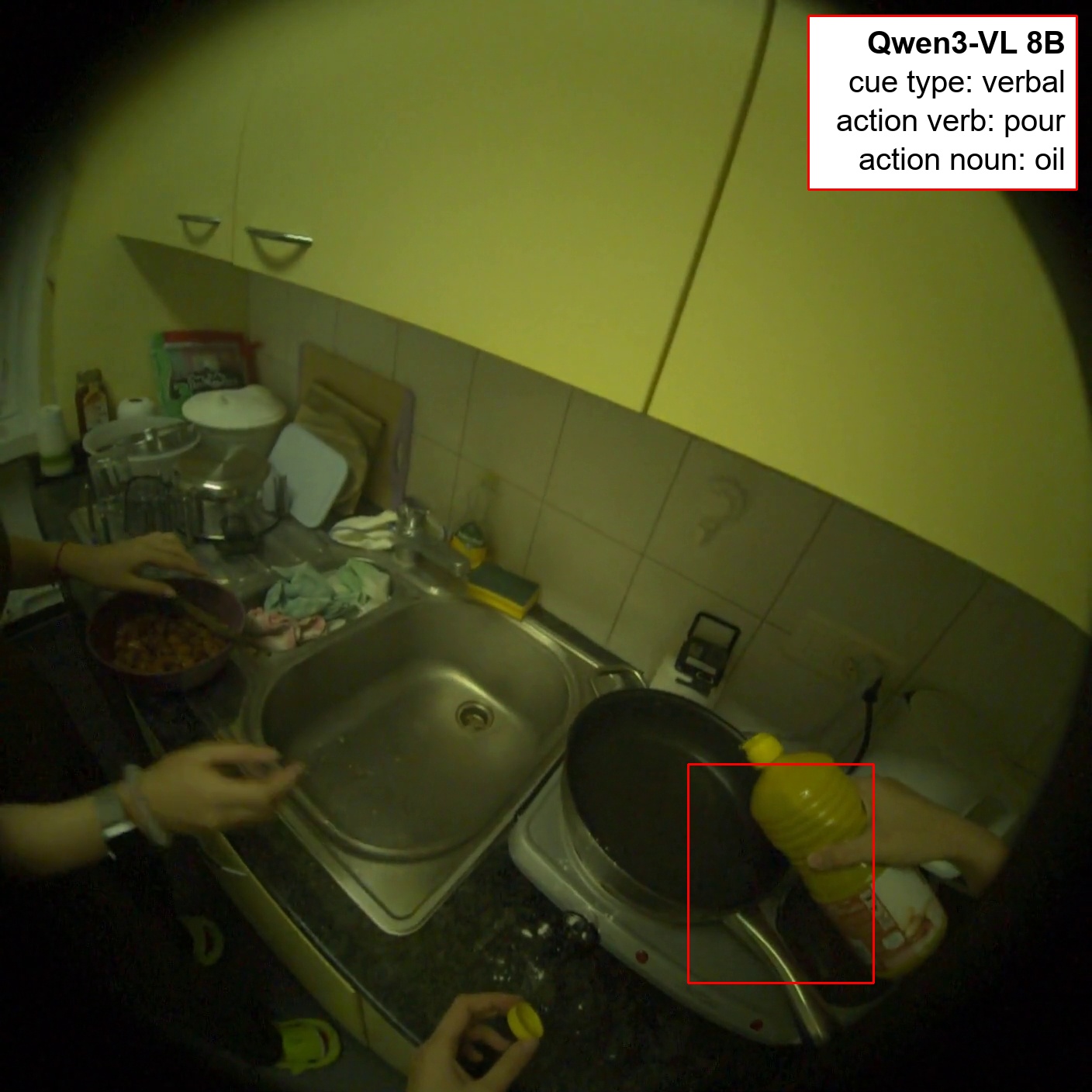}
        %\caption{Second}
    \end{subfigure}
    \begin{subfigure}[t]{0.3\textwidth}
        \centering
        \includegraphics[width=\linewidth]{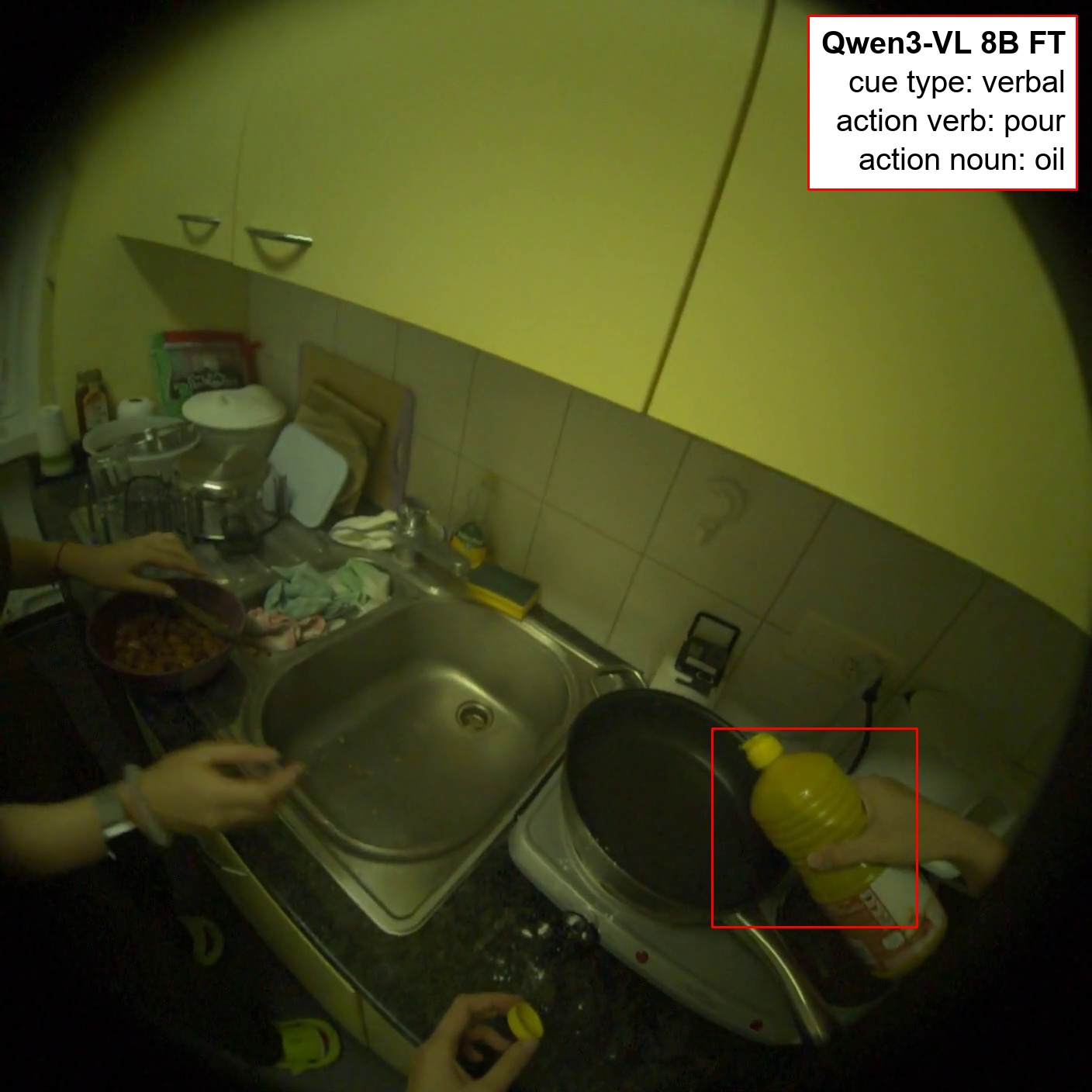}
        %\caption{Second}
    \end{subfigure}

    \vspace{0.5em}

    \begin{subfigure}[t]{0.3\textwidth}
        \centering
        \includegraphics[width=\linewidth]{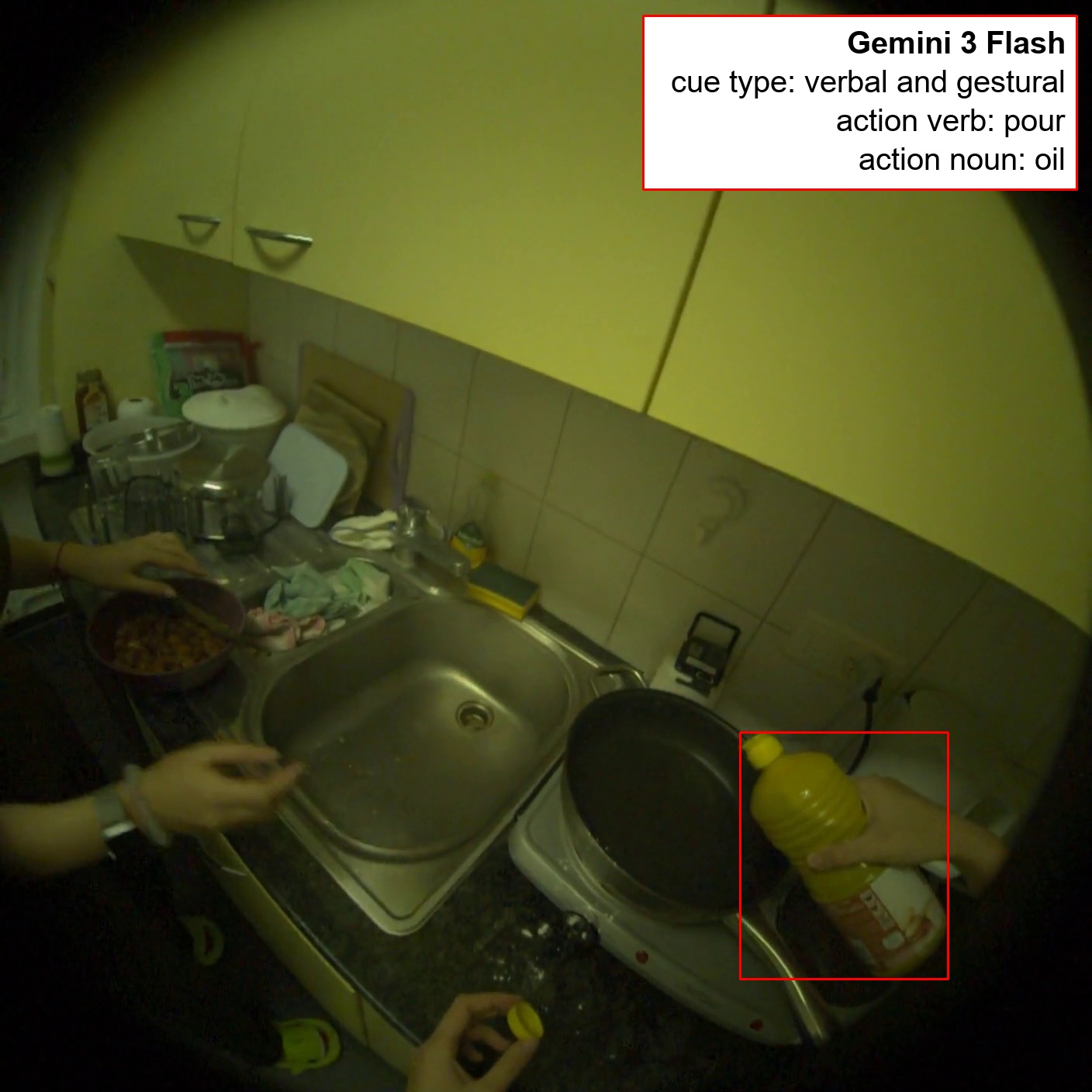}
        %\caption{Second}
    \end{subfigure}
    \begin{subfigure}[t]{0.3\textwidth}
        \centering
        \includegraphics[width=\linewidth]{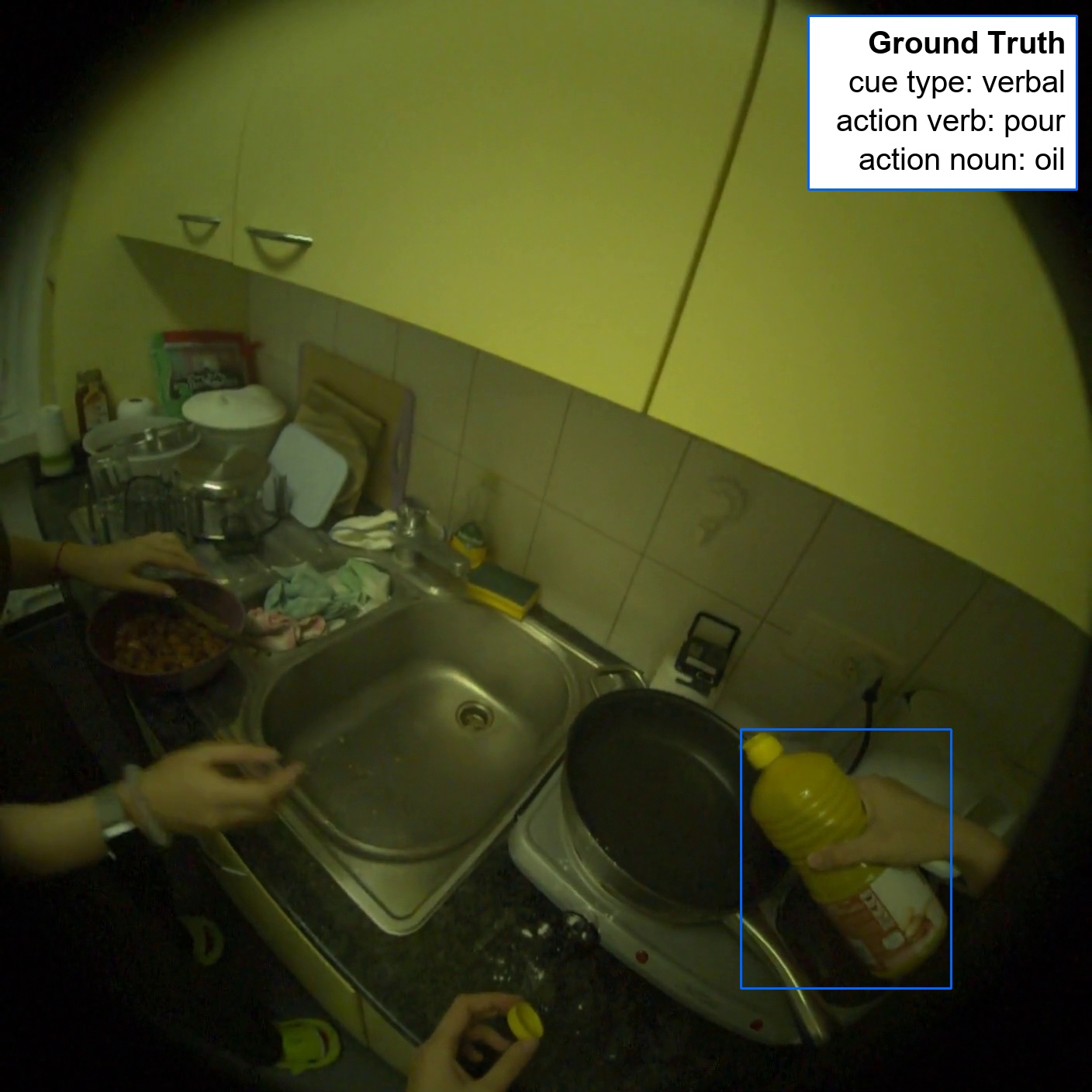}
        %\caption{Second}
    \end{subfigure}

    \caption{\textbf{Baseline comparison for the Socially Conditioned Object Interaction Anticipation task.} We visualize and compare the predictions of Qwen3-VL 8B with those of Qwen3-VL 8B finetuned on the CoMind training set and Gemini 3 Flash, and provide the ground-truth solution given by our dataset. Finetuning on our training set improves the test set predictions.}
    \label{fig:supp_scoia_baseline_comp}
\end{figure}

\begin{figure}[t]
    \centering
    
    \begin{subfigure}[t]{0.48\textwidth}
        \centering
        \includegraphics[width=\linewidth]{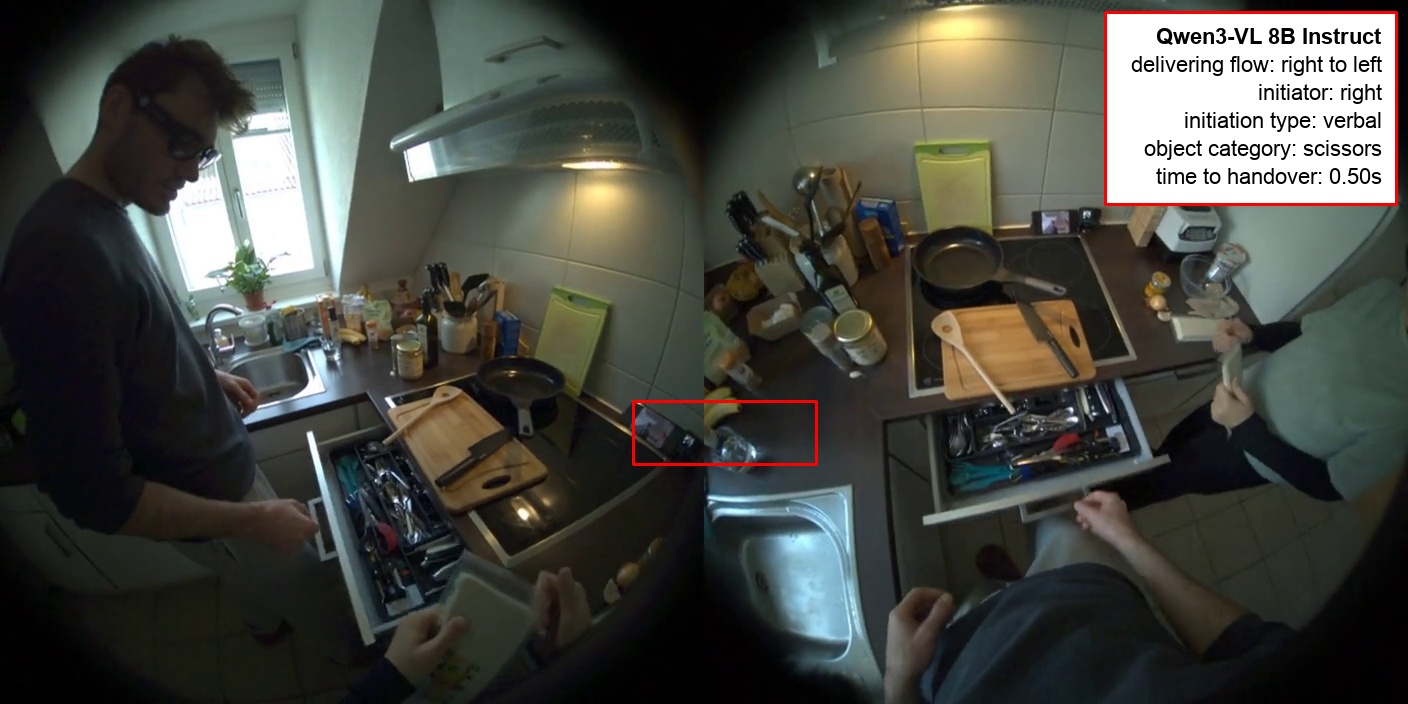}
        %\caption{Second}
    \end{subfigure}
    \hfill
    \begin{subfigure}[t]{0.48\textwidth}
        \centering
        \includegraphics[width=\linewidth]{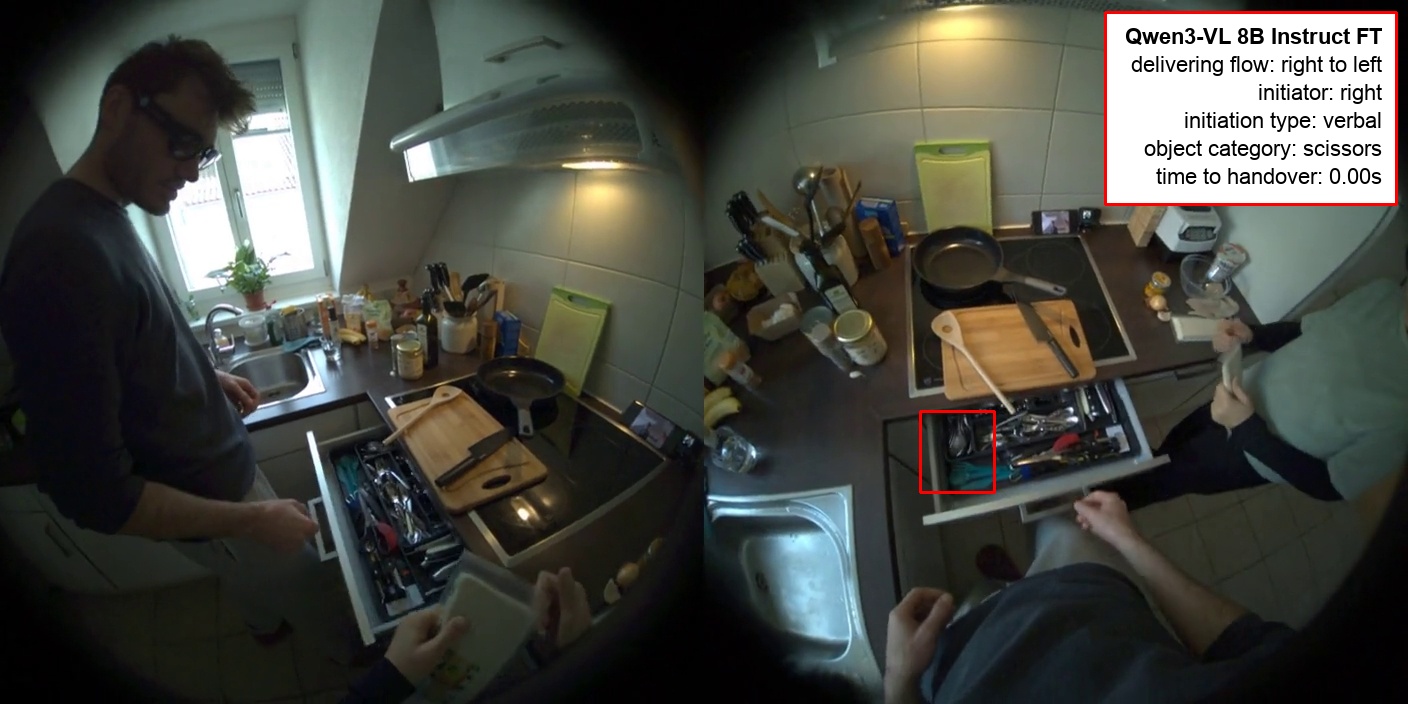}
        %\caption{Second}
    \end{subfigure}
    
    \vspace{0.5em}
    
    \begin{subfigure}[t]{0.48\textwidth}
        \centering
        \includegraphics[width=\linewidth]{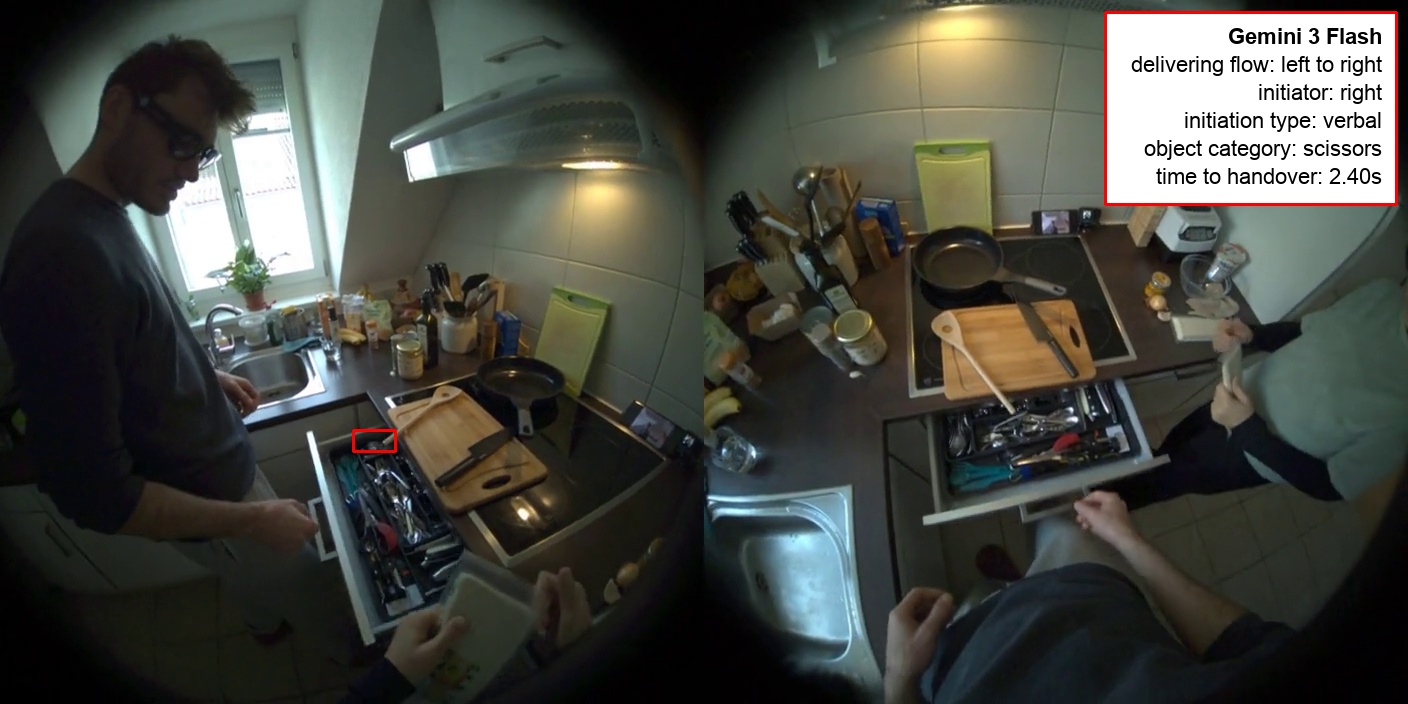}
        %\caption{Second}
    \end{subfigure}
    \hfill
    \begin{subfigure}[t]{0.48\textwidth}
        \centering
        \includegraphics[width=\linewidth]{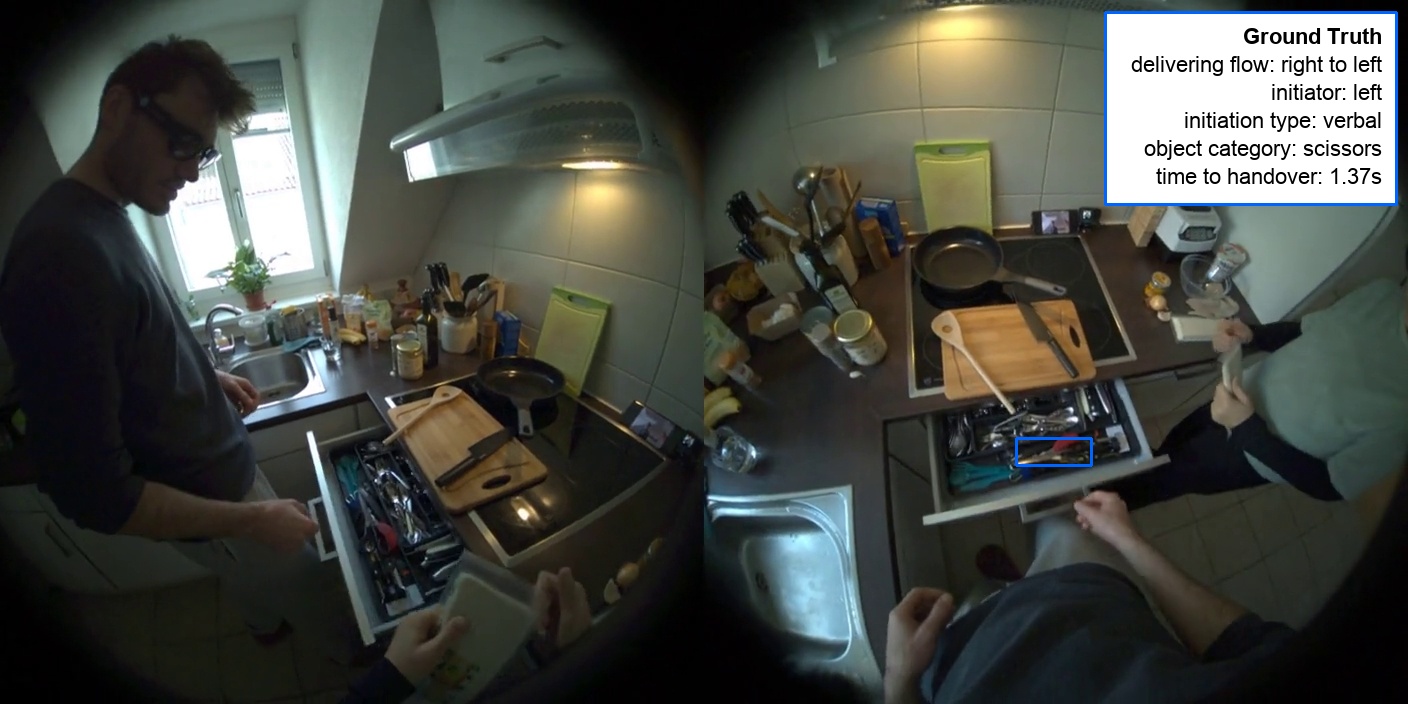}
        %\caption{Second}
    \end{subfigure}

    \caption{\textbf{Baseline comparison for the Collaborative Handover Prediction task.} We visualize and compare the predictions of Qwen3-VL 8B with those of Qwen3-VL 8B finetuned on the CoMind training set and Gemini 3 Flash, and provide the ground-truth solution given by our dataset. Finetuning on our training set improves the test set predictions.}
    \label{fig:supp_handover_baseline_comp}
\end{figure}

\begin{figure}[t]
    \centering
    \begin{subfigure}[t]{0.48\textwidth}
        \centering
        \includegraphics[width=\linewidth]{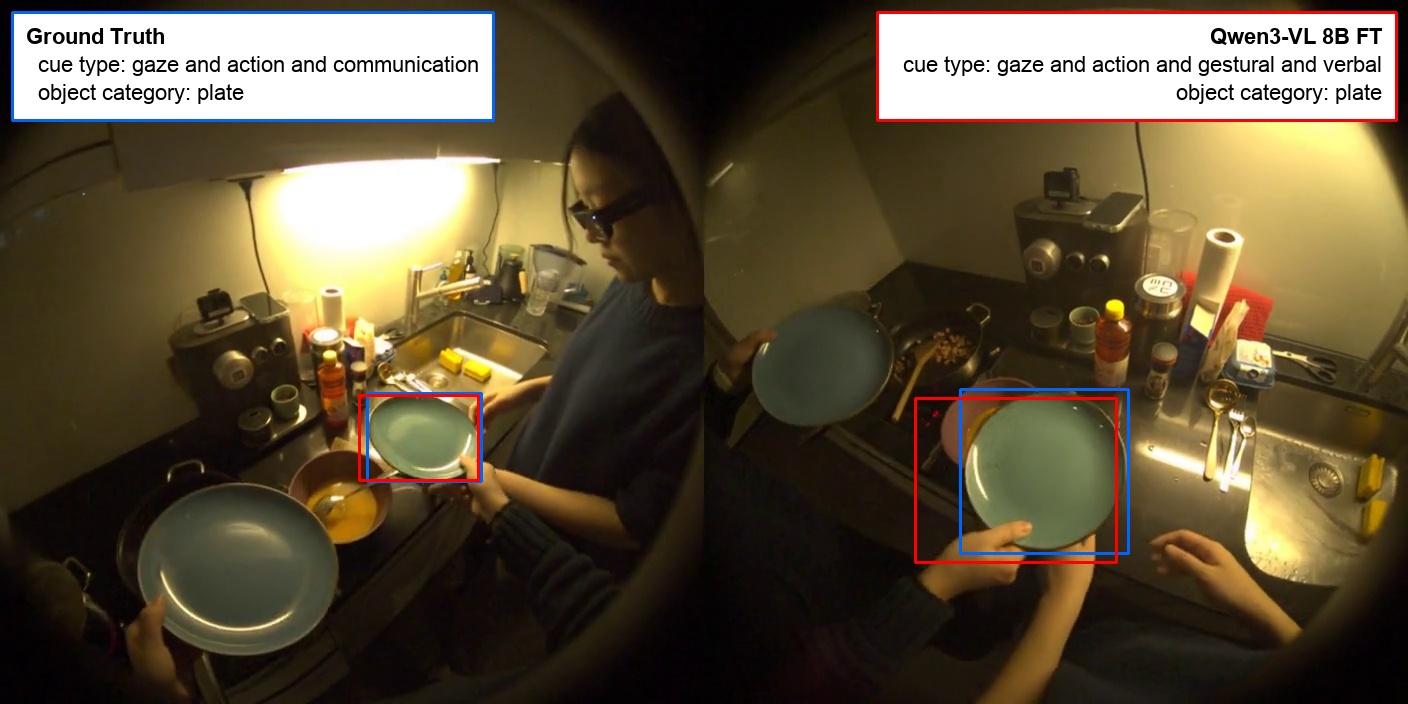}
        %\caption{First}
    \end{subfigure}
    \hfill
    \begin{subfigure}[t]{0.48\textwidth}
        \centering
        \includegraphics[width=\linewidth]{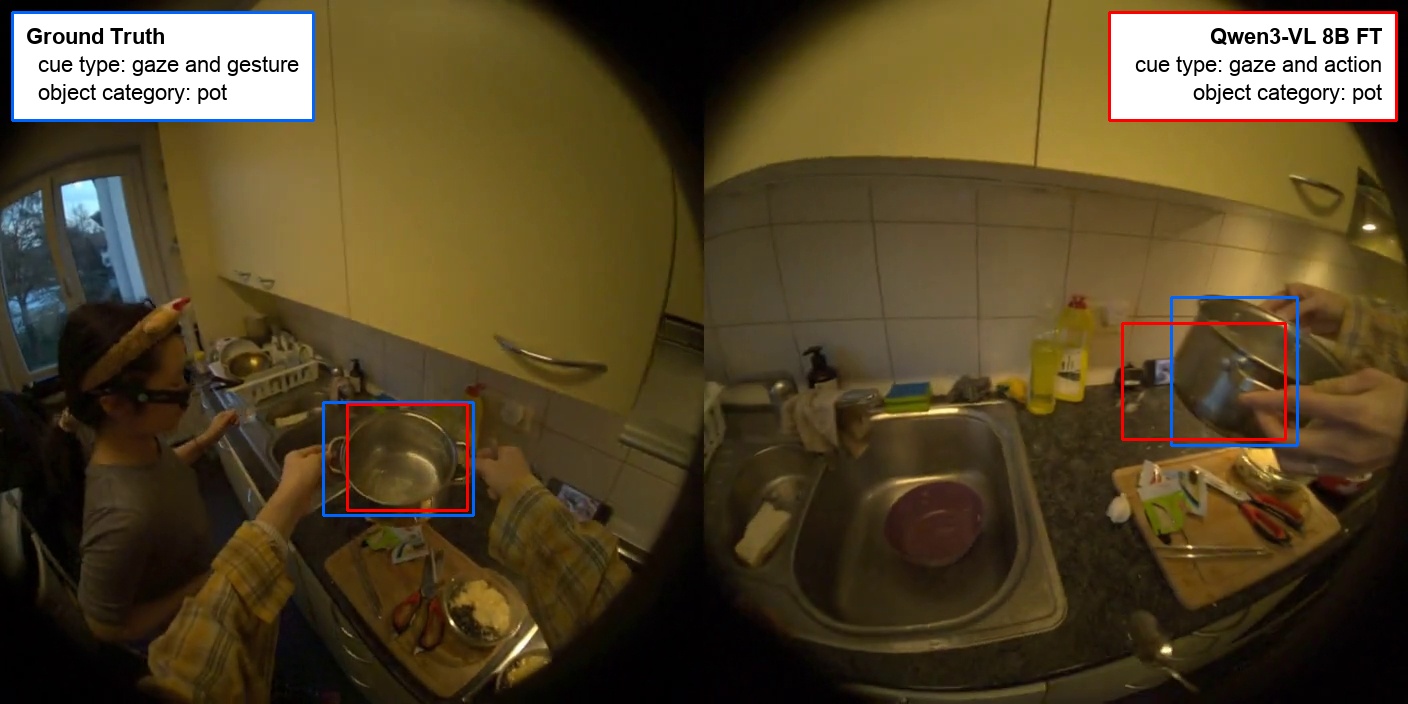}
        %\caption{Second}
    \end{subfigure}

    \vspace{0.5em}

    \begin{subfigure}[t]{0.48\textwidth}
        \centering
        \includegraphics[width=\linewidth]{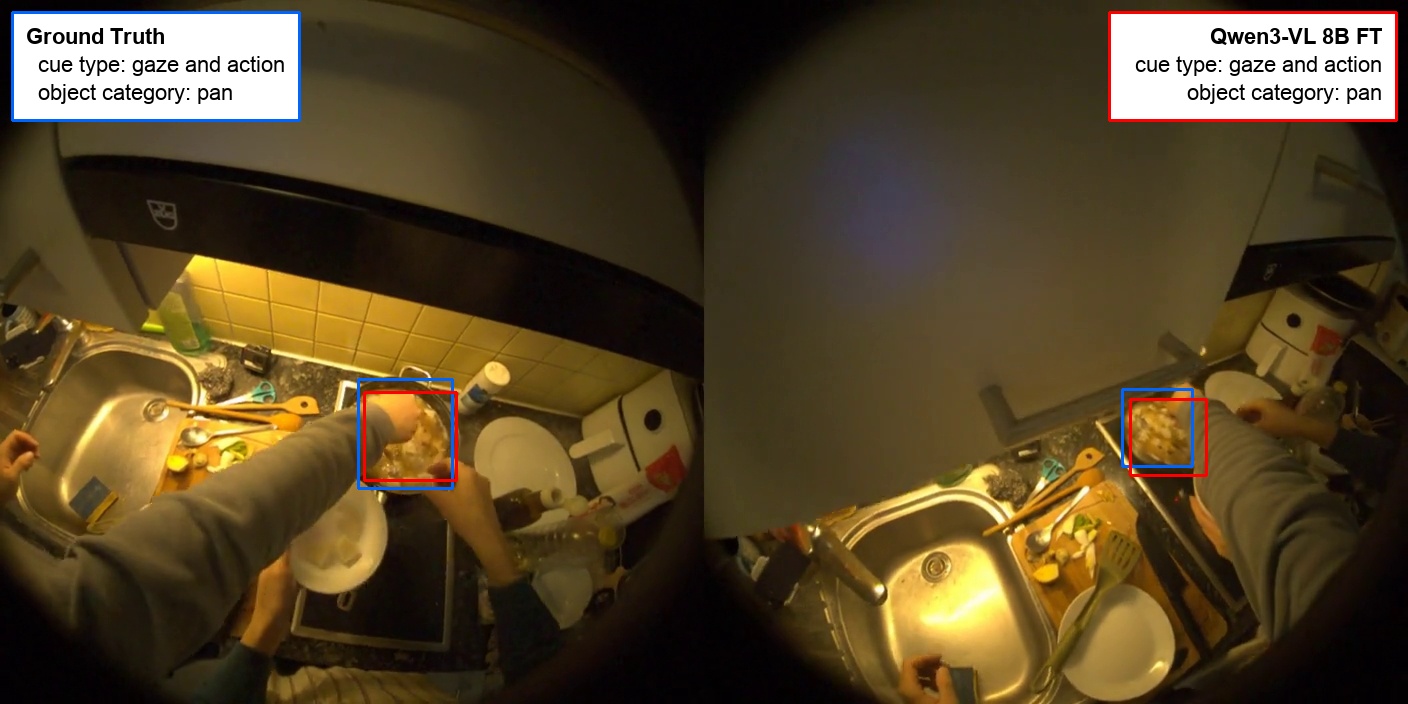}
        %\caption{First}
    \end{subfigure}
    \hfill
    \begin{subfigure}[t]{0.48\textwidth}
        \centering
        \includegraphics[width=\linewidth]{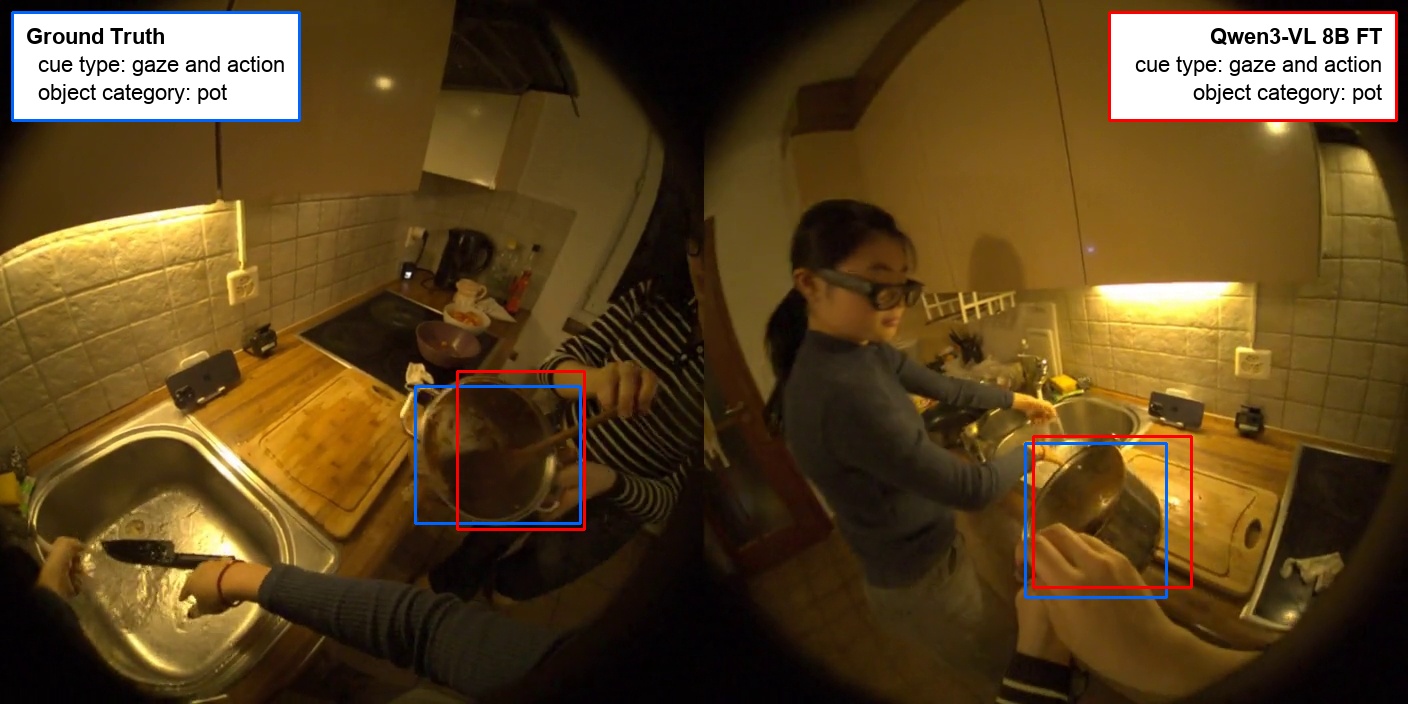}
        %\caption{Second}
    \end{subfigure}

    \vspace{0.5em}

    \begin{subfigure}[t]{0.48\textwidth}
        \centering
        \includegraphics[width=\linewidth]{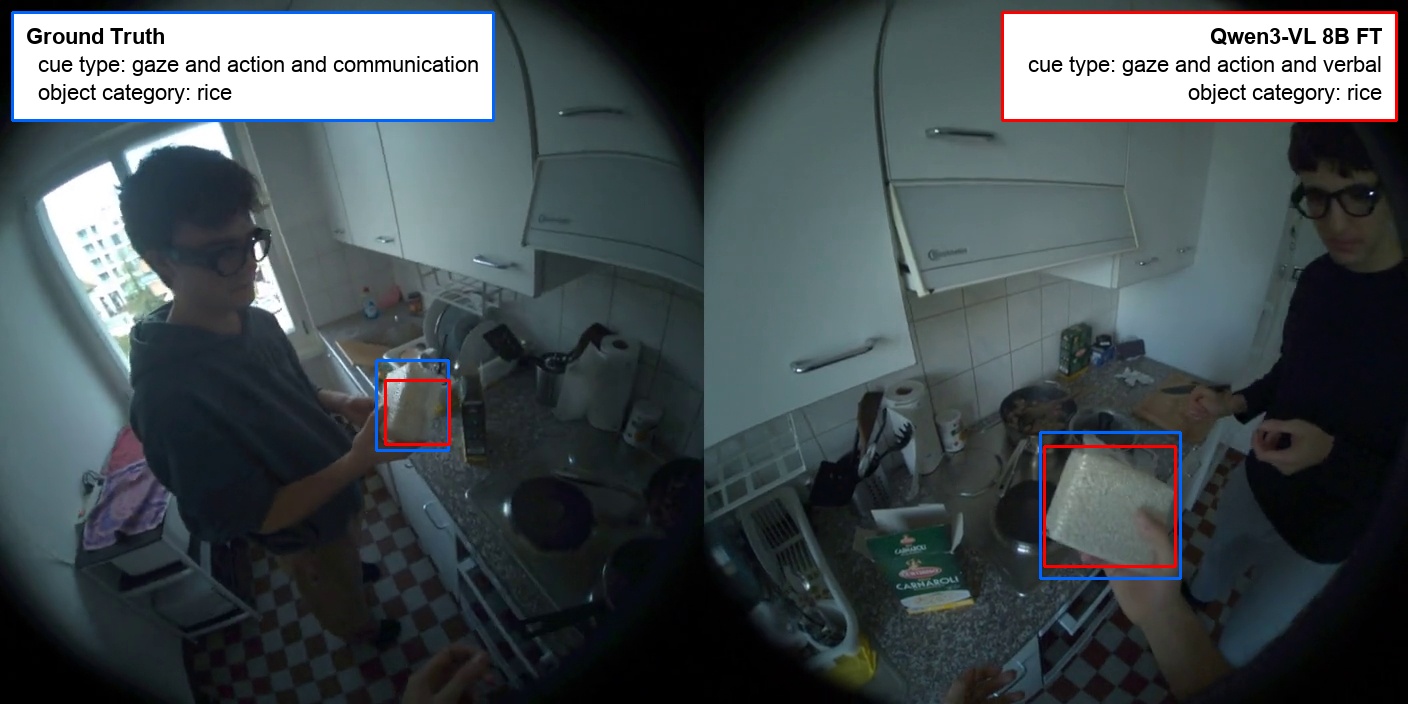}
        %\caption{Second}
    \end{subfigure}
    \hfill
    \begin{subfigure}[t]{0.48\textwidth}
        \centering
        \includegraphics[width=\linewidth]{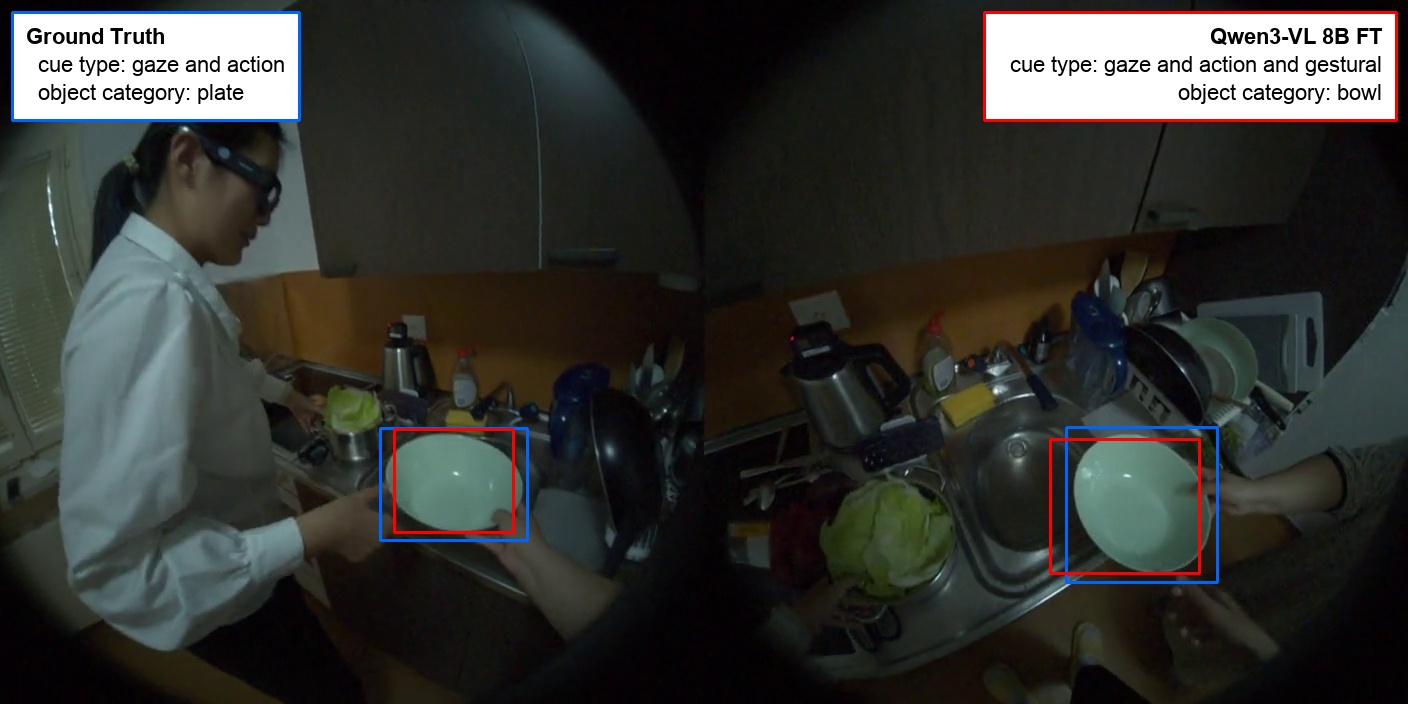}
        %\caption{First}
    \end{subfigure}

    \caption{\textbf{Further predictions on the Joint Attention Estimation task.} We visualize the predictions of Qwen3-VL 8B finetuned on our training set in red, together with the ground-truth annotations in blue.}
    \label{fig:supp_joint_attn_example}
\end{figure}

\begin{figure}[t]
    \centering

    \begin{subfigure}[t]{0.32\textwidth}
        \centering
        \includegraphics[width=\linewidth]{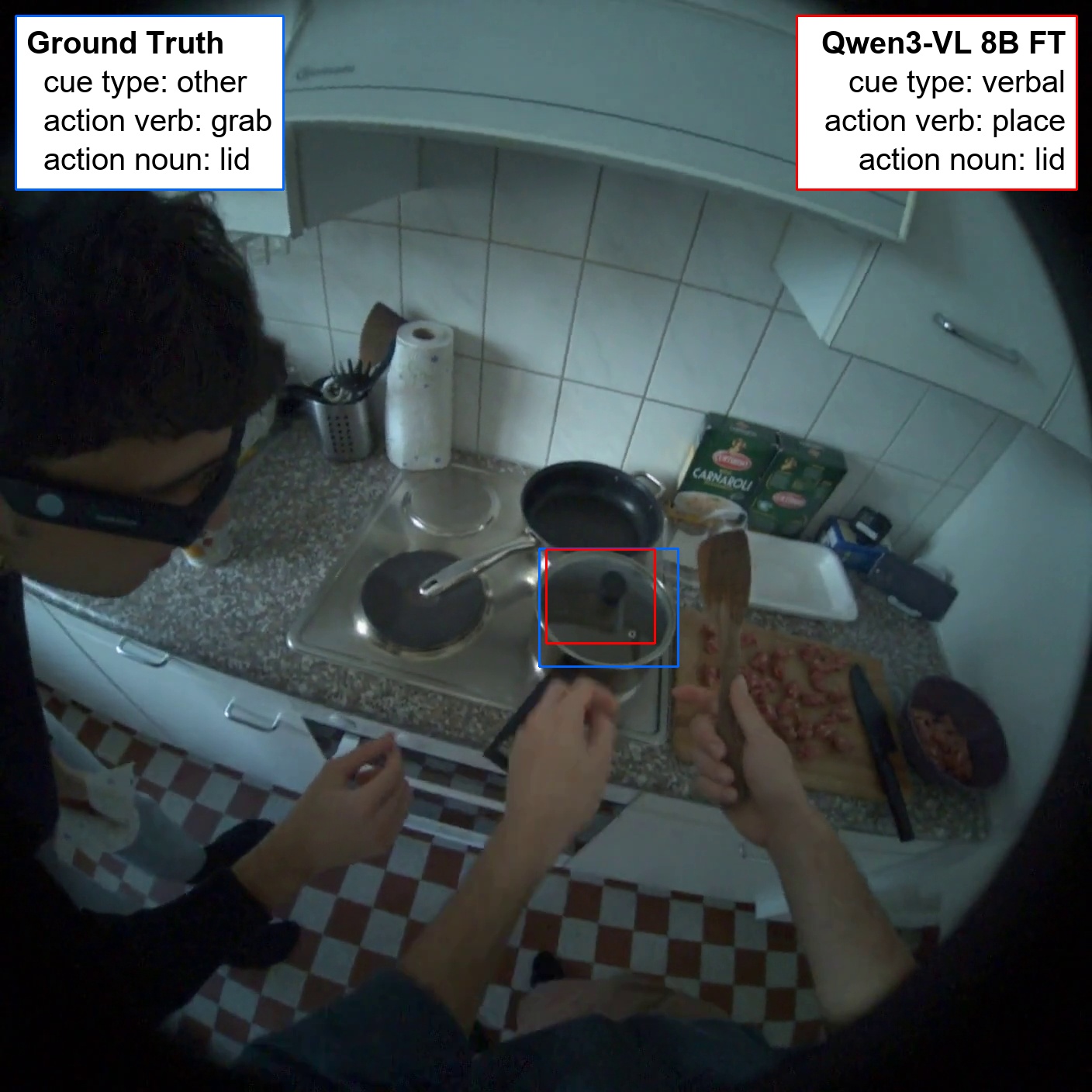}
        %\caption{First panel}
        \label{fig:panel1a}
    \end{subfigure}
    \hfill
    \begin{subfigure}[t]{0.32\textwidth}
        \centering
        \includegraphics[width=\linewidth]{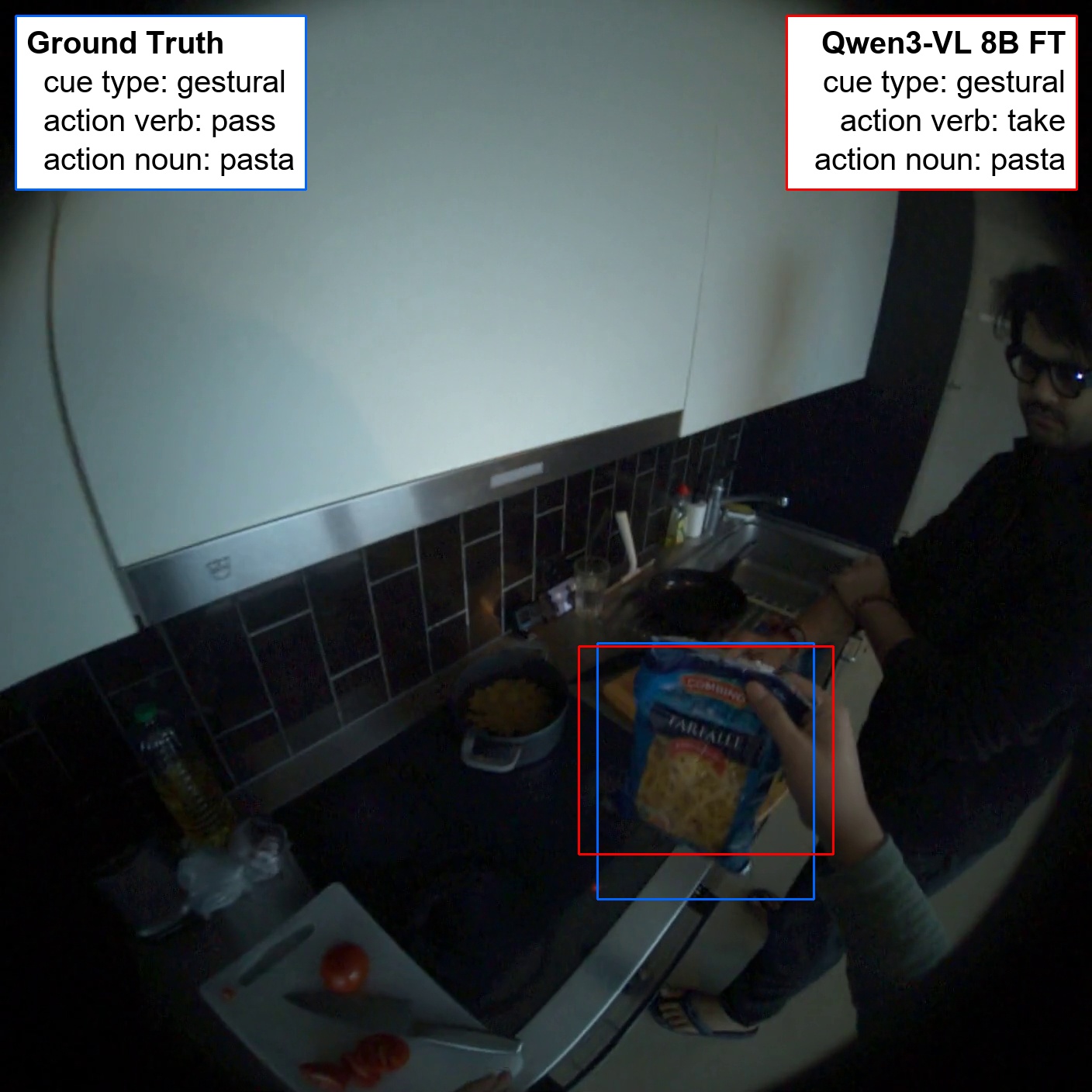}
        %\caption{Second panel}
        \label{fig:panel2a}
    \end{subfigure}
    \hfill
    \begin{subfigure}[t]{0.32\textwidth}
        \centering
        \includegraphics[width=\linewidth]{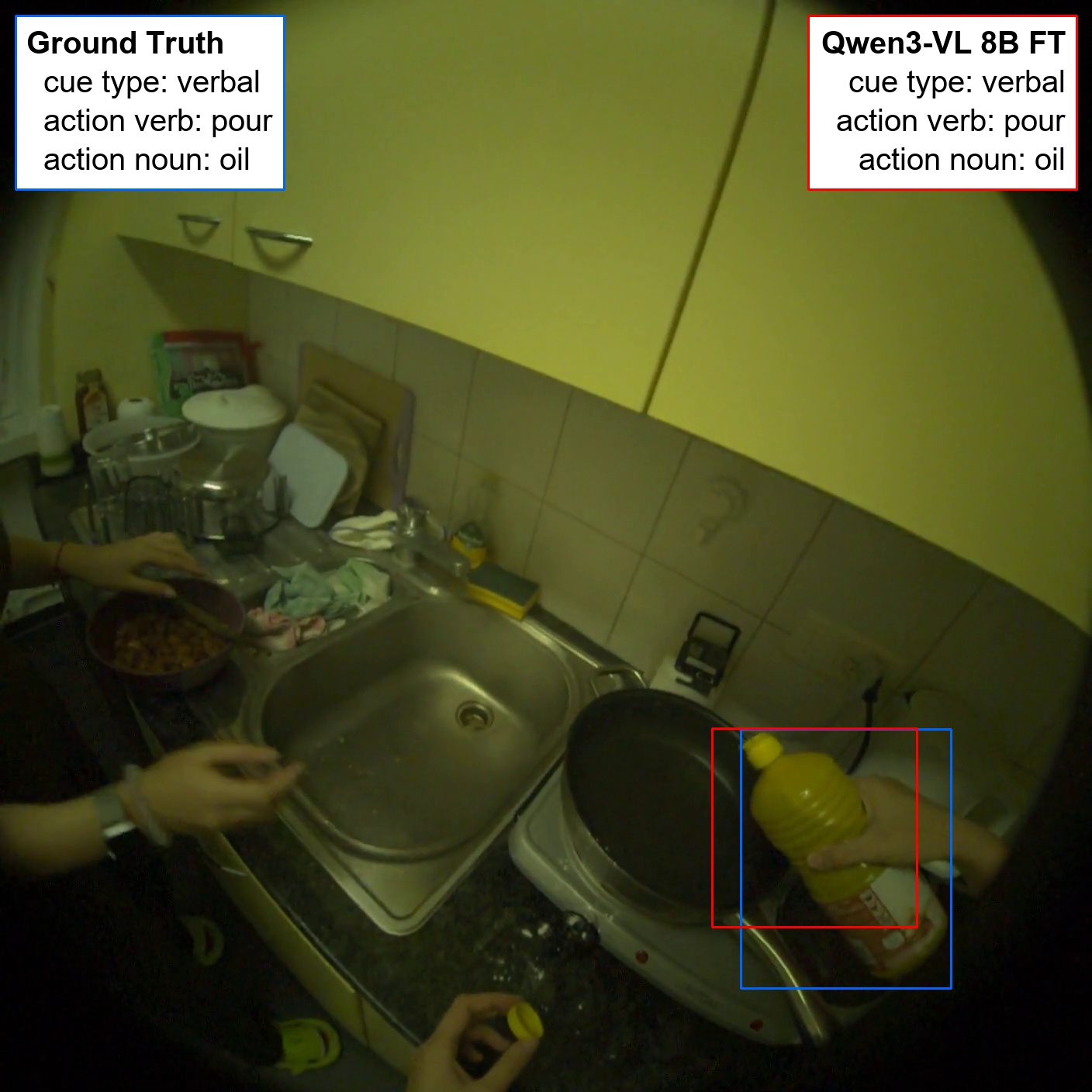}
        %\caption{Third panel}
        \label{fig:panel3a}
    \end{subfigure}

    \vspace{0.5em}

    \begin{subfigure}[t]{0.32\textwidth}
        \centering
        \includegraphics[width=\linewidth]{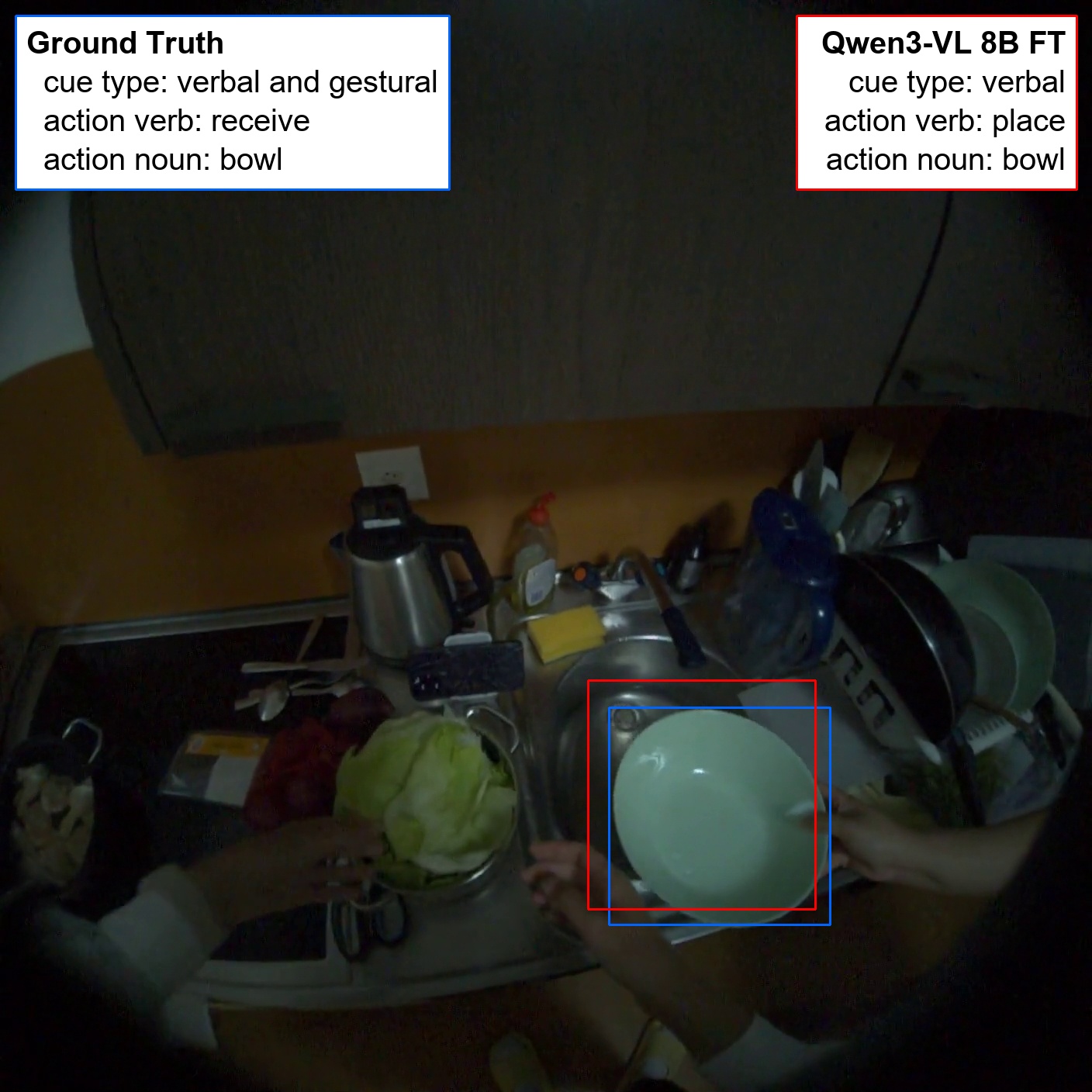}
        %\caption{First panel}
        \label{fig:panel1}
    \end{subfigure}
    \hfill
    \begin{subfigure}[t]{0.32\textwidth}
        \centering
        \includegraphics[width=\linewidth]{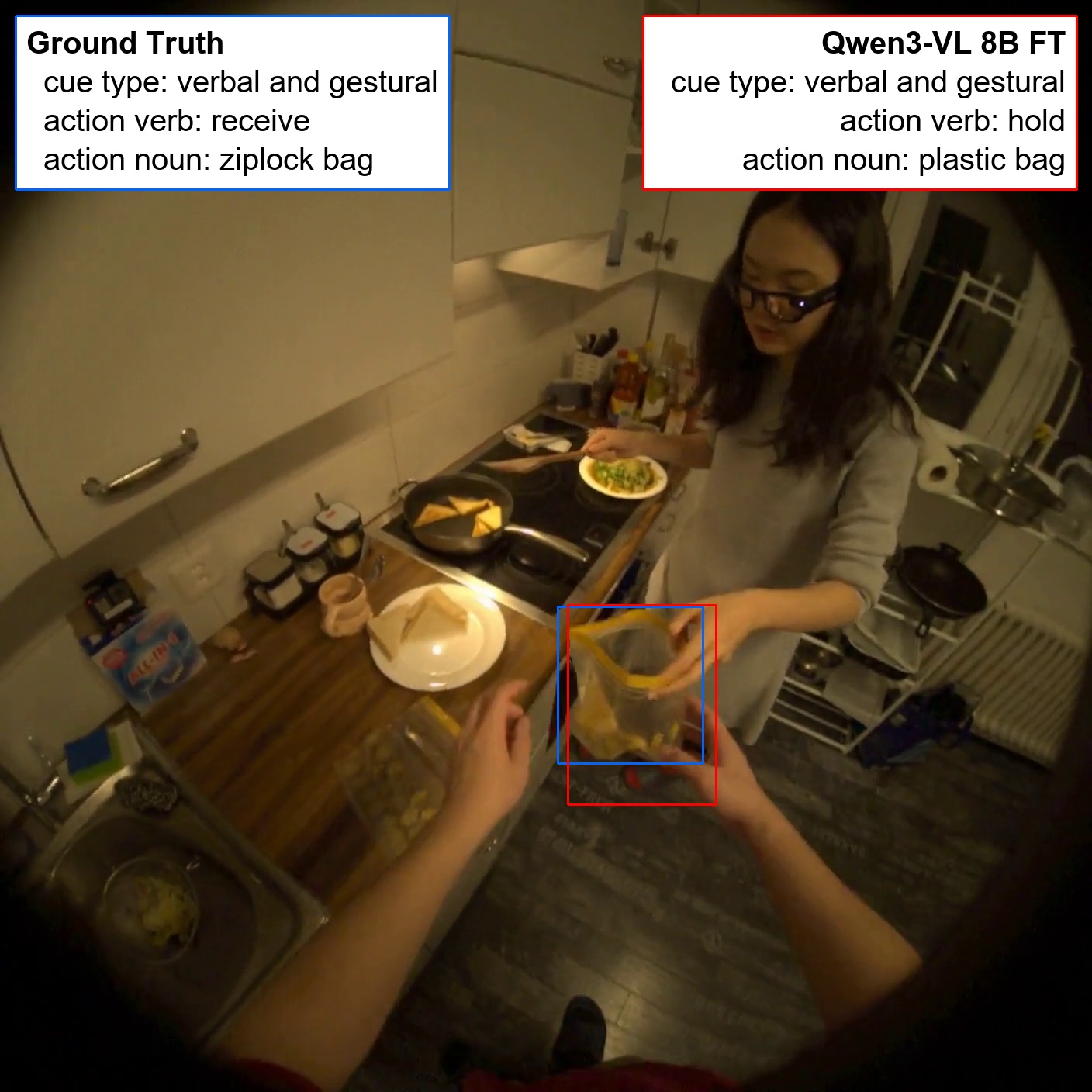}
        %\caption{Second panel}
        \label{fig:panel2}
    \end{subfigure}
    \hfill
    \begin{subfigure}[t]{0.32\textwidth}
        \centering
        \includegraphics[width=\linewidth]{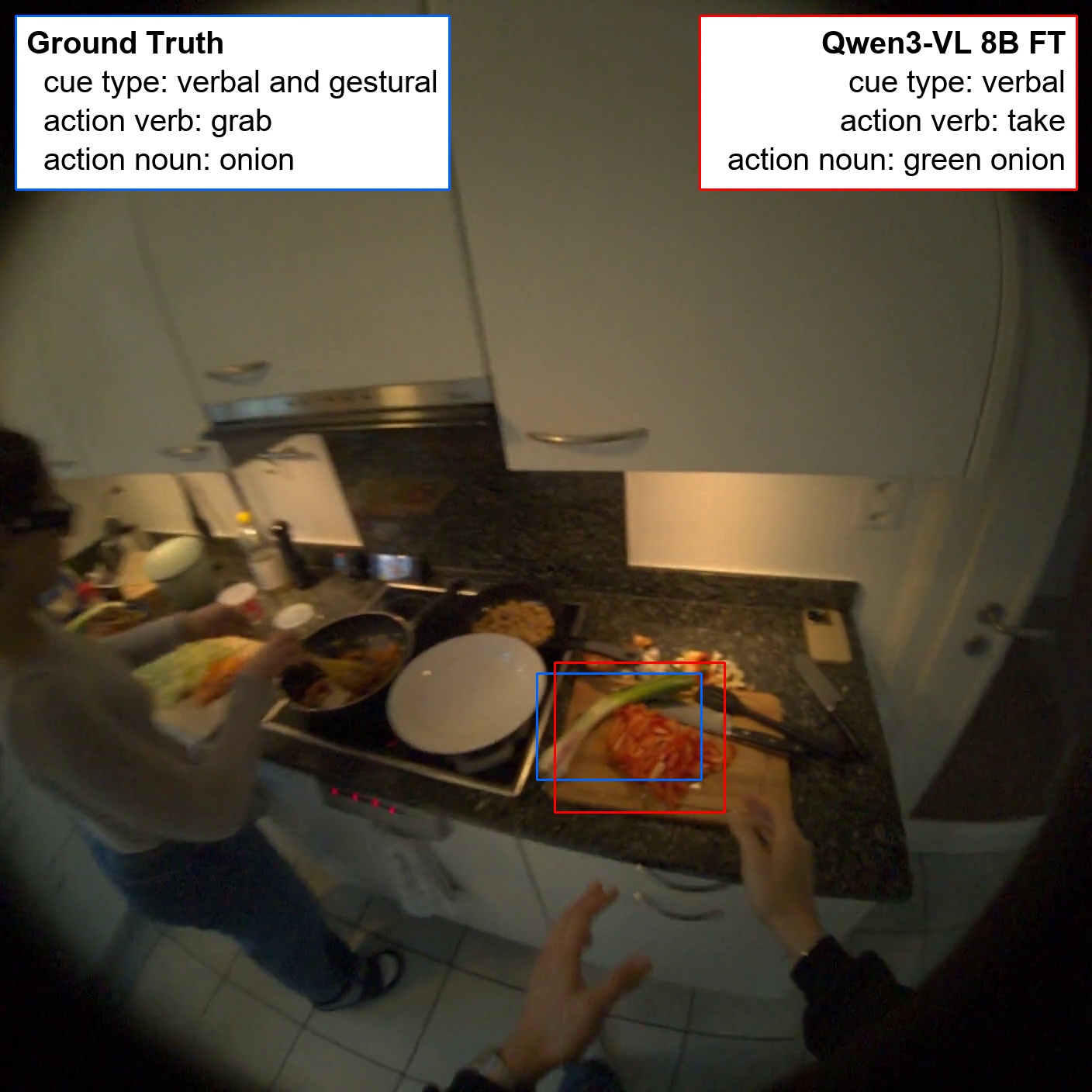}
        %\caption{Third panel}
        \label{fig:panel3}
    \end{subfigure}
    \caption{\textbf{Further predictions on the Socially Conditioned Object Interaction Anticipation task.} We visualize the predictions of Qwen3-VL 8B finetuned on our training set in red, together with the ground-truth annotations in blue.}
    \label{fig:supp_scoia_example}
\end{figure}

We provide visual comparisons of various baselines' predictions in Figs.~\ref{fig:supp_joint_attention_baseline_comp}, \ref{fig:supp_scoia_baseline_comp}, \ref{fig:supp_handover_baseline_comp}. Specifically, we visualize the predictions of Qwen-3 VL 8B, Qwen-3 VL 8B FT (finetuned on CoMind), and Gemini 3 Flash, together with the ground truth. The improved predictions of Qwen-3 VL 8B FT compared to the non-finetuned version validate the usefulness of training on CoMind for tasks associated with social reasoning. Further qualitative examples are presented in Figs.~\ref{fig:supp_joint_attn_example} and \ref{fig:supp_scoia_example}.

%%%%%%%%%%%%%%%%%%%%%
%% Joint Attention %%
%%%%%%%%%%%%%%%%%%%%%

%%%%%%%%%%%%%%%%%%%%%%%%%%%%%%%%%%%%%%%%%%%%%%%%%%%%%%%%%%
%% Socially Conditioned Object Interaction Anticipation %%
%%%%%%%%%%%%%%%%%%%%%%%%%%%%%%%%%%%%%%%%%%%%%%%%%%%%%%%%%%

%%%%%%%%%%%%%%%%%%%%%%%%%%%%%%%%%%%%%%%
%% Collaborative Handover Prediction %%
%%%%%%%%%%%%%%%%%%%%%%%%%%%%%%%%%%%%%%%

%%%%%%%%%%%%%%%%%%%%%
%% Joint Attention %%
%%%%%%%%%%%%%%%%%%%%%

%%%%%%%%%%%%%%%%%%%%%%%%%%%%%%%%%%%%%%%%%%%%%%%%%%%%%%%%%%
%% Socially Conditioned Object Interaction Anticipation %%
%%%%%%%%%%%%%%%%%%%%%%%%%%%%%%%%%%%%%%%%%%%%%%%%%%%%%%%%%%

%\begin{figure}[t]
%    \centering
%    \begin{subfigure}[t]{0.48\textwidth}
%        \centering
%        \includegraphics[width=\linewidth]{figures/supp_cherry_pick_vis/handover/cherry_pick_v1/b19c41f5-719d-4a93-92e2-0555cad8a607_023020_Qwen3-VL_8B_Instruct_FT.jpg}
        %\caption{Second}
%    \end{subfigure}
%    \hfill
%    \begin{subfigure}[t]{0.48\textwidth}
%        \centering
%        \includegraphics[width=\linewidth]{figures/supp_cherry_pick_vis/handover/cherry_pick_v1/f0b49334-565a-4d30-a010-15d92e511c9a_059377_Qwen3-VL_8B_Instruct_FT.jpg}
        %\caption{Second}
%    \end{subfigure}
%    
%    \caption{Handover}
%    \label{fig:supp_handover_example}
%\end{figure}

%TODO: Zhao

%\begin{itemize}
%    \item Baseline vs. ours examples (qualitative)
%    \item Pick things to visualize where fine-tuning actually improves the metric
%    \item Needed for each of 3 different tasks
%    \item Visualize failure cases
%    \item Is there any systematic error bias from VLMs?
%\end{itemize}

\section{Downstream Applications and Future Tasks}

\begin{figure}[t]
    \centering
    \begin{subfigure}[t]{0.48\textwidth}
        \centering
        \includegraphics[width=\linewidth]{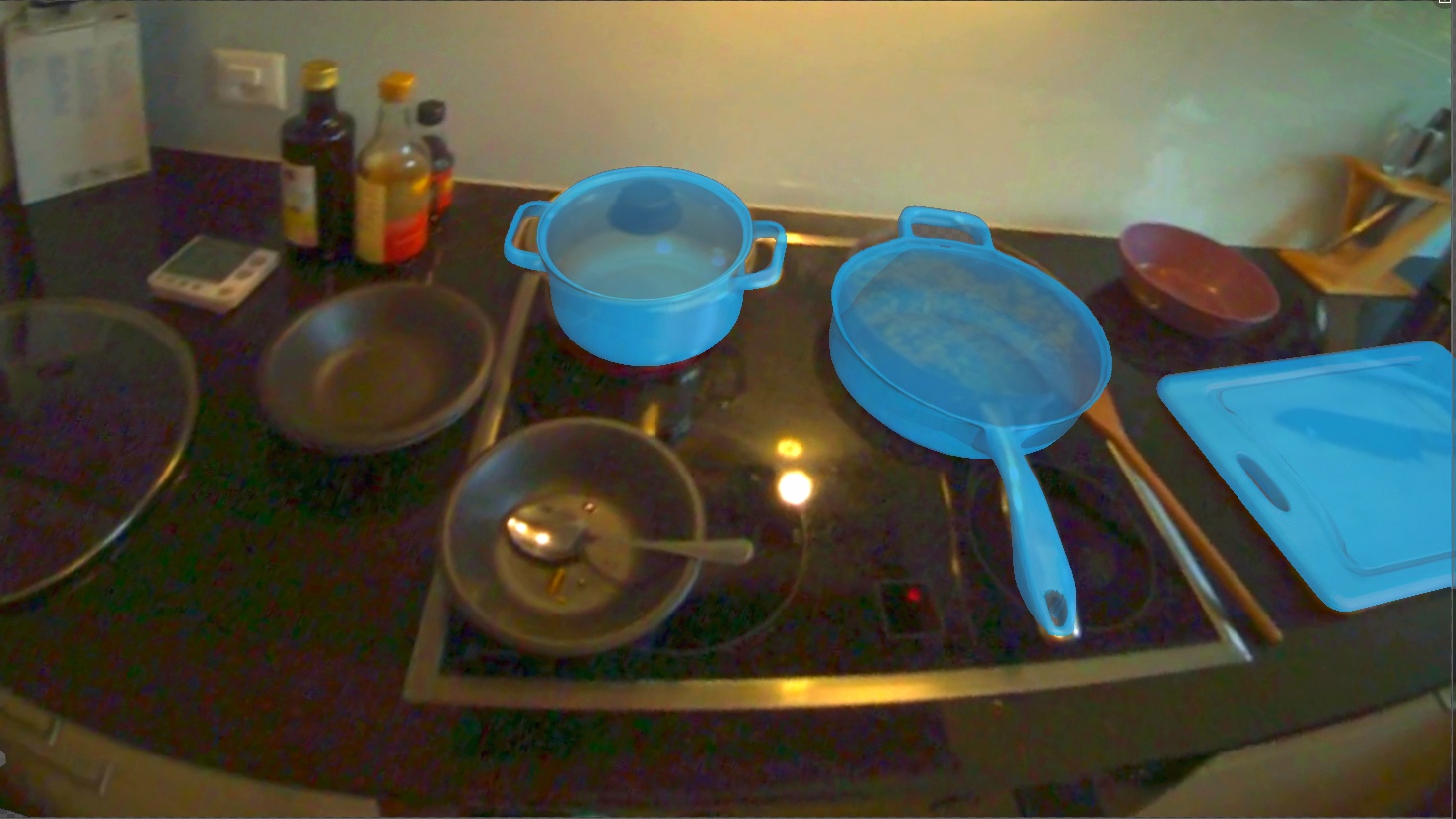}
        \caption{6D object pose estimation}
    \end{subfigure}
    \hfill
    \begin{subfigure}[t]{0.48\textwidth}
        \centering
        \includegraphics[width=\linewidth]{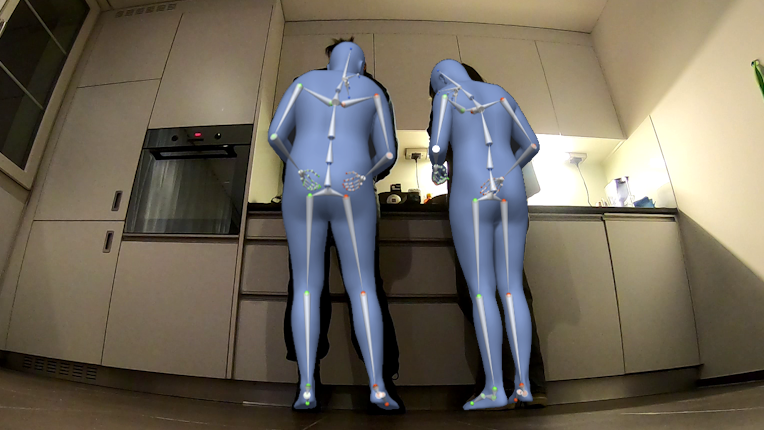}
        \caption{Human body pose estimation}
    \end{subfigure}

    \caption{\textbf{Pseudo ground-truth extraction with foundation models.} Our dataset provides modalities well-suited for pseudo ground-truth extraction using foundation models.}
    \label{fig:supp_downstream_applications}
\end{figure}

Our dataset provides multiple modalities which allow the construction of pseudo ground-truth beyond the provided annotations and data. For example, the front and back views of the participants can be used to reconstruct their body poses. While the back view offers a good view of most of the participants' bodies, the front view complements this by providing a good view of the participants' upper body, compensating for the self-occlusion of the arms and hands. Additionally, the provided object meshes can be used to reconstruct the object poses at frames where these objects are visible. The pseudo ground-truth can for instance be used to pretrain methods for later finetuning with more accurate data. We visualize possible pseudo ground-truth outputs for object pose reconstruction and body pose estimation in \autoref{fig:supp_downstream_applications}. The body poses were reconstructed using \cite{yang2026sam3dbody}.  % using methods such as \cite{foundationposewen2024}

\begin{figure}[ht]
\centering
% Replace with your actual image path
\includegraphics[width=\linewidth]{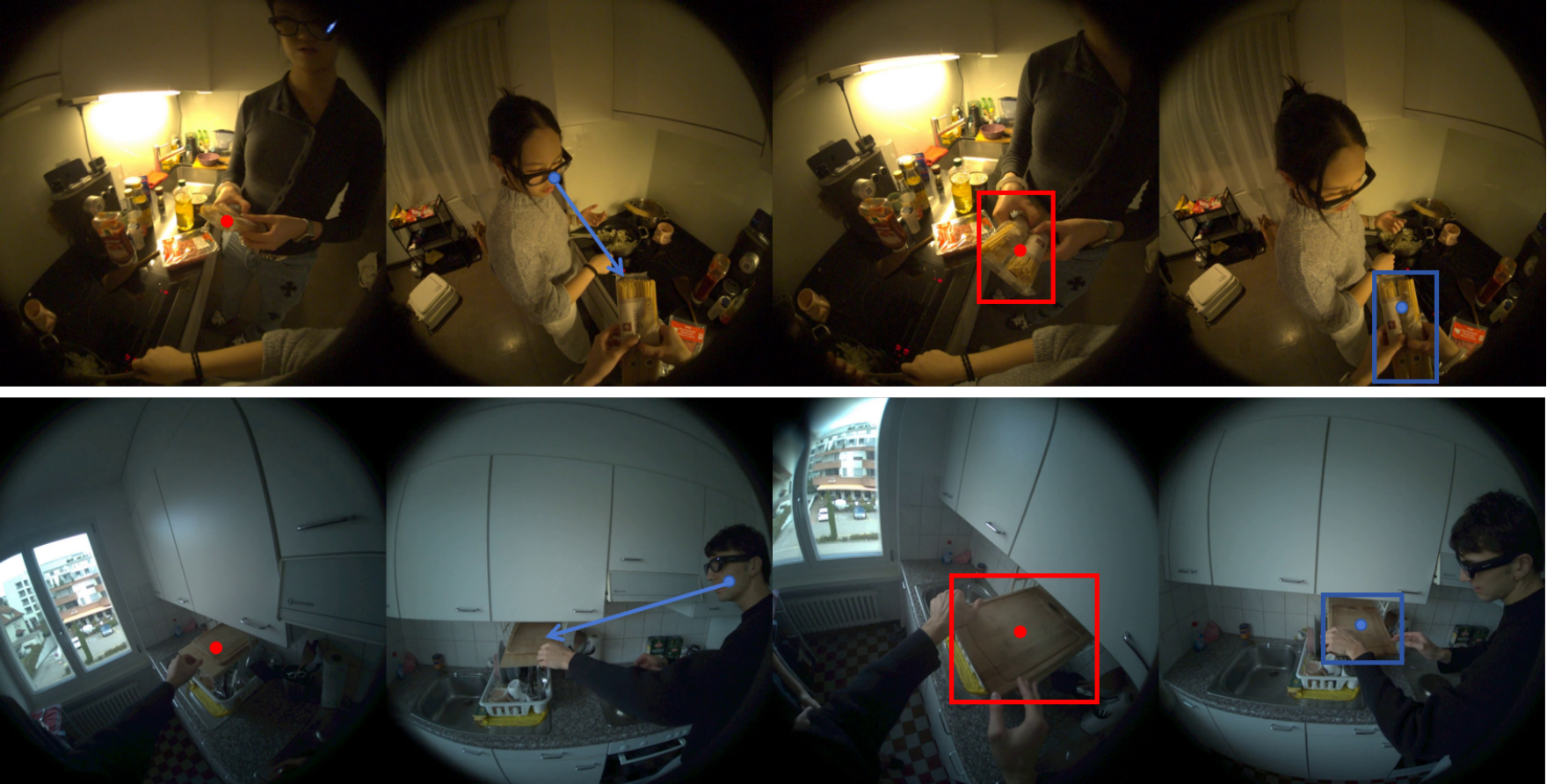} 
\caption{\textbf{Gaze Following as a precursor to Joint Attention.} In collaborative tasks, gaze following frequently serves as a vital geometric and social prior before joint attention is fully established. CoMind's synchronous dual views capture the partner's head orientation and line of sight (blue vectors). The specific gaze points of the leader and the helper are indicated by red and blue dots, respectively. This setup effectively grounds the target objects (bounding boxes) in the shared workspace prior to the interaction.}
\label{fig:gaze_following}
\end{figure}

Furthermore, our dataset naturally supports the study of \textit{gaze following}, an important social cue that frequently precedes and facilitates the establishment of joint attention \cite{tomasello2014joint, recasens2015they}. In dynamic collaborative setups, identifying where a partner is looking provides a strong visual prior for predicting imminent object interactions. As shown in \autoref{fig:gaze_following}, the synchronous dual views in CoMind uniquely capture this temporal dynamic. One perspective records the initiator's head orientation and gaze direction, while the other captures the shared workspace. The gaze vectors (blue arrows) explicitly demonstrate how a partner's line of sight grounds the target objects (bounding boxes) right before a collaborative action or joint attention event occurs. This rich visual context enables future research to develop predictive models that leverage mutual gaze for early action anticipation and social intent understanding.

% Beyond social cues, CoMind provides a robust geometric foundation for 3D spatial reasoning. By leveraging the semi-dense point clouds and camera poses estimation provided by the Aria MPS Service \cite{meta_mps_projectaria}, we can lift our 2D annotations into the 3D physical space. Specifically, similar to \cite{perrett2025hd, ashutosh2025fiction}, this lifting can be achieved by combining dense monocular depth estimates with the sparse 2d-to-3d correspondences from MPS. As shown in \autoref{fig:3d_bbox}, this process enables the reconstruction of 3D object bounding boxes within the environment point cloud. Crucially, this geometric grounding implies that all our proposed benchmark tasks can be naturally extended from the 2D image plane to the 3D world coordinate system. Such a transition from 2D pixels to 3D spatial reasoning is fundamental for advancing embodied AI and robotic manipulation.

\begin{figure}[ht]
\centering
\includegraphics[width=\linewidth]{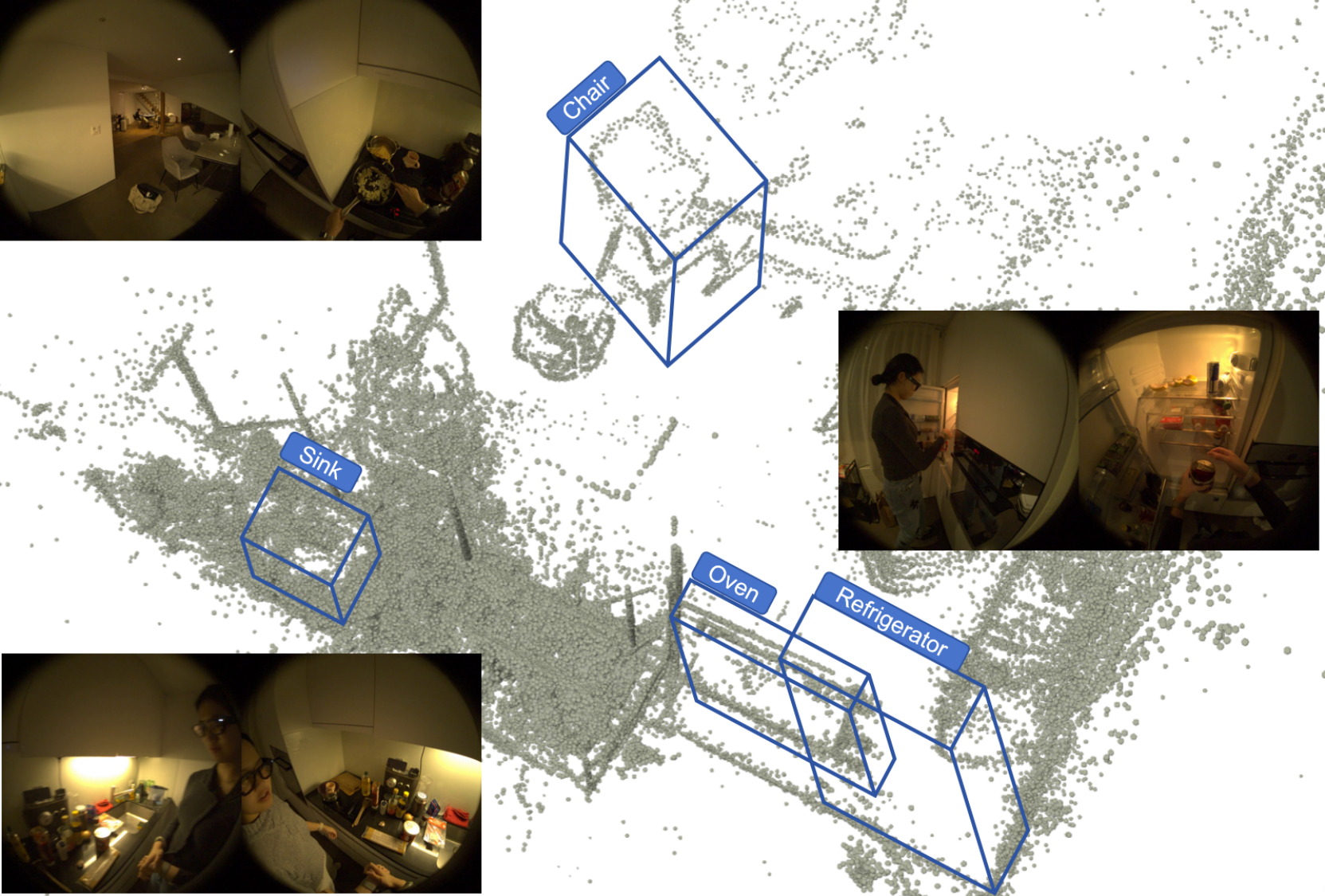} 
\caption{\textbf{Potential for 3D Spatial Reasoning.} By aligning dense depth estimates with the sparse 3D correspondences provided by the Aria MPS, our 2D bounding box annotations can be geometrically lifted into the 3D space. This figure shows reconstructed 3D bounding boxes within the scene point cloud, demonstrating CoMind's capability to support future 3D prediction tasks. Only representative static object bounding boxes shown for clarity.}
\label{fig:3d_bbox}
\end{figure}

Beyond social cues, CoMind provides a robust geometric foundation for 3D spatial reasoning. By leveraging the semi-dense point clouds and camera pose estimation provided by the Aria MPS Service \cite{meta_mps_projectaria}, we can lift our 2D annotations of static scene elements into the 3D physical space. Specifically, future work can apply 3D point cloud instance segmentation network \cite{ngo2023isbnet, Yang_2025_ICCV}, to the reconstructed 3d object bounding boxes. The exact 3D bounding boxes and geometries of these objects can then be robustly grounded by projecting our 2D labels onto the 3D segmented instances using the provided calibrated camera poses. As shown in \autoref{fig:3d_bbox}, this projection process enables the conceptual reconstruction of 3D object bounding boxes within the environment point cloud. Crucially, this geometric grounding implies that all our proposed benchmark tasks can be naturally extended from the 2D image plane to the 3D world coordinate system. Such a transition from 2D pixels to 3D spatial reasoning is fundamental for advancing embodied AI and robotic manipulation.

% \autoref{fig:supp_downstream_applications}

% \todo{Dingxi}


\section{Finetuning Hyperparameters}

We use LoRa \cite{hu2022lora} to finetune the open-source Qwen3-VL models. Specifically, we choose $r=64,\ \alpha=128$. We finetune the models for 5 epochs (Joint Attention: 3 epochs) on the training set samples of the respective tasks. The learning rate is set to $2\times10^{-4}$, and we use a batch size of 4 to accommodate increased memory usage from passing context frames.

\section{Communication Analysis}

%We analyze direct string matches between automatically generated transcriptions from the helper’s Aria glasses audio stream and object labels at L1, L2, and L3, and we report an \emph{Any-level} score that counts a single match when the same string appears at multiple consolidation levels (for example, identical L1 and L2 labels), avoiding double counting (Table~\ref{tab:helper_mentions}). This provides a conservative estimate of explicit object mentions in speech.

We evaluate matches in only the final 10 seconds before each prediction frame (Table~\ref{tab:last10s_match_rates}) to measure how often participants explicitly mention the object that is about to be interacted with. This window is informative for communication dynamics and anticipatory referencing. The analysis is intentionally strict: it only captures direct string matches and does not account for paraphrases, synonyms, or context-dependent references.

\begin{table}[ht]
\centering
\caption{\textbf{Last-10s transcript matches to ground-truth object labels (rates).} 
For each annotation, we check whether the transcript in the prior 10 seconds 
contains a direct string match to the ground-truth label at each level. 
\emph{Any level} counts a single match if any of the Level 1--3 ground-truth labels match.}
\label{tab:last10s_match_rates}
\resizebox{\linewidth}{!}{
\begin{tabular}{l c c c c}
\toprule
Task & Level 1 & Level 2 & Level 3 & Any level \\
\midrule
Socially Conditioned Object Interaction & 13.57\% & 7.96\% & 0.06\% & 13.97\% \\
Collaborative Handover Prediction & 14.16\% & 9.04\% & 0.00\% & 14.85\% \\
Joint Attention Detection & $\phantom{1}$5.39\% & 2.81\% & 0.02\% & $\phantom{1}$5.61\% \\
\bottomrule
\end{tabular}
% \begin{tabular}{l c c c c}
% \toprule
% Task & Level 1 & Level 2 & Level 3 & Any level \\
% \midrule
% Socially Conditioned Object Interaction & 14.02\% & 7.73\% & 0.00\% & 14.83\% \\
% Collaborative Handover Prediction & 13.76\% & 8.99\% & 0.00\% & 15.21\% \\
% Joint Attention Detection & $\phantom{1}$5.76\% & 2.90\% & 0.03\% & $\phantom{1}$6.21\% \\
% \bottomrule
% \end{tabular}
}
\vspace{-12pt}
\end{table}
%%%

\section{Noun Hierarchy \& Consolidation Details}

\myparagraph{Verb consolidation}
Action verbs were consolidated into a fixed set of canonical interaction classes in order to reduce lexical variation while preserving distinctions that remain meaningful for embodied collaboration. This mapping resolves minor spelling errors, formatting differences, and closely related synonyms that describe the same underlying action (e.g., \textit{take}, \textit{pick}, and \textit{hold} are mapped to \textit{grab}; \textit{put}, \textit{put down}, and \textit{putdown} are mapped to \textit{place}; \textit{stir}, \textit{fold}, and \textit{beat} are mapped to \textit{mix}). The final verb vocabulary contains 21 canonical classes: \textit{grab}, \textit{receive}, \textit{pass}, \textit{place}, \textit{open}, \textit{add}, \textit{turn on}, \textit{cut}, \textit{pour}, \textit{press}, \textit{move}, \textit{rotate}, \textit{turn off}, \textit{scoop}, \textit{mix}, \textit{dispose}, \textit{apply}, \textit{clean}, \textit{close}, \textit{roll}, and \textit{stretch}. These categories were chosen to preserve action-level differences that are visually and functionally relevant, while avoiding fragmentation caused by free-form wording.

\myparagraph{Noun consolidation}
Noun labels were consolidated using a three-level hierarchy. Level~1 preserves object identity as closely as possible. At this level, we primarily normalize surface form by correcting spelling and formatting inconsistencies, resolving singular/plural variants, and mapping alternative names for the same object to a single canonical label. Thus, Level~1 generally leaves the underlying object unchanged and serves mainly to standardize how the same object is referred to across annotations and predictions.

Level~2 groups Level~1 nouns into semantically and functionally related mid-level categories. These groupings were defined according to the role of the object in the interaction and its semantic similarity to other objects in the dataset. For example, ingredients are grouped into classes such as \textit{vegetable}, \textit{meat}, \textit{egg}, \textit{grain}, \textit{dairy}, \textit{spice}, \textit{sauce}, and \textit{veggie protein}; manipulated tools are grouped into classes such as \textit{kitchen utensil}, \textit{spoon}, \textit{spatula}, and \textit{scrubber}; and environmental or infrastructure objects are grouped into classes such as \textit{container}, \textit{cupboard}, \textit{prep area}, \textit{sink area}, and \textit{stove}. The goal of Level~2 is to preserve interaction-relevant distinctions while reducing unnecessary label sparsity.

\begin{figure*}[t]
  \centering
  \includegraphics[width=0.75\textwidth]{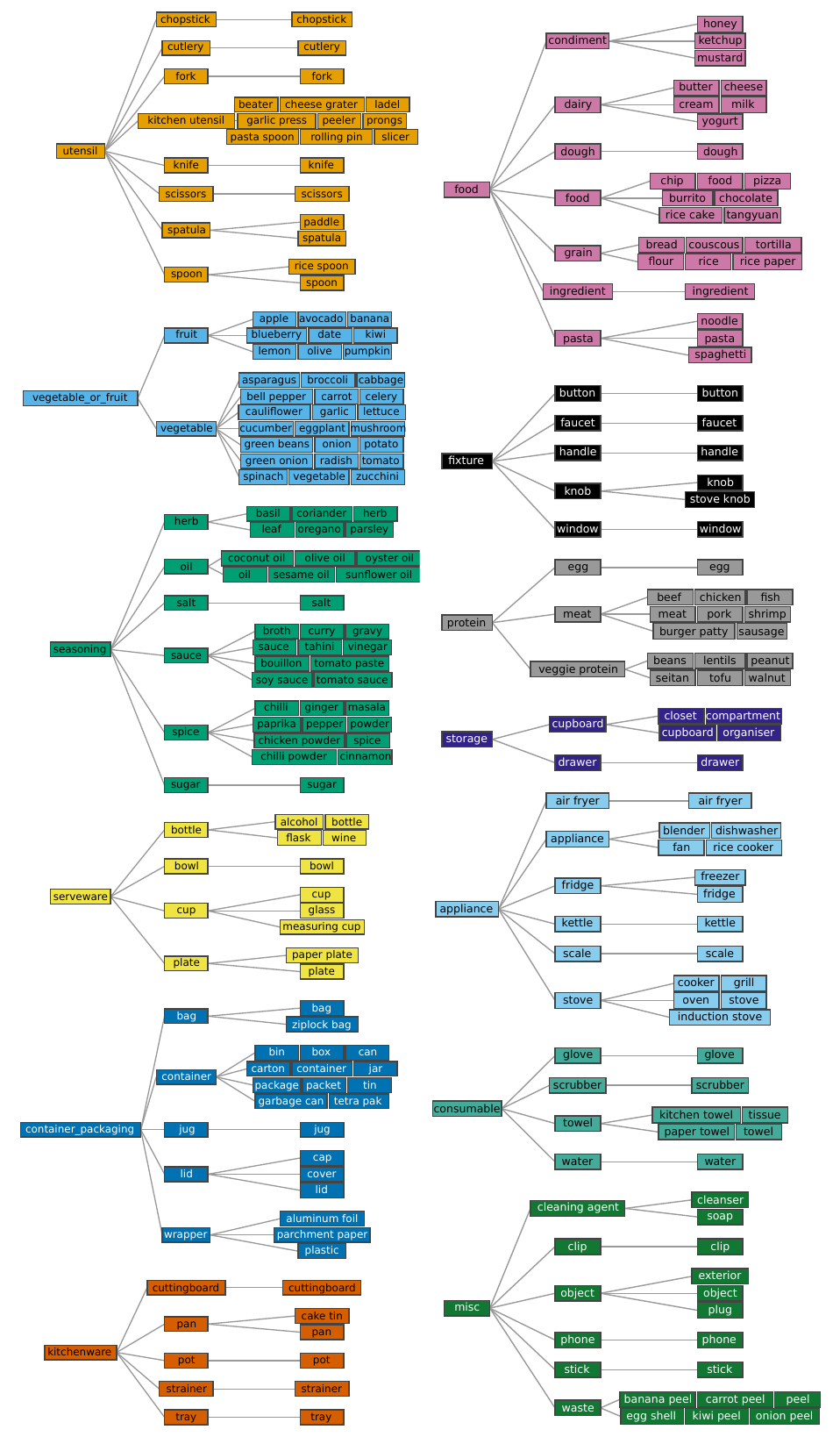}
  \vspace{-8pt}
\caption{\textbf{Object hierarchy tree for the CoMind Dataset.} Boxes show L3 (left), L2 (middle), and L1 (right) categories; lines indicate parent–child relations. Colors are keyed to L3 groups.}
  \label{fig:object-hierarchy-tree}
  \vspace{-16pt}
\end{figure*}

Level~3 maps Level~2 categories into a small set of broad super-categories: \textit{appliance}, \textit{area}, \textit{consumable}, \textit{container\_packaging}, \textit{fixture}, \textit{food}, \textit{kitchenware}, \textit{misc}, \textit{protein}, \textit{seasoning}, \textit{serveware}, \textit{storage}, \textit{utensil}, and \textit{vegetable\_or\_fruit}. These categories were designed to be broad enough for stable aggregation, but not so broad that they remove the dominant semantic structure of the dataset. Because the dataset is strongly focused on kitchen and food-related activity, collapsing all edible items into a single \textit{food} category would obscure a substantial amount of meaningful variation. We therefore retain several distinct high-level food-related groups at Level~3, such as \textit{food}, \textit{protein}, \textit{seasoning}, and \textit{vegetable\_or\_fruit}, while consolidating non-food objects into comparably broad functional categories. A detailed visualization of the full consolidation hierarchy is shown in \ref{fig:object-hierarchy-tree}.

\section{Input Ablations}

We ablate the importance of different types of input in \autoref{tab:ablations}. Specifically, we provide results for Gemini 3 Flash and Qwen3-VL 8B instruct on the Socially Conditioned Object Interaction Anticipation task, varying whether context frames and/or transcripts are provided as part of the input. The prediction frame is always provided and the prompts are slightly adapted depending on the presence/absence of certain input types. We observe that passing transcripts to both models greatly helps them with predicting the cue type and the action noun (i.e., object category). Furthermore, context frames improve the accuracy on action verb and action noun prediction. 

\begin{table}[t]
  \centering
  \caption{\textbf{Ablation study on context frames and transcripts on the Socially Conditioned Object Interaction Anticipation task}. $\mathrm{IoU@0.5}$ denotes the fraction of predicted bounding boxes with $\mathrm{IoU} > 0.5$ with the ground-truth object to be interacted with. Cue Type reports cue classification accuracy, and Act. Verb resp.~Act. Noun (L1) measures the fraction of correct verb resp.~L1 object category predictions. Providing transcripts greatly improves cue type and act.~noun accuracy. Context frames improve act.~verb and act.~noun accuracy.}
  \resizebox{\linewidth}{!}{
  \label{tab:ablations}
  \setlength{\tabcolsep}{5pt}
  \begin{tabular}{lcc cccc}
    \toprule
    \textbf{Model} & \textbf{Context} & \textbf{Transcript} & \textbf{IoU} & \textbf{Cue} & \textbf{Act.} & \textbf{Act.} \\
    & \textbf{Frames} & & \textbf{@0.50} & \textbf{Type} & \textbf{Verb} & \textbf{Noun (L1)} \\
    \midrule
    \multirow{4}{*}{Gemini 3 Flash}
      & --         & --           & 0.1867 & 0.2954 & 0.0437 & 0.2384 \\
      & --         & \checkmark   & \textbf{0.1954} & \textbf{0.5404} & 0.0623 & 0.3748 \\
      & \checkmark & --           & 0.1449 & 0.3272 & 0.0517 & 0.2874 \\
      & \checkmark & \checkmark   & 0.1859 & 0.5298 & \textbf{0.0781} & \textbf{0.3775} \\
    \midrule
    \multirow{3}{*}{Qwen3-VL 8B I.}
      & --         & --           & \textbf{0.0909} & 0.0887 & 0.0596 & 0.2199 \\
      & --         & \checkmark   & 0.0024 & 0.4159 & 0.0848 & 0.3258 \\
      & \checkmark & --           & 0.0020 & 0.1192 & 0.0768 & 0.2450 \\
      & \checkmark & \checkmark           & 0.0114 & \textbf{0.4172} & \textbf{0.0993} & \textbf{0.3298} \\
    \bottomrule
  \end{tabular}
}
\end{table}

\section{Prompting Details}

\noindent We use the three task-specific prompts for the respective tasks. To maintain a fair evaluation, we provide the same task-specific prompt to all models. The prompts for the Joint Attention Estimation, Socially Conditioned Object Interaction Anticipation and Collaborative Handover Prediction Tasks are given in Figures~\ref{fig:prompt_joint_attention}, \ref{fig:prompt_scoia} and \ref{fig:prompt_handover}, respectively.\\[0.3cm]
%%%%%%%%%%%%%%%%%%%%%
%% Joint Attention %%
%%%%%%%%%%%%%%%%%%%%%

\begin{figure}
\begin{verbatimquote}You are an expert system for analyzing and predicting joint attention between two persons separately filming a single scene from their perspectives. The persons are working in a human-to-human collaboration scenario. Joint attention occurs when both individuals simultaneously focus their perception and cognitive state onto the same shared object or event.
You are provided with a dual-view (side-by-side) image frame showing two egocentric views from two persons (left person and right person). Both halves show the EXACT SAME physical scene from two different spatial perspectives.
Additionally, here is the transcribed speech of the two persons during the time period leading up to the moment in the provided image frame: "{transcript}"

TASK:
Carefully observe the visual interactions and social cues in the frame and analyze the moment of joint attention shown in the frame. Answer the following questions sequentially:
1. Cue type: Analyze the frame and classify the underlying cue(s) that reveal the joint attention. Choose from one or more of these four categories (1 to 4 choices):
   - <gaze>: The eye gaze or head pose of both individuals clearly converges onto the same target object or spatial area.
   - <action>: The joint attention is established through a physical, functional manipulation (e.g. grasping, handing over, or manipulating a tool).
   - <gestural>: The attention is directed via a communicative body movement (e.g. pointing a finger, holding an object up for display).
   - <verbal>: The shared focus is established via verbal means (e.g. following a spoken instruction or comment).
2. Object grounding: Look at the provided image frame. Based on your answers to the previous questions, locate and identify the precise object of joint attention in BOTH views SEPARATELY:
   - Object category: Provide a short, precise category name of the object being mutually attended to by both persons.
   - Left object bounding box: Provide the bounding box coordinates of the object of joint attention specifically within the LEFT person's egocentric view.
   - Right object bounding box: Provide the bounding box coordinates of the EXACT SAME object specifically within the RIGHT person's egocentric view.

OUTPUT FORMAT:
Provide your final answer as a JSON list containing a single dictionary. You may think step-by-step beforehand, but do not provide any additional text in the output.
[
  {
    "cue type": correct subset of ["<gaze>", "<action>", "<gestural>", "<verbal>"],
    "object category": "<string>",
    "left object bounding box": [<ymin>, <xmin>, <ymax>, <xmax>],
    "right object bounding box": [<ymin>, <xmin>, <ymax>, <xmax>]
  }
]

CRITICAL RULES:
- All bounding box coordinates MUST be normalized integers between [0, 1000] relative to the overall CONCATENATED image dimensions.
- For `left object bounding box`, its `<xmin>` and `<xmax>` values MUST STRICTLY fall between 0 and 500.
- For `right object bounding box`, its `<xmin>` and `<xmax>` values MUST STRICTLY fall between 500 and 1000.
\end{verbatimquote}
\caption{\textbf{Prompt for the Joint Attention Estimation task.} }
\label{fig:prompt_joint_attention}
\end{figure}

%%%%%%%%%%%%%%%%%%%%%%%%%%%%%%%%%%%%%%%%%%%%%%%%%%%%%%%%%%
%% Socially Conditioned Object Interaction Anticipation %%
%%%%%%%%%%%%%%%%%%%%%%%%%%%%%%%%%%%%%%%%%%%%%%%%%%%%%%%%%%

\begin{figure}
\begin{verbatimquote}
You are an expert handover recognition and prediction AI analyzing a human-to-human collaboration scenario.
You are provided with a series of sequential dual-view (side-by-side) images showing two egocentric views from two persons (left person and right person) over the span of 10 seconds, and a final single reference image frame.
Additionally, here is the transcribed speech of the two persons during the considered time period: "{transcript}"

TASK:
Carefully observe the human behaviors and social cues and answer the following questions sequentially:
1. Handover Timing: Approximately how many seconds after the reference image frame will the object handover between two persons occur? (Provide a decimal time offset).
2. Delivering Flow: Whether the object will be handed over from left person to right person or from right person to left person? (Choose STRICTLY: <left to right> or <right to left>).
3. Initiator: Who initiates the handover? 'Initiation' refers to the first cue that prompts the handover to occur. (Choose STRICTLY: <left> or <right>).
4. Initiation Type: Classify the initiator's cue STRICTLY into one of these four categories:
   - <verbal>: The cue is PURELY spoken (e.g., a verbal request/command).
   - <gestural>: The cue is PURELY a physical gesture (e.g., reaching out, pointing) with NO speech.
   - <verbal and gestural>: BOTH speech and a physical gesture are used simultaneously to demand the object.
   - <implicit>: ALL OTHER cases. The handover occurs naturally due to situational context, mutual routine, or proactive offering, with NO explicit prior speaking or reaching out from the receiver.
5. Object Grounding: Look at the PROVIDED SINGLE REFERENCE IMAGE FRAME. Based on the context of the video clip and your answers to the previous questions, locate the precise object that is most likely going to be handed over in the future:
   - Provide a short, precise object category name.
   - Then, provide the bounding box of this object within the PROVIDED SINGLE REFERENCE IMAGE FRAME, specifically in the GIVER'S view (determine who the giver is, left view or right view, based on your answer to the 'Delivering Flow').

OUTPUT FORMAT:
Provide your final answer as a JSON list containing a single dictionary. You may think step-by-step beforehand, but do not provide any additional text in the output.
[
  {
    "handover happen offset": <int>,
    "delivering flow": "<left to right> or <right to left>",
    "initiator": "<left> or <right>",
    "initiation type": "<verbal>, <gestural>, <verbal and gestural> or <implicit>",
    "object category": "<string>",
    "object bounding box": [<ymin>, <xmin>, <ymax>, <xmax>]
  }
]

CRITICAL RULES:
- The bounding box coordinates MUST be normalized integers between [0, 1000] relative to the overall image dimensions.
\end{verbatimquote}
\caption{\textbf{Prompt for the Socially Conditioned Object Interaction Anticipation task.}}
\label{fig:prompt_scoia}
\end{figure}

%%%%%%%%%%%%%%%%%%%%%%%%%%%%%%%%%%%%%%%
%% Collaborative Handover Prediction %%
%%%%%%%%%%%%%%%%%%%%%%%%%%%%%%%%%%%%%%%

\begin{figure}
\begin{verbatimquote}You are an expert handover recognition and prediction AI analyzing a human-to-human collaboration scenario.
You are provided with a series of sequential dual-view (side-by-side) images showing two egocentric views from two persons (left person and right person) over the span of 10 seconds, and a final single reference image frame.
Additionally, here is the transcribed speech of the two persons during the considered time period: "{transcript}"

TASK:
Carefully observe the human behaviors and social cues and answer the following questions sequentially:
1. Handover Timing: Approximately how many seconds after the reference image frame will the object handover between two persons occur? (Provide a decimal time offset).
2. Delivering Flow: Whether the object will be handed over from left person to right person or from right person to left person? (Choose STRICTLY: <left to right> or <right to left>).
3. Initiator: Who initiates the handover? 'Initiation' refers to the first cue that prompts the handover to occur. (Choose STRICTLY: <left> or <right>).
4. Initiation Type: Classify the initiator's cue STRICTLY into one of these four categories:
   - <verbal>: The cue is PURELY spoken (e.g., a verbal request/command).
   - <gestural>: The cue is PURELY a physical gesture (e.g., reaching out, pointing) with NO speech.
   - <verbal and gestural>: BOTH speech and a physical gesture are used simultaneously to demand the object.
   - <implicit>: ALL OTHER cases. The handover occurs naturally due to situational context, mutual routine, or proactive offering, with NO explicit prior speaking or reaching out from the receiver.
5. Object Grounding: Look at the PROVIDED SINGLE REFERENCE IMAGE FRAME. Based on the context of the video clip and your answers to the previous questions, locate the precise object that is most likely going to be handed over in the future:
   - Provide a short, precise object category name.
   - Then, provide the bounding box of this object within the PROVIDED SINGLE REFERENCE IMAGE FRAME, specifically in the GIVER'S view (determine who the giver is, left view or right view, based on your answer to the 'Delivering Flow').

OUTPUT FORMAT:
Provide your final answer as a JSON list containing a single dictionary. You may think step-by-step beforehand, but do not provide any additional text in the output.
[
  {
    "handover happen offset": <int>,
    "delivering flow": "<left to right> or <right to left>",
    "initiator": "<left> or <right>",
    "initiation type": "<verbal>, <gestural>, <verbal and gestural> or <implicit>",
    "object category": "<string>",
    "object bounding box": [<ymin>, <xmin>, <ymax>, <xmax>]
  }
]

CRITICAL RULES:
- The bounding box coordinates MUST be normalized integers between [0, 1000] relative to the overall image dimensions.
\end{verbatimquote}
\caption{\textbf{Prompt for the Collaborative Handover Prediction task.}}
\label{fig:prompt_handover}
\end{figure}

\vspace{-0.8cm}
\section{Annotation Details}

\begin{figure}
    \centering
    \includegraphics[width=0.98\linewidth]{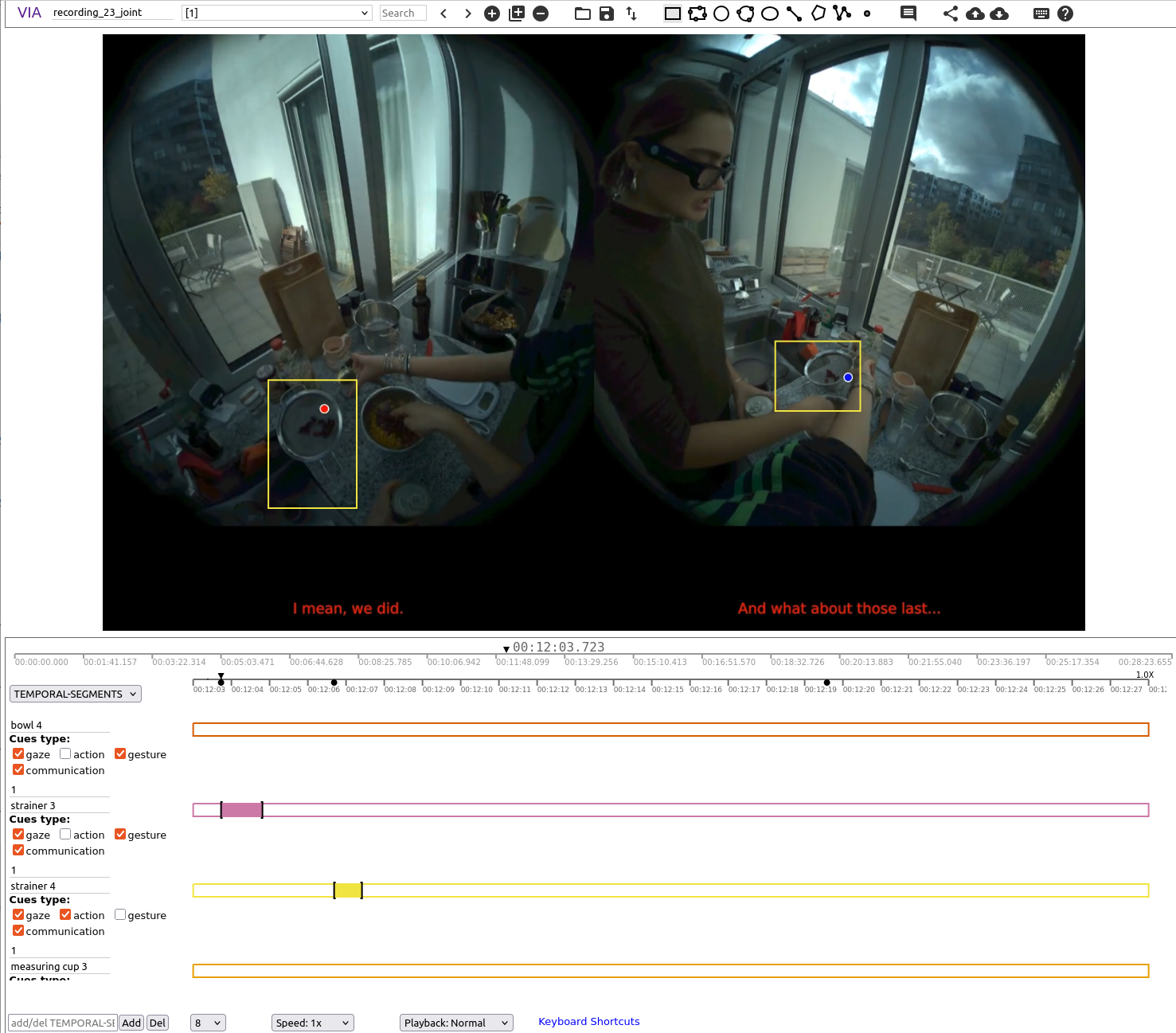}
    \caption{\textbf{Annotation interface for the Joint Attention Estimation Task.} The task-specific annotation interface is intuitive, accessible over the internet and synchronizes outputs to a central database, making annotation and supervision easily scalable. Transcripts are provided as subtitles to facilitate and speed up annotation.}
    \label{fig:via_jointattn}
\end{figure}

\begin{figure}
    \centering
    \includegraphics[width=0.98\linewidth]{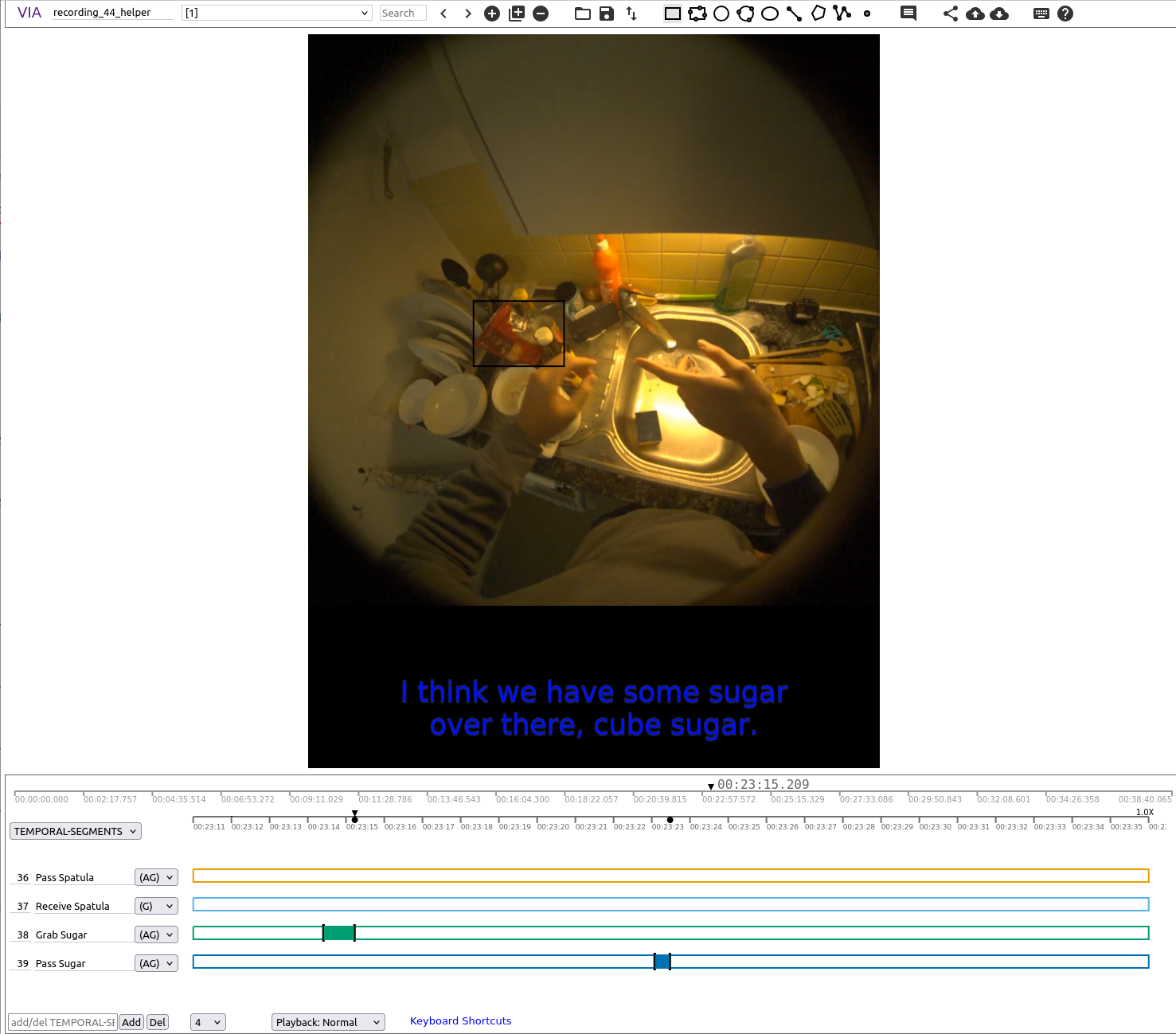}
    \caption{\textbf{Annotation interface for the Socially Conditioned Object Interaction Anticipation task.} The task-specific annotation interface is intuitive, accessible over the internet and synchronizes outputs to a central database, making annotation and supervision easily scalable. Transcripts are provided as subtitles to facilitate and speed up annotation.}
    \label{fig:via_scoia}
\end{figure}

\begin{figure}
    \centering
    \includegraphics[width=0.98\linewidth]{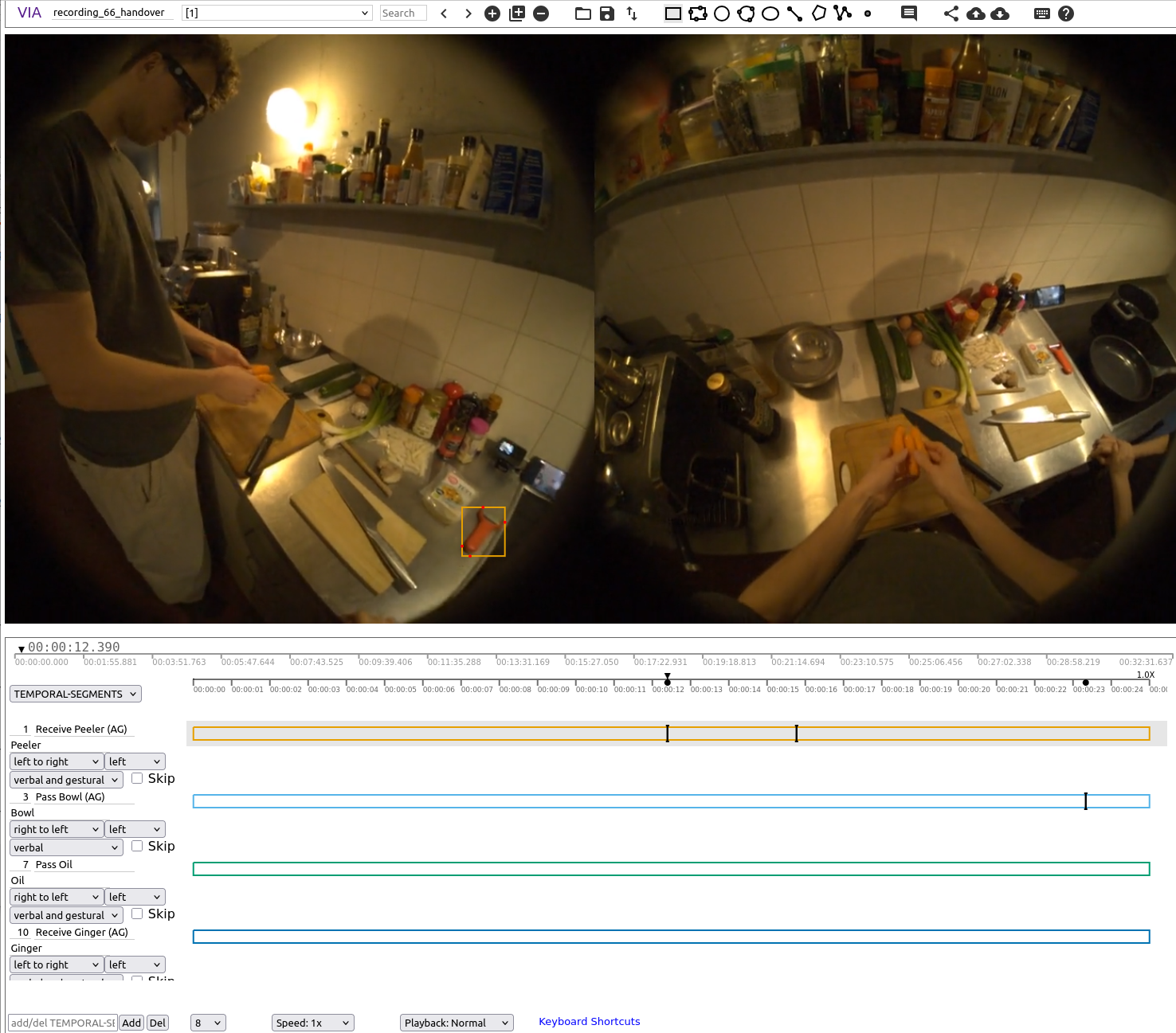}
    \caption{\textbf{Annotation interface for the Collaborative Handover Prediction task.} The task-specific annotation interface is intuitive, accessible over the internet and synchronizes outputs to a central database, making annotation and supervision easily scalable.}
    \label{fig:via_handover}
\end{figure}

\myparagraph{Annotation statistics}
Table~\ref{tab:benchmark_annotation_stats} summarizes the annotation coverage for the three benchmark tasks. 
For each task we report the number of annotated interaction events completed. 
We also report the resulting annotation density measured as events per hour of video.

The Joint Attention Detection task contains the largest number of annotations with 8,513 events, corresponding to an average density of 209.1 events per hour.  % 31.2 hours of data
The Socially Conditioned Object Interaction task includes 1,747 annotated interaction events, corresponding to 42.9 events per hour.  % across 63 recordings (36.18 hours)
Finally, the Collaborative Handover Prediction task contains 586 annotated handover events, corresponding to 14.4 events per hour. % across 79 recordings totaling 43.3 hours
These statistics illustrate the differing annotation densities across tasks, reflecting the varying frequency and temporal structure of the underlying interaction types.

\begin{table}[ht]
\centering
\caption{\textbf{Annotation statistics for the benchmark tasks.} For each task we report the number of recordings containing annotations, the total duration of the corresponding streams, the number of annotated interaction events, and the resulting annotation density measured as events per hour of video.}
\label{tab:benchmark_annotation_stats}
\setlength{\tabcolsep}{7pt}
\small
\resizebox{\linewidth}{!}{
\begin{tabular}{l c c c c}
\toprule
Task  & \# Events & Events / hour \\
\midrule
Socially Conditioned Object Interaction & 1,747 & $\phantom{1}$42.9 \\
Collaborative Handover Prediction & $\phantom{1}$586 & $\phantom{1}$14.4 \\
Joint Attention Detection & 8,513 & 209.1 \\
\bottomrule
\end{tabular}
}
\vspace{-12pt}
\end{table}

\myparagraph{Annotation interfaces}
We design task-specific annotation interfaces based on the VIA Video Annotator \cite{dutta2019vgg}. The interfaces are visualized in Figs.~\ref{fig:via_jointattn}, \ref{fig:via_scoia} and \ref{fig:via_handover}. Each interface is tailored to easily annotate the labels required for its associated task. The interfaces are accessible by a secret web link and synchronize to a central database. The videos to be annotated are uploaded to a content delivery network to eliminate the need of downloading large video files and allow dynamic updating of the videos if necessary. This system makes annotation and review possible using every machine with a web browser, allowing for effortless scaling of the annotation force and easy reviewing of the annotations.

\myparagraph{Annotation instructions} 
% Instructions provided to the annotators are attached at the end of this document. Care was taken to keep the language easy to understand and maintain a dynamic list of edge cases and instructions on how to handle them.
The instructions provided to annotators are included at the end of this document. Special care was taken to ensure that the language remains easy to understand. In addition, a dynamic list of edge cases and corresponding handling guidelines was maintained throughout the annotation process.

\section{Ethical Considerations}

\begin{figure}[htbp]
    \centering
    \includegraphics[width=0.9\textwidth]{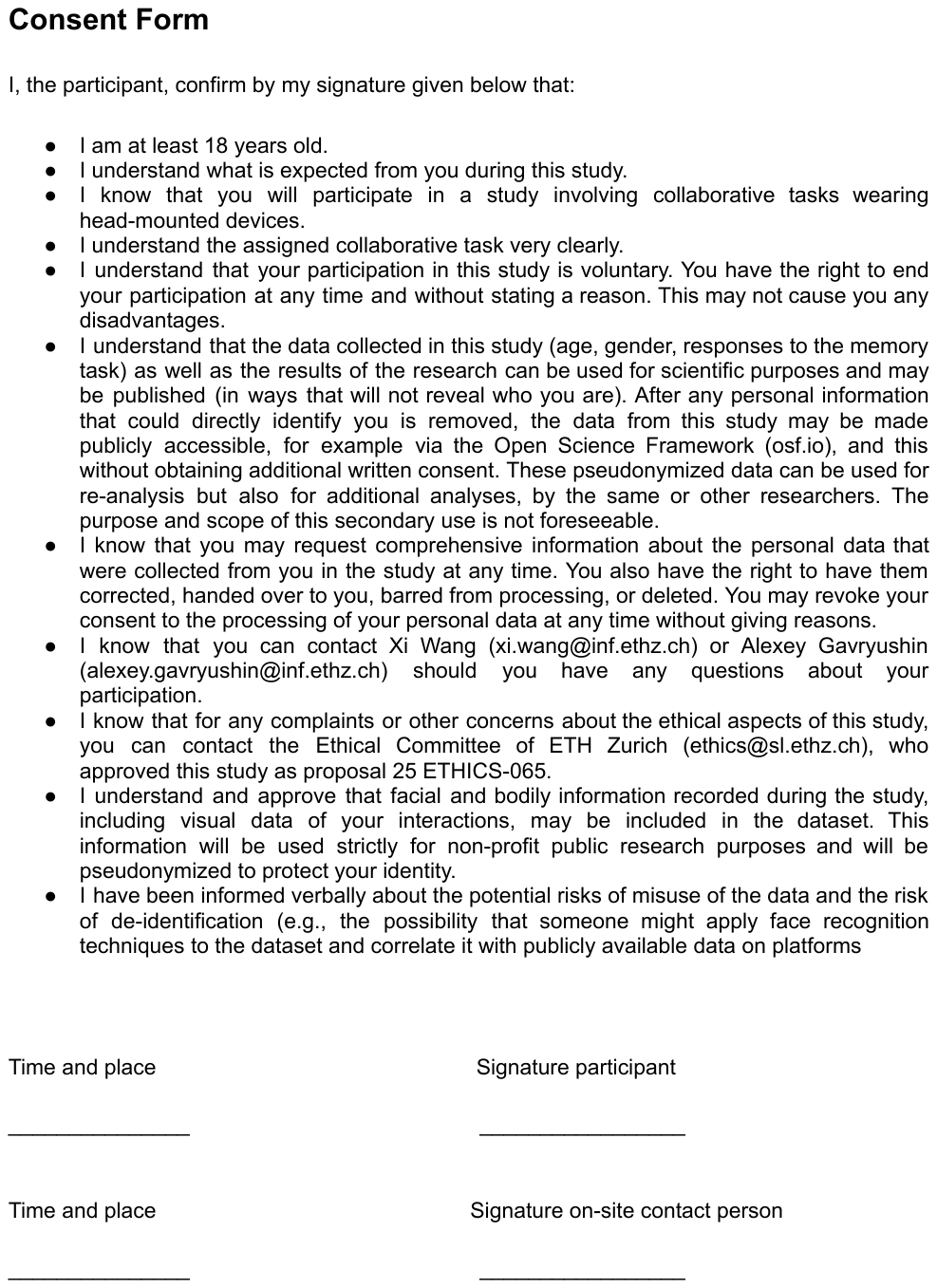}
    \caption{Consent form.}
    \label{fig:consent_form}
\end{figure}

The data collection received full approval from ETH Zurich's Ethics Commission (Project 25 ETHICS-065; approved by Vice President for Research Prof.\ Christian Wolfrum and Chair of the Ethics Commission Prof.\ Lutz Wingert) and was conducted in accordance with established ethical standards. All participants provided written informed consent covering the release of their videos with identifiable faces. Faces are intentionally left unblurred, as gaze and expression are core signals for Theory of Mind. The consent form can be found in \cref{fig:consent_form}.

%\input{figures/teaser}

%\input{sec/0_abstract}

%\input{sec/1_introduction}

%\input{sec/2_related_work}

%\input{sec/3_task_definition}

%\input{sec/4_collection}

% \input{sec/5_dataset_statistics}

%\input{sec/6_experiments}

%\input{sec/7_conclusion}

% Now we have reached the maximum length of an ECCV \ECCVyear{} submission (excluding references and acknowledgements).
% References should start immediately after the main text, but can continue past p.\ 14 if needed. 
% \clearpage  % TODO FINAL: This \clearpage needs to be removed from both review and camera-ready versions.

% \section*{Acknowledgements}
% Please insert your acknowledgments here.


%\bibliographystyle{splncs04}
%\bibliography{main}

% Originally here:

\clearpage

\includepdf[pages=-,offset=0 0]{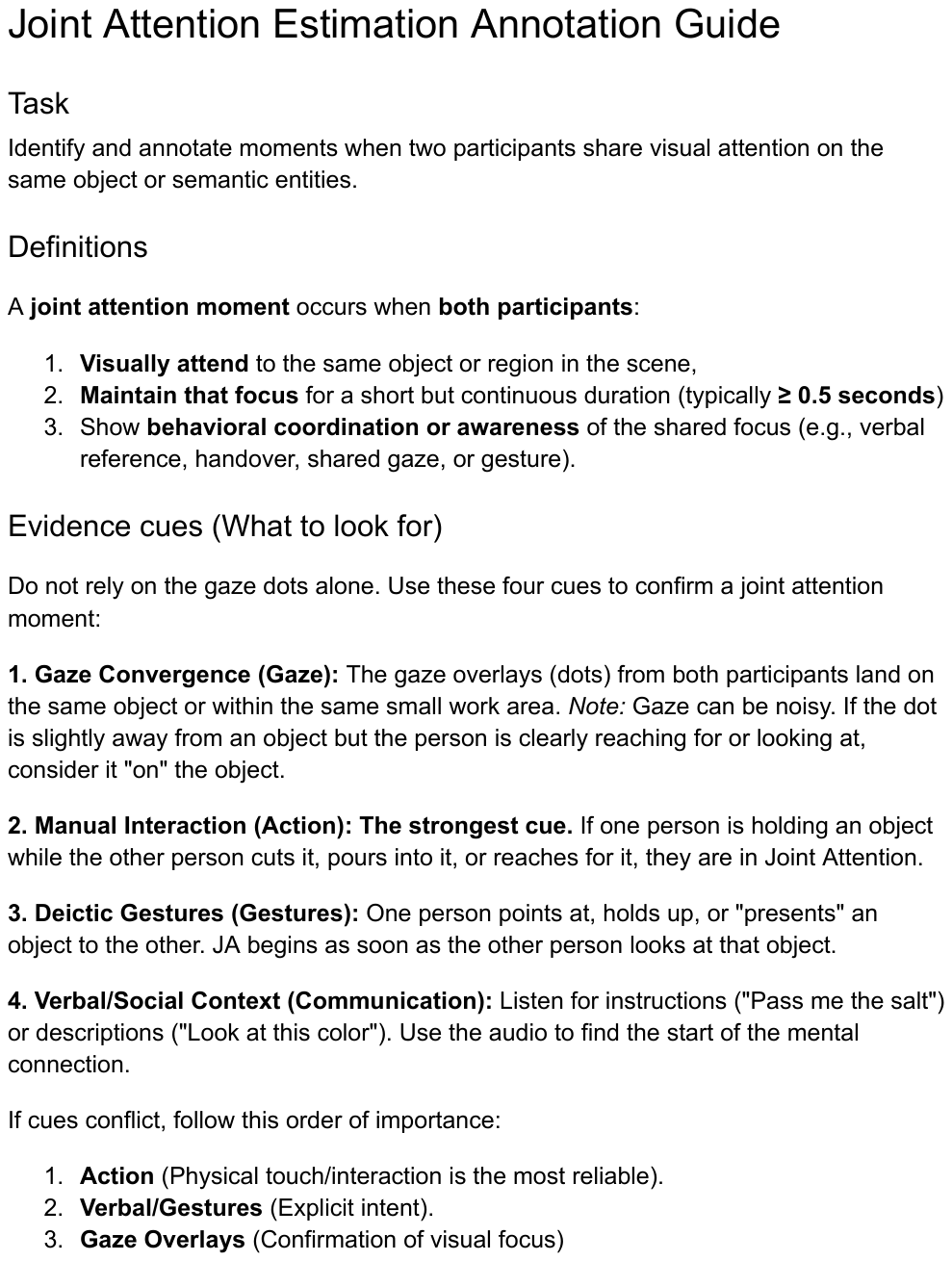}
\includepdf[pages=-,offset=0 0]{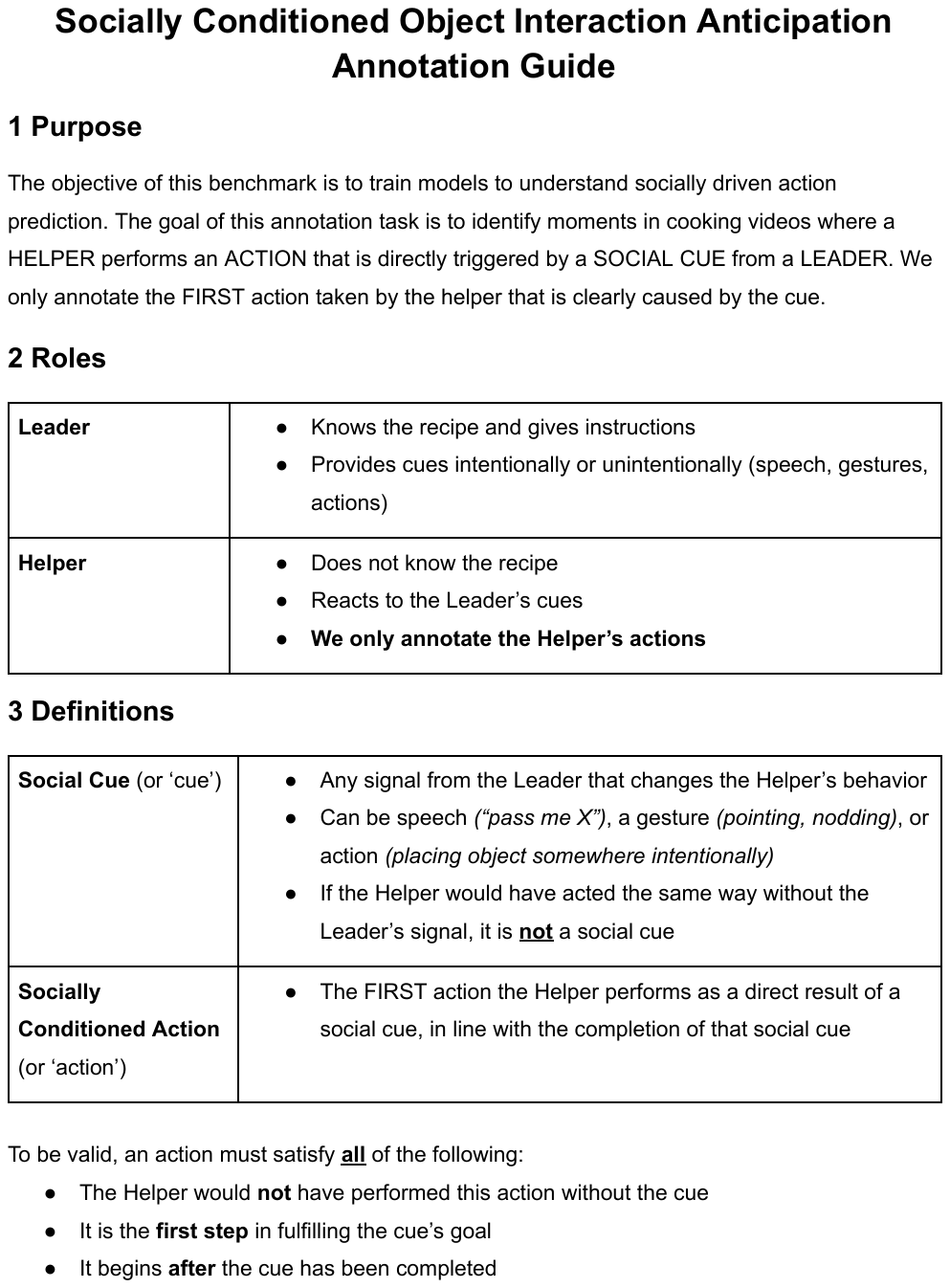}
\includepdf[pages=-,offset=0 0]{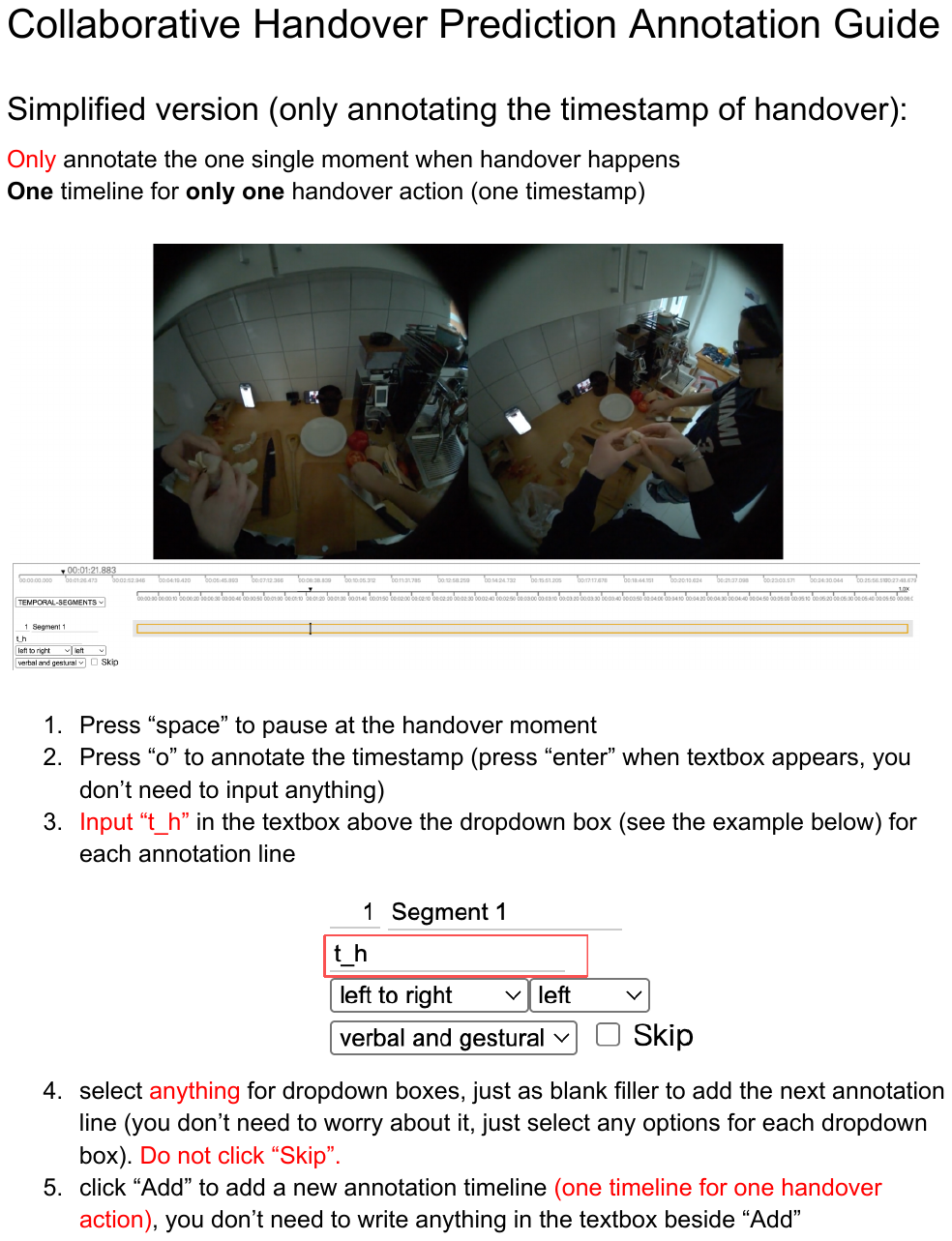}

\end{document}